\documentclass[10pt,twocolumn,letterpaper]{article}

\usepackage{cvpr}              

\usepackage[dvipsnames]{xcolor}

\definecolor{cvprblue}{rgb}{0.21,0.49,0.74}
\usepackage[pagebackref,breaklinks,colorlinks,citecolor=cvprblue]{hyperref}
\usepackage{xcolor,colortbl}
\usepackage{graphicx}
\usepackage{adjustbox}
\usepackage{verbatim} 
\usepackage{multirow}
\usepackage[accsupp]{axessibility}
 
\definecolor{oldgold}{rgb}{0.81, 0.71, 0.23}
\definecolor{silver}{rgb}{0.75, 0.75, 0.75}
\definecolor{copper}{rgb}{0.72, 0.45, 0.2}

\definecolor{custom_colour}{rgb}{0.86, 0.82, 1.0}

\definecolor{lightpastelpurple}{rgb}{0.69, 0.61, 0.85}
\definecolor{lilac}{rgb}{0.78, 0.64, 0.78}
\definecolor{wisteria}{rgb}{0.79, 0.63, 0.86}
\definecolor{mediumpurple}{rgb}{0.58, 0.44, 0.86}

\newcommand{\mtwo}{$\mathcal{M}_{2D \rightarrow 3D}$}
\newcommand{\mthree}{$\mathcal{M}_{3D \rightarrow 2D}$}

\def\paperID{7284} 
\def\confName{CVPR}
\def\confYear{2024}

\title{Multimodal Industrial Anomaly Detection by Crossmodal Feature Mapping}

\author{ 
    Alex Costanzino$^*$ \hspace{0.5cm} Pierluigi Zama Ramirez$^*$ \hspace{0.5cm} Giuseppe Lisanti \hspace{0.5cm} Luigi Di Stefano \\
    \small CVLAB, Department of Computer Science and Engineering (DISI) -- University of Bologna, Italy \\
    \normalsize\url{https://cvlab-unibo.github.io/CrossmodalFeatureMapping/}
}

\begin{document}
\maketitle
\def\thefootnote{*}\footnotetext{\emph{These authors contributed equally to this work}.}
\begin{abstract}
\label{sec:abstract}
    Recent advancements have shown the potential of leveraging both point clouds and images to localize anomalies. 
    Nevertheless, their applicability in industrial manufacturing is often constrained by significant drawbacks, such as the use of memory banks, which lead to a substantial increase in terms of memory footprint and inference time.
    We propose a novel light and fast framework that learns to map features from one modality to the other on nominal samples and detect anomalies by pinpointing inconsistencies between observed and mapped features.
    Extensive experiments show that our approach achieves state-of-the-art detection and segmentation performance, in both the standard and few-shot settings, on the MVTec 3D-AD dataset while achieving faster inference and occupying less memory than previous multimodal AD methods. 
    Furthermore, we propose a layer pruning technique to improve memory and time efficiency with a marginal sacrifice in performance.
\end{abstract}
\section{Introduction}
\label{sec:introduction}
    Industrial Anomaly Detection (AD) aims to identify unusual characteristics or defects in products, serving as a vital component within quality inspection processes. 
    Collecting data to exemplify anomalies is challenging due to their rarity and unpredictability. 
    Therefore, most works focus on unsupervised approaches, \ie, algorithms trained only on samples without defects, also referred to as \emph{nominal} samples. 
    Currently, most existing AD methods are geared toward analyzing RGB images. However, in many industrial settings, anomalies are hard to recognize effectively based solely on colour images, \eg, due to varying light conditions conducive to false detection and surface deviations that may not appear as unlikely colours. 
    Deploying colour images and surface information acquired by 3D sensors can tackle the above issues and substantially improve AD. 
    
\begin{figure}[t]
        \centering
            \includegraphics[width=\linewidth]{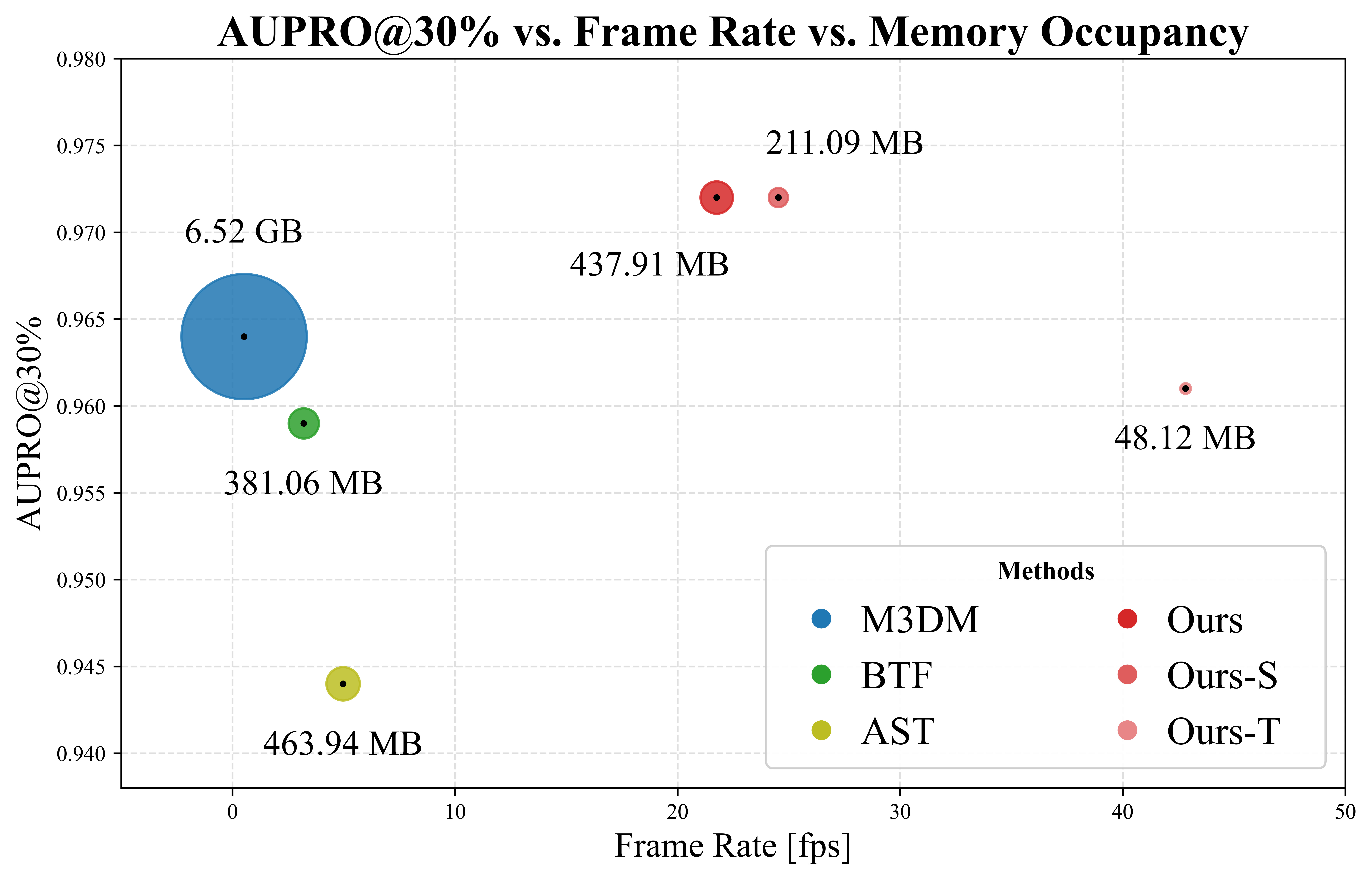}
        \caption{\textbf{Performance, speed and memory occupancy of Multimodal Anomaly Detection methods.}
        The chart reports defect segmentation performance (AUPRO@30\%) vs inference speed (Frame Rate on an NVIDIA 4090 GPU). 
        }
        \label{fig:teaser}
    \end{figure}

    Recently, researchers have started to explore novel avenues thanks to the introduction of benchmark datasets for 3D anomaly detection, such as MVTec 3D-AD~\cite{bergmann2022mvtec} and Eyecandies~\cite{bonfiglioli2022eyecandies}. 
    Indeed, both provide RGB images alongside pixel-registered 3D information for all data samples, thereby fostering the development of new, multimodal AD approaches~\cite{wang2023multimodal, horwitz2023back, RudWeh2023}.
    Unsupervised multimodal AD methods like BTF~\cite{horwitz2023back} and M3DM~\cite{wang2023multimodal} rely on large memory banks of multimodal features. 
    They achieve excellent performance (AUPRO@30\% metric in~\cref{fig:teaser}) at the cost of extensive memory requirements and slow inference (\cref{fig:teaser}). 
    In particular, M3DM outperforms BTF by leveraging frozen feature extractors trained by self-supervision on large datasets, \ie, ImageNet and Shapenet, for 2D and 3D features, respectively.
    Another recent multimodal method, AST~\cite{RudWeh2023}, follows a teacher-student paradigm conducive to a faster architecture (\cref{fig:teaser}).  
    Yet, AST does not exploit the spatial structure of the 3D data but employs this information just as an additional input channel in a 2D network architecture. 
    This results in inferior performance compared to M3DM and BTF  (\cref{fig:teaser}).

\begin{figure*}[ht]
    \centering
        \includegraphics[width=0.88\linewidth]{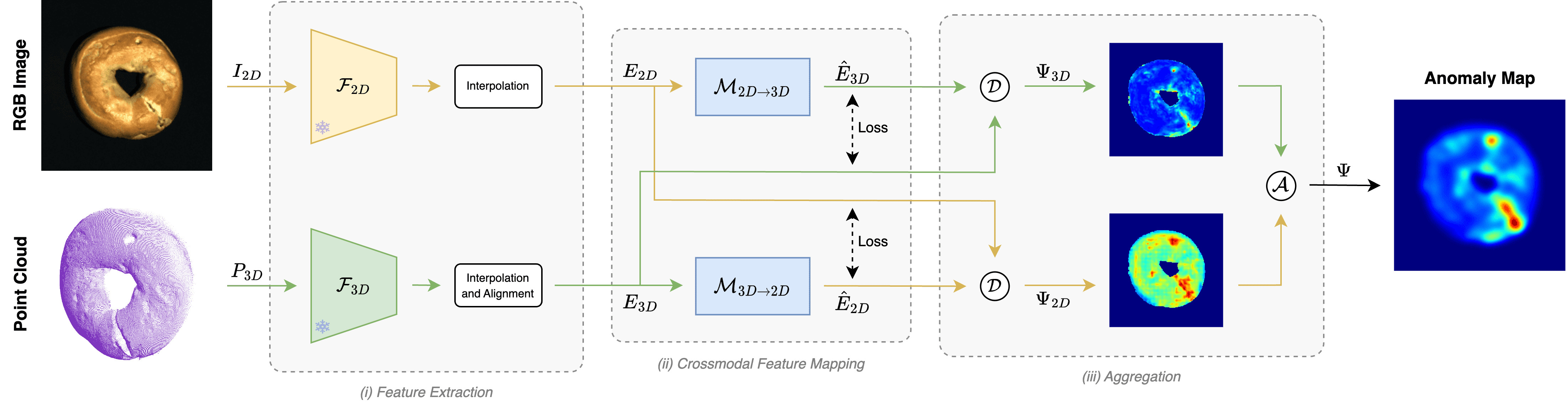}
    \caption{\textbf{Proposed pipeline.} Given an RGB Image $I_{2D}$ and a Point Cloud $P_{3D}$: a pair of feature extractors,  $\mathcal{F}_{2D},\mathcal{F}_{3D}$,  extract pixel-aligned feature maps, $E_{2D},E_{3D}$, by Transformer architectures. Then, a pair of crossmodal feature mappings, $\mathcal{M}_{2D \to 3D}, \mathcal{M}_{3D \to 2D}$, map the extracted features from one modality to the other, processing the features at each pixel independently. Lastly, extracted, $E_{2D},E_{3D}$ and mapped, $\hat{E}_{3D},\hat{E}_{3D}$, features are compared through a discrepancy function $\mathcal{D}$, to create modality-specific anomaly maps, $\Psi_{2D},\Psi_{3D}$, that are then combined by an aggregation function, $\mathcal{A}$, to obtain the final anomaly map $\Psi$.}
    \label{fig:architecture}
\end{figure*}

    %
    %
    In this paper, we propose a novel paradigm to exploit the relationship between features extracted from different modalities and improve multimodal AD. 
    The core idea behind our method, described in~\cref{fig:architecture}, is to learn two \emph{crossmodal} mapping functions, $\mathcal{M}_{2D \to 3D}$  and  $\mathcal{M}_{3D \to 2D}$, between the latent spaces of frozen 2D and 3D feature extractors, $\mathcal{F}_{2D}$ and $\mathcal{F}_{3D}$, respectively. 
    Thus, given a 2D feature computed by the 2D extractor, $\mathcal{M}_{2D \to 3D}$ learns to predict the corresponding 3D feature calculated by the 3D extractor, and, likewise, $\mathcal{M}_{3D \to 2D}$ learns to predict a 2D feature for a given 3D feature.
    %
    As we learn the two mapping functions on nominal data, we expect them to capture crossmodal relationships peculiar to good samples, while anomalies, by their quintessential nature, realize mappings unseen at training time, such as a 2D feature never observed in conjunction with a certain 3D feature, or vice versa.
    Hence, at inference time, we compute an anomaly map $\Psi$ by estimating and aggregating the discrepancies $(\Psi_{3D}, \Psi_{2D})$ between the actual features provided by the two frozen extractors and those predicted by the crossmodal mapping functions.
    
    This framework is amenable to realising multimodal AD effectively and efficiently. 
    Indeed, no obvious, trivial solutions would lead the crossmodal mapping networks to generalize to defective samples.
    For instance, as input and output features are extracted from different modalities, the networks cannot learn identity mappings, as may have happened in previous reconstruction-based AD methods~\cite{liu2023deep}. 
    Moreover, as we will discuss in~\cref{sec:method}, modelling the relationship between 2D and 3D features in nominal data provides high sensitivity toward all kinds of anomalies.  
    Finally, the feature mapping functions can be implemented as lightweight neural networks, such as small and shallow MLPs.
    This yields very fast inference alongside limited memory occupancy.

    As shown in~\cref{fig:teaser}, our novel AD approach based on crossmodal mapping functions achieves state-of-the-art performance on MVTec 3D-AD, outperforming the best resource-intensive method based on memory banks (Ours vs M3DM), while delivering much faster inference. 
    Additionally, we have observed that learning mappings between features from shallower layers of the frozen extractors can yield massive gains in terms of memory requirements and inference speed with a relatively limited impact on the effectiveness of our method. 
    Thus, we can prune the deepest layers of both the 2D and 3D feature extractors to obtain \emph{Small} and \emph{Tiny} variants of our framework (\cref{fig:teaser}: Ours-S, Ours-T) that require much less memory and run faster. 
    Remarkably, the \emph{Small} architecture still provides state-of-the-art performance on MVTec 3D-AD while requiring less than half memory compared to the full model, whereas the \emph{Tiny} architecture runs almost twice as fast and outperforms BTF and AST.   
    Finally, we point out that our method can be trained even with a few nominal samples.
    To properly evaluate our approach in this challenging scenario, we build the first few-shot multimodal AD benchmark from MVTec 3D-AD, and we note that our method achieves state-of-the-art anomaly segmentation performance.
    
    Our contributions can be summarized as follows:
    \begin{itemize}
        \item We propose a novel framework for unsupervised multimodal AD based on mapping features across modalities;
        \item By using modality-specific features extracted from frozen 2D and 3D extractors, we attain state-of-the-art detection and segmentation performance on MVTec 3D-AD, while reaching performance comparable to the state-of-the-art on Eyecandies. 
        \item Our method is capable of very fast inference and requires less memory than state-of-the-art solutions.
        \item We reach state-of-the-art performance on the proposed few-shot AD benchmark built on top of MVTec 3D-AD;
        \item We develop a strategy to prune networks without overly compromising performance. In this way, we achieve remarkably faster inference and large memory savings.
    \end{itemize}
    
\section{Related Work}
\label{sec:related_work}

    \noindent
    \textbf{Unsupervised Image Anomaly Detection.}
        Unsupervised AD approaches~\cite{liu2023deep} analyzing RGB Images can be divided into two broad categories.
        The general idea behind the first is to learn how to reconstruct images of nominal samples using auto-encoders~\cite{bergmann2018improving,zavrtanik2021draem,hou2021divide,ristea2022self}, in-painting~\cite{pirnay2022inpainting}, or diffusion models~\cite{wyatt2022anoddpm}. 
        Then, at test time, as the trained model cannot correctly reconstruct anomalous images, a per-pixel anomaly map can be computed by analyzing the discrepancy between the input and reconstructed image.
        The second category of approaches focuses instead on the feature space defined by deep neural networks~\cite{reiss2021panda, yi2020patch, zhang2021anomaly, sohn2021learning, yoa2021self, li2021cutpaste, yang2023memseg, massoli2021mocca, yu2021fastflow, rudolph2021same, rippel2021modeling, gudovskiy2022cflow, chiu2023self, defard2021padim, bergmann2020uninformed, wang2021student_teacher, cao2022informative, salehi2021multiresolution, deng2022anomaly}.
        Deep Feature Reconstruction (DFR)~\cite{yang2020dfr} trains an auto-encoder on the features extracted from nominal samples. 
        Then, similarly to image reconstruction methods, it identifies anomalies in the test samples by analysing the difference between reconstructed and original features.
        The increasing availability of effective, general-purpose feature extractors~\cite{caron2021emerging, oquab2023dinov2,he2022masked}, has fostered interest in anomaly detection methods that deploy features extracted by frozen models~\cite{patchcore2022roth,Cohen2020SubImageAD,bergman2020deep}. 
        At training time, the features computed from nominal samples by a frozen extractor are stored in a memory bank. 
        At inference time, the features extracted from the input image by the frozen model are compared to those stored in the bank to identify anomalies. 
        These approaches achieve remarkable performance, albeit at the cost of slow inference --- since each feature vector extracted from the input image has to be compared to all the nominal ones stored in the bank --- and significant memory occupancy --- since larger memory banks better capture the variability of nominal features. 
        
    \noindent
    \textbf{Multimodal RGB-3D Anomaly Detection.}
        Multimodal approaches exploit both RGB images and 3D data to enhance the robustness and effectiveness of anomaly detection.  
        Following the influential work on benchmarking image-based AD~\cite{bergmann2019mvtec}, a recent paper~\cite{bergmann2022mvtec} has introduced the MVTec 3D-AD dataset, alongside an experimental validation including several baselines, such as distribution mapping techniques based on GANs and variational models (\ie, VAEs), as well as auto-encoders. 
        Inspired by PatchCore~\cite{patchcore2022roth}, BTF~\cite{horwitz2023back} investigates the use of memory banks for 3D anomaly detection.
        The authors propose to add 3D features to the 2D features provided by a frozen convolutional model to enhance anomaly detection performance.
        They test several 3D features and achieve the best results using hand-crafted descriptors extracted from Point Clouds~\cite{fpfh}.
        M3DM~\cite{wang2023multimodal} improved over BTF by employing rich and distinctive 2D and 3D features extracted by frozen Transformer-based \emph{foundation models} trained by self-supervision on large datasets. 
        The authors also propose a learned function to fuse 2D and 3D features into multimodal features stored in memory banks alongside those computed from the individual modalities. 
        However, reliance on large feature banks renders M3DM overly expensive in terms of memory and time (\cref{fig:teaser}). 
        Similarly to M3DM~\cite{wang2023multimodal}, our method deploys 2D and 3D features computed by frozen Transformer-based models. Yet, we do not employ any memory bank and, instead, propose a novel crossmodal feature mapping paradigm that can be realized by two lightweight neural networks. 
        Using the same feature extractors as M3DM~\cite{wang2023multimodal}, we achieve better performance on MVTec 3D-AD while requiring way less memory and running remarkably faster (\cref{fig:teaser}).    
\section{Method}
\label{sec:method}
    Our multimodal AD approach relies on learning crossmodal mappings between features extracted from nominal samples to pinpoint anomalies based on the discrepancy between predicted and observed features. 
    %
    As depicted in \cref{fig:architecture}, this is realized by \emph{(i)} a pair of frozen feature extractors $\mathcal{F}_{2D},\mathcal{F}_{3D}$; \emph{(ii)} a pair of feature mappings networks $\mathcal{M}_{2D \to 3D}, \mathcal{M}_{3D \to 2D}$; and \emph{(iii)} an aggregation module.
    

    \subsection{Feature Extraction}
    \label{sec:extalign}
        The initial step in our pipeline involves extracting features for every pixel in a 2D image denoted as $I_{2D}$ and for each point in a 3D Point Cloud represented by $P_{3D}$. As explained in \cref{sec:introduction}, in our framework, both feature extractors have been trained on large external datasets and are kept frozen, \ie, their weights will never be updated.

        \noindent
        \textbf{2D Feature Extraction and Interpolation.}
        Given an image $I_{2D}$ with dimensions $H \times W \times C$
        , we process it with a 2D feature extractor, denoted as $\mathcal{F}_{2D}$, yielding a feature map with dimensions $H_{f} \times W_{f} \times D_{2D}$. Since the dimensions $H_{f}$ and $W_{f}$ are smaller than the original $H$ and $W$, we apply a bilinear upsampling operation to obtain $E_{2D}$, which is a feature map with dimensions $H \times W \times D_{2D}$, thereby obtaining a feature vector for each pixel location.

        \noindent
        \textbf{3D Feature Extraction and Interpolation.}
        Given a point cloud of dimensions $N \times 3$, we process it with a 3D feature extractor, $\mathcal{F}_{3D}$, obtaining a set of $N_{f}$ feature vectors of size $D_{3D}$. 
        Each feature vector, $f_\mathbf{c}$, is associated with a specific point within the original point cloud, $\mathbf{c} \in P_{3D}$. 
        Indeed, many 3D feature extractors (\eg,~\cite{pang2022masked}) do not estimate features for each input point but only for a subset of them, \ie, $N_{f} < N$.
        %
        Thus, to obtain a feature vector, $f_\mathbf{p}$, for each point of the cloud, $\mathbf{p} \in P_{3D}$, we follow a procedure similar to~\cite{wang2023multimodal}. 
        Here, $f_\mathbf{p}$ is computed as a weighted sum of the three feature vectors that, among the $N_{f}$ extracted by  $\mathcal{F}_{3D}$, have the closest centres to $\mathbf{p}$.
        In this way, we obtain ${E}'_{3D}$, a set of $N$ interpolated feature vectors of size $D_{3D}$.

        \noindent
        \textbf{Feature Alignment.}
        According to the standard setting in multimodal AD~\cite{bergmann2022mvtec,bonfiglioli2022eyecandies}, we assume pixel-registered 3D data and images. Thus, we know the corresponding pixel location associated with each 3D point. As $E_{2D}$ and ${E}'_{3D}$ have been interpolated to match the original image and point cloud resolutions, we can project ${E}'_{3D}$ into the 2D image plane, obtaining $E_{3D}$, a feature map of dimensions $H \times W \times D_{3D}$. In this process, we set to zero the vectors at the pixel locations where we do not have a corresponding 3D feature. 
        Finally, we apply a $3 \times 3$ smoothing kernel on $E_{3D}$. 
        At the end of this procedure, we obtain $E_{2D}$ and $E_{3D}$, two feature maps aligned at the pixel level.

    \subsection{Crossmodal Feature Mapping} 
    \label{sec:crossmodal_mapping}
        Once $E_{2D}$ and $E_{3D}$ have been obtained, we deploy two Feature Mapping functions, implemented as lightweight MLPs, \mtwo{} and \mthree{}. \mtwo{} maps a feature vector of size $D_{2D}$ into another one of size $D_{3D}$, while \mthree{} does the opposite.
        Each network predicts features of one modality from the other, processing each pixel location independently.
        Thus, given a pixel location $i$, and the corresponding 2D and 3D feature, $E^i_{2D}$ and $E^i_{3D}$, we can obtain the predicted feature of the other modality as:
        %
        %
        \begin{equation}
             \hat{E}^i_{3D} = \mathcal{M}_{2D \to 3D}(E^i_{2D}) \quad \hat{E}^i_{2D}= \mathcal{M}_{3D \to 2D}(E^i_{3D})
        \end{equation}
        %
        When processing pixel locations without a 3D point associated with it, we set to zero the corresponding predicted feature.
        By processing all pixels, we obtain the predicted feature maps $\hat{E}_{3D}, \hat{E}_{2D}$, of dimensions $H \times W \times D_{2D}$ and $H \times W \times D_{3D}$, respectively. 
        
        \noindent
        \textbf{Training.}
        At training time, \mtwo{} and \mthree{} are jointly optimized on all the nominal samples of a dataset by minimizing the cosine distance between the feature maps computed from the input data of both modalities and the predicted ones.
        Thus, the per-pixel loss is:
        \begin{equation}
            \mathcal{L}^{i} = 
            \Bigg(1 - \frac{E_{2D}^i \cdot \hat{E}_{2D}^i}{\|E_{2D}^i\| \|\hat{E}_{2D}^i\|}\Bigg) + \Bigg(1 - \frac{E_{3D}^i \cdot \hat{E}_{3D}^i}{\|E_{3D}^i\| \|\hat{E}_{3D}^i\|}\Bigg) 
        \end{equation}
        %

        \begin{figure}[t]
            \centering
            \includegraphics[width=0.98\linewidth]{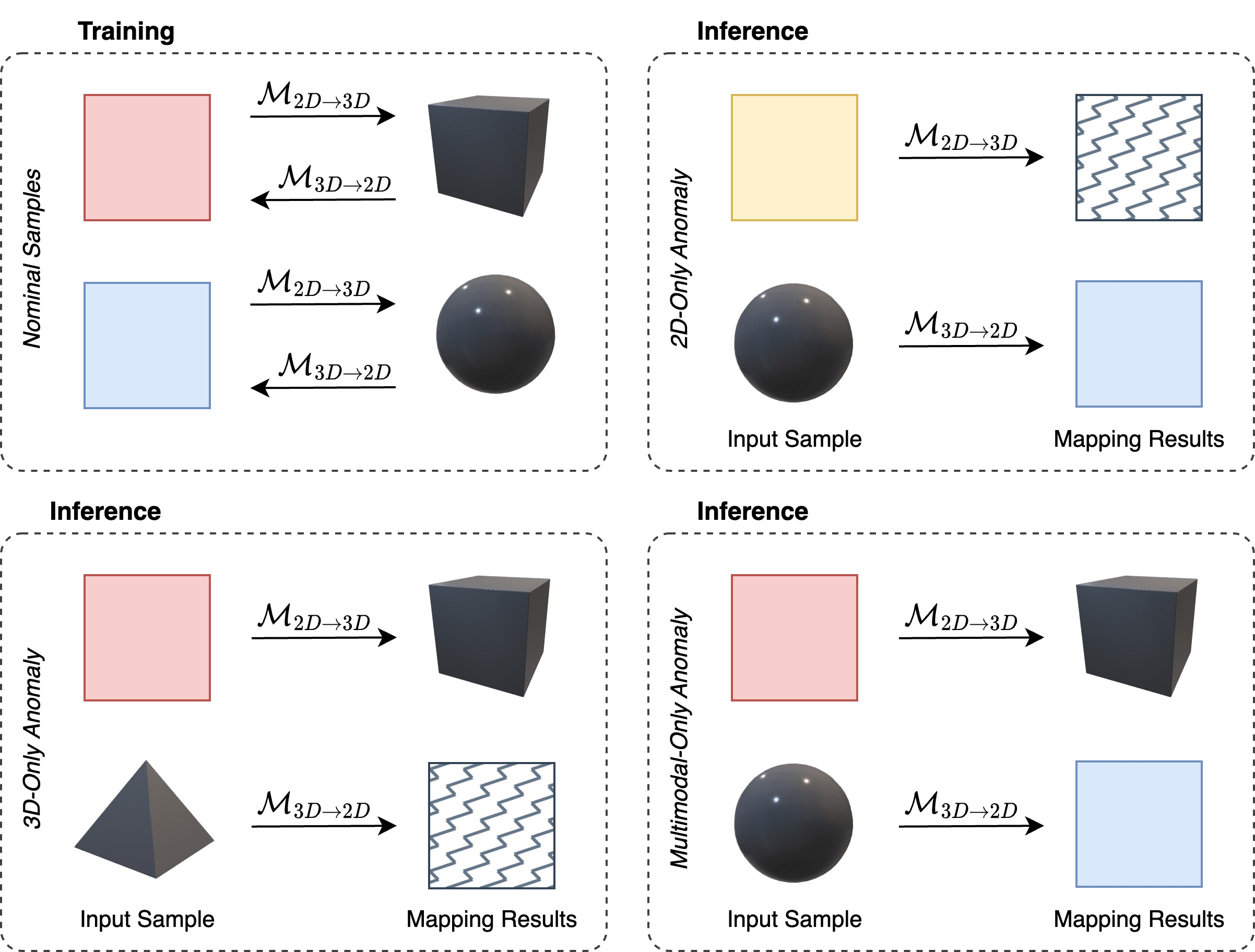}
            \caption{\textbf{Toy example of anomaly scenarios with corresponding behaviour of cross-modal mappings.} \emph{Top left:} Nominal samples. \emph{Top right:} 2D-Only with an RGB anomaly. \emph{Bottom left:} 3D-Only with an anomalous shape. \emph{Bottom right:} Multimodal-Only, with nominal RGB and 3D data but anomalous correlation.}
            \label{fig:sketch}
        \end{figure}

        \noindent
        \textbf{Rationale.}
        As pointed out in \cref{sec:introduction}, this novel paradigm offers high sensitivity toward all kinds of anomalies. Let us conceptualize this property with the toy examples presented in \cref{fig:sketch}.
        At training time (top left), we observe red 2D patterns on flat 3D surfaces and blue 2D patterns on curved 3D surfaces:   $\mathcal{M}_{2D \to 3D}$ and $\mathcal{M}_{3D \to 2D}$ learn to predict the relationships between the features extracted from these data. 
        At inference time, if an anomalous, \eg, yellow, 2D pattern appears on a curved surface  (top right), $\mathcal{M}_{3D \to 2D}$ predicts the 2D feature corresponding to a blue pattern, whilst the observed 2D feature concerns a yellow pattern. 
        Moreover, $\mathcal{M}_{2D \to 3D}$ receives an input feature unseen at training time, which would unlikely yield as output the 3D feature of the actual curved surface. Thus, our method senses a discrepancy between prediction and observation for the 2D and the 3D features.
        %
        Similar considerations apply to a nominal 2D pattern on an anomalous 3D surface (bottom left): both predictions disagree with the observations. 
        This is the case also when both modalities exhibit anomalies (not shown in \cref{fig:sketch}): both inputs are unseen at training time, so both crossmodal predictions are unlikely to match the observations. 
        Finally, we highlight the case mandating multimodal AD: the individual modalities comply with the nominal distributions, but their co-occurrence is anomalous.
        This may be exemplified by a red pattern on a curved surface  (bottom right): again, as $\mathcal{M}_{2D \to 3D}$ outputs the 3D feature of the flat patch and $\mathcal{M}_{3D \to 2D}$ the 2D feature of the blue one, both predictions disagree with the observations. 
    
        It is worth pointing out that, due to the variability of the nominal samples, the mappings between 2D and 3D features may not be \emph{unique}. For instance, in \cref{fig:sketch}, there might be both flat and curved surfaces coloured in red, and this \emph{one-to-many} mapping makes it hard for $\mathcal{M}_{2D \to 3D}$ to learn the correct 3D feature to be associated to the 2D feature of a red patch. Consequently, when presented with a red patch, $\mathcal{M}_{2D \to 3D}$ may predict the wrong 3D feature or an unlikely one, causing a discrepancy between the predicted and observed 3D features.
        Yet, $\mathcal{M}_{3D \to 2D}$ can  predict the 2D feature of the red patch, due to the $3D \to 2D$ mapping being \emph{many-to-one}. Thus, we may avoid a false detection by pinpointing anomalies only when both predictions disagree with the observations. Of course, due to even higher variability across nominal samples, we may also face \emph{one-to-many}  $3D \to 2D$ mappings, \eg, considering again \cref{fig:sketch}, both blue and red image patches on curved 3D patches. In such a case, when presented with a red patch on a curved surface, $\mathcal{M}_{2D \to 3D}$ may wrongly predict the feature of a flat patch and $\mathcal{M}_{3D \to 2D}$ that of a blue patch, ending up in a false anomaly detection due to both predictions disagreeing with the observations.  

        Nonetheless, in our framework, we can address the issue of potential  \emph{one-to-many} feature mappings across modalities by leveraging on the highly contextualized 2D and 3D features provided by Transformer architectures~\cite{caron2021emerging, pang2022masked}. Indeed, a contextualized 2D feature, \eg, describing a red patch surrounded by blue and purple patches, tends to correspond to a specific contextualized 3D feature, \eg, representing a flat patch just to the right of a rippling surface area. In other words, the highly contextualized 2D and 3D features extracted by Transformers are less prone to realize \emph{one-to-many} crossmodal mappings. 
        For the above reasons, we employ Transformers  for both   $\mathcal{F}_{2D}$ and $\mathcal{F}_{3D}$.

    \subsection{Aggregation}
    \label{sec:anomaly_score}
        At inference time, test samples are forwarded to the Feature Extraction and Mapping networks to obtain two pairs of extracted and predicted feature maps $(E_{2D}, \hat{E}_{2D}), (E_{3D}, \hat{E}_{3D})$.
        After  $\ell_2$-normalization of  all the individual feature vectors, the extracted and predicted maps are compared pixel-wise by a discrepancy function, $\mathcal{D}$, to obtain modality-specific anomaly maps $\Psi_{2D}, \Psi_{3D}$:
        \begin{equation}
            \Psi_{2D} = \mathcal{D}(E_{2D}, \hat{E}_{2D}) \quad \Psi_{3D} = \mathcal{D}(E_{3D}, \hat{E}_{3D})
        \end{equation}
        We employ the Euclidean distance as discrepancy $\mathcal{D}$.
        %
        %
        
        The above anomaly maps are then combined using an aggregation function $\mathcal{A}$ to get the final anomaly map $\Psi = \mathcal{A}(\Psi_{2D}, \Psi_{3D})$.
        As discussed in~\cref{sec:crossmodal_mapping} with the help of~\cref{fig:sketch}, pinpointing anomalies only when both predictions disagree with the observations provides high sensitivity across all kinds of anomalies and good robustness toward false detection. Therefore, we use the pixel-wise product as aggregation function: $\Psi = \Psi_{2D} \cdot \Psi_{3D}$, which can be thought of as a logical AND: the anomaly score at any pixel location is high only if this is so for both the modality-specific scores, \ie, anomaly detection must be corroborated by both modalities.  


        The aggregated anomaly map is finally smoothed by a Gaussian of kernel with $\sigma = 4$, similarly to common practice~\cite{wang2023multimodal, patchcore2022roth, Cohen2020SubImageAD}.
        The global anomaly score required to perform sample-level anomaly detection is obtained as the maximum value of the anomaly map $\Psi$. 
    
        \begin{table*}[ht]
    \centering
    \resizebox{0.75\textwidth}{!}{%
        \begin{tabular}{c||c||cccccccccc||c}
            &
            \textbf{Method} &
            \textit{Bagel} &
            \textit{Cable Gland} &
            \textit{Carrot} &
            \textit{Cookie} &
            \textit{Dowel} &
            \textit{Foam} &
            \textit{Peach} &
            \textit{Potato} &
            \textit{Rope} &
            \textit{Tire} &
            \textbf{Mean} \\ 

            \hline
            
            \multirow{10}{*}{\rotatebox{90}{\textbf{I-AUROC}}}
            & DepthGAN~\cite{bergmann2022mvtec} & 0.538 & 0.372 & 0.580 & 0.603 & 0.430 & 0.534 & 0.642 & 0.601 & 0.443 & 0.577 & 0.532 \\
            & DepthAE~\cite{bergmann2022mvtec}  & 0.648 & 0.502 & 0.650 & 0.488 & 0.805 & 0.522 & 0.712 & 0.529 & 0.540 & 0.552 & 0.595 \\
            & DepthVM~\cite{bergmann2022mvtec}  & 0.513 & 0.551 & 0.477 & 0.581 & 0.617 & 0.716 & 0.450 & 0.421 & 0.598 & 0.623 & 0.555 \\
            & VoxelGAN~\cite{bergmann2022mvtec} & 0.680 & 0.324 & 0.565 & 0.399 & 0.497 & 0.482 & 0.566 & 0.579 & 0.601 & 0.482 & 0.517 \\
            & VoxelAE~\cite{bergmann2022mvtec}  & 0.510 & 0.540 & 0.384 & 0.693 & 0.446 & 0.632 & 0.550 & 0.494 & 0.721 & 0.413 & 0.538 \\
            & VoxelVM~\cite{bergmann2022mvtec}  & 0.553 & 0.772 & 0.484 & 0.701 & 0.751 & 0.578 & 0.480 & 0.466 & 0.689 & 0.611 & 0.609 \\
            & BTF~\cite{horwitz2023back}        & 0.918 & 0.748 & 0.967 & 0.883 & 0.932 & 0.582 & 0.896 & 0.912 & 0.921 & 0.886 & 0.865 \\
            & AST~\cite{RudWeh2023}             & 0.983 & 0.873 & 0.976 & 0.971 & 0.932 & 0.885 & \textbf{0.974} & \textbf{0.981} & \textbf{1.000} & 0.797 & 0.937 \\
            & M3DM~\cite{wang2023multimodal}    & \textbf{0.994} & \textbf{0.909} & \underline{0.972} & {0.976} & {0.960} & \textbf{0.942} & \underline{0.973} & 0.899 & 0.972 & {0.850} & {0.945} \\
            & Ours                              & \textbf{0.994} & \underline{0.888} & \textbf{0.984} & \textbf{0.993} & \underline{0.980} & {0.888} & 0.941 & {0.943} & \underline{0.980} & \textbf{0.953} & \underline{0.954} \\
            & Ours-M                            & \underline{0.988} & 0.875 & \textbf{0.984} & \underline{0.992} & \textbf{0.997} & \underline{0.924} & {0.964} & \underline{0.949} & 0.979 & \underline{0.950} & \textbf{0.960} \\

            \hline 
            
            \multirow{10}{*}{\rotatebox{90}{\textbf{AUPRO@30\%}}} 
            & DepthGAN~\cite{bergmann2022mvtec} & 0.421 & 0.422 & 0.778 & 0.696 & 0.494 & 0.252 & 0.285 & 0.362 & 0.402 & 0.631 & 0.474 \\
            & DepthAE~\cite{bergmann2022mvtec}  & 0.432 & 0.158 & 0.808 & 0.491 & 0.841 & 0.406 & 0.262 & 0.216 & 0.716 & 0.478 & 0.481 \\
            & DepthVM~\cite{bergmann2022mvtec}  & 0.388 & 0.321 & 0.194 & 0.570 & 0.408 & 0.282 & 0.244 & 0.349 & 0.268 & 0.331 & 0.335 \\
            & VoxelGAN~\cite{bergmann2022mvtec} & 0.664 & 0.620 & 0.766 & 0.740 & 0.783 & 0.332 & 0.582 & 0.790 & 0.633 & 0.483 & 0.639 \\
            & VoxelAE~\cite{bergmann2022mvtec}  & 0.467 & 0.750 & 0.808 & 0.550 & 0.765 & 0.473 & 0.721 & 0.918 & 0.019 & 0.170 & 0.564 \\
            & VoxelVM~\cite{bergmann2022mvtec}  & 0.510 & 0.331 & 0.413 & 0.715 & 0.680 & 0.279 & 0.300 & 0.507 & 0.611 & 0.366 & 0.471 \\
            & BTF~\cite{horwitz2023back}        & {0.976} & 0.969 & 0.979 & \textbf{0.973} & 0.933 & 0.888 & 0.975 & {0.981} & 0.950 & 0.971 & 0.959 \\
            & AST~\cite{RudWeh2023}             & 0.970 & 0.947 & \underline{0.981} & 0.939 & 0.913 & 0.906 & {0.979} & \underline{0.982} & 0.889 & 0.940 & 0.944 \\
            & M3DM~\cite{wang2023multimodal}    & 0.970 & \underline{0.971} & 0.979 & \underline{0.950} & {0.941} & {0.932} & 0.977 & 0.971 & {0.971} & {0.975} & {0.964} \\
            & Ours                              & \underline{0.979} & \textbf{0.972} & \textbf{0.982} & 0.945 & \underline{0.950} & \textbf{0.968} & \underline{0.980} & \underline{0.982} & \underline{0.975} & \underline{0.981} & \underline{0.971} \\
            & Ours-M                            & \textbf{0.980} & {0.966} & \textbf{0.982} & {0.947} & \textbf{0.959} & \underline{0.967} & \textbf{0.982} & \textbf{0.983} & \textbf{0.976} & \textbf{0.982} & \textbf{0.972} \\
            
            \hline
            
        \end{tabular}}
    \caption{\textbf{I-AUROC and AUPRO@30\% on MVTec 3D-AD for multimodal AD methods.} Best results in \textbf{bold}, runner-ups \underline{underlined}.}
    \label{tab:aupro_iauroc_comparisons_mvtec}
\end{table*}

        \begin{table*}[!ht]
    \centering
    \resizebox{0.75\textwidth}{!}{%
        \begin{tabular}{c||cccccccccc||c}

            \textbf{Method} &
            \textit{Bagel} &
            \textit{Cable Gland} &
            \textit{Carrot} &
            \textit{Cookie} &
            \textit{Dowel} &
            \textit{Foam} &
            \textit{Peach} &
            \textit{Potato} &
            \textit{Rope} &
            \textit{Tire} &
            \textbf{Mean} \\ 

            \hline

            BTF~\cite{horwitz2023back}        & {0.428} & 0.365 & 0.452 & {0.431} & 0.370 & 0.244 & 0.427 & {0.470} & 0.298 & 0.345 & 0.383 \\
            AST~\cite{RudWeh2023}             & 0.388 & 0.322 & {0.470} & 0.411 & 0.328 & 0.275 & \underline{0.474} & \underline{0.487} & 0.360 & {0.474} & {0.398} \\
            M3DM~\cite{wang2023multimodal}    & 0.414 & {0.395} & 0.447 & 0.318 & \textbf{0.422} & {0.335} & 0.444 & 0.351 & {0.416} & 0.398 & 0.394 \\
            Ours                              & \underline{0.459} & \textbf{0.431} & \underline{0.485} & \textbf{0.469} & {0.394} & \textbf{0.413} & {0.468} & \underline{0.487} & \underline{0.464} & \underline{0.476} & \underline{0.455} \\
            Ours-M                            & \textbf{0.480} & \underline{0.398} & \textbf{0.490} & \underline{0.467} & \underline{0.413} & \underline{0.408} & \textbf{0.481} & \textbf{0.494} & \textbf{0.468} & \textbf{0.488} & \textbf{0.459} \\
            
            \hline
            
        \end{tabular}}
    \caption{\textbf{AUPRO@1\% on MVTec 3D-AD for multimodal AD methods.} Best results in \textbf{bold}, runner-ups \underline{underlined}.}
    \label{tab:aupro_fpr-rate_mvtec}
\end{table*}

    \subsection{Layers Pruning}
    \label{sec:pruning}
\begin{figure}[t]
        \centering
            \includegraphics[width=\linewidth]{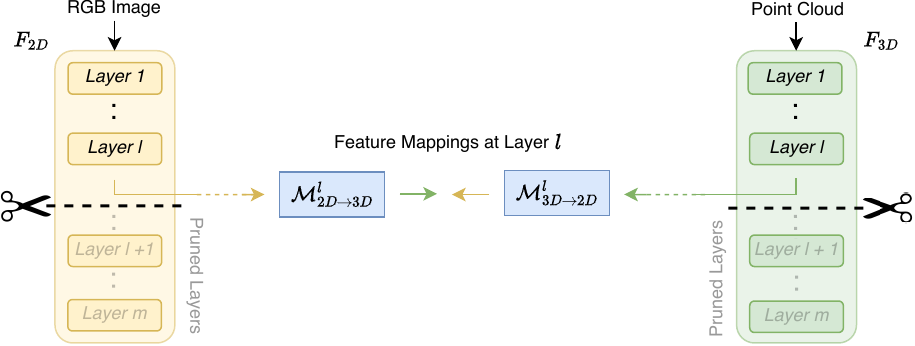}
        \caption{
        \textbf{Layers Pruning.} The Feature Mapping networks can be fed with features from different layers of the two Transformers.} 
        \label{fig:pruning}
    \end{figure}

        %
        The Feature Extractors employed in our solution~\cite{caron2021emerging, pang2022masked} are based on Transformer encoders composed of $m$ layers. 
        The distinguishing factor between features at different layers lies in the varying degree of self-attention processing applied to the original input. As the input features descend the encoder layers, they exhibit an increased contextualization. 
        We observed that learning mappings between features from shallower layers of the frozen extractors can yield remarkable gains in terms of memory requirements and inference speed with a limited impact on effectiveness. 
        %
        Thus, as shown in \cref{fig:pruning}, we perform layer-pruning by choosing an intermediate layer $l$ in the 2D and 3D frozen feature extractors ($\mathcal{F}_{2D}, \mathcal{F}_{3D}$) and discarding those from $l + 1$ to the last.
        Consequently, crossmodal networks, $M^{l}_{2D \rightarrow 3D}$ and $M^{l}_{3D \rightarrow 2D}$, map features of layer $l$ of the 2D encoder into those of layer $l$ of the 3D encoder and vice versa.
        For instance, in Ours-T ($l=1$), we trim both encoders after the first layer, discarding those from second to last, and apply the crossmodal mappings on the features extracted by the first layers.
        In contrast, our reference model learns the crossmodal mapping networks between features from the last layer of both encoders.
        %
        %


\section{Experimental Settings}
\label{sec:experimental_settings}
    
    \textbf{Datasets and Metrics.} 
    We evaluate our framework on two multimodal AD benchmarks.  MVTec 3D-AD~\cite{bergmann2022mvtec} consists of $10$ categories of industrial objects, totalling $2656$ train samples, $294$ validation samples and $1197$ test samples. 
    Eyecandies~\cite{bonfiglioli2022eyecandies} is a synthetic dataset featuring photo-realistic images of $10$ categories of food items in an industrial conveyor scenario. It contains $10k$ train samples, $1k$ validation samples and $4k$ test samples. 
    Both datasets provide RGB images alongside pixel-registered 3D information for each sample. Thus, we have RGB information at each pixel location paired with $(x,y,z)$ coordinates.
    We employ the evaluation metrics proposed by MVTec 3D-AD. Thus, we assess image anomaly detection performance by the Area Under the Receiver Operator Curve (I-AUROC) computed on the global anomaly score.
    We estimate the anomaly segmentation performance by the pixel-level Area Under the Receiver Operator Curve (P-AUROC) and the Area Under the Per-Region Overlap (AUPRO).
    All previous works employ $0.3$ as the False Positive Rate (FPR) integration threshold to calculate the AUPRO. 
    We reckon such a value may often turn out too loose for real industrial applications, allowing too many false positives. 
    Hence, we also compute AUPRO based on the tighter $0.01$ threshold. 
    We denote AUPROs with integration thresholds $0.3$ and $0.01$ as AUPRO@30\%, and AUPRO@1\%, respectively.
    We report results with additional thresholds in the Supplementary.

    \noindent
    \textbf{Implementation Details.}
    We employ the same frozen Transformers as  M3DM~\cite{wang2023multimodal} to realize the $\mathcal{F}_{2D}$ and $\mathcal{F}_{3D}$ feature extractors, \ie, DINO ViT-B/8~\cite{kolesnikov2021vit, caron2021emerging} trained on ImageNet~\cite{deng2009imagenet} and Point-MAE~\cite{pang2022masked} trained on ShapeNet~\cite{shapenet2015}, respectively. 
    Thus, $\mathcal{F}_{2D}$  processes $224 \times 224$ RGB images and outputs $28 \times 28 \times 768$ feature maps, which are bi-linearly up-sampled to $224 \times 224 \times 768$ before feeding the features to \mtwo{}.  
    $\mathcal{F}_{3D}$ processes $1024$ groups of $32$ points obtained with FPS~\cite{pointent2}, yielding a feature vector of dimensionality $1152$ for each group.
    As described in in~\cref{sec:extalign}, these features are interpolated and aligned to $224 \times 224 \times 1152$  before being fed to \mthree{}.

    Both \mtwo{} and \mthree{} consist of just three linear layers, each but the last one followed by GeLU activations.  
    The number of units per layer is $768, 960, 1152$ for \mtwo{} and $1152, 960, 768$ for \mthree{}.
    The two networks are trained jointly for $250$ epochs using Adam~\cite{adam} with a learning rate of $0.001$.
    
    As done in~\cite{horwitz2023back, wang2023multimodal, RudWeh2023}, we fit a plane with RANSAC on the 3D point cloud and consider a point as background if the distance to the plane is less than $0.005$. 
    Background points are discarded from the
    in input to $\mathcal{F}_{3D}$.
    This procedure accelerates the processing of 3D features 
    and mitigates background noise in anomaly maps. 

    Moreover, as described in~\cref{sec:pruning} to obtain lighter versions of our framework we prune both feature extractors at layer $l$ equal to $1$, $4$, $8$, to obtain \emph{Tiny}, \emph{Small} and \emph{Medium} architectures referred to as Ours-T, Ours-S, and Ours-M. 
    
    We conducted experiments using both our and the original code from the authors of other multimodal AD methods on a single NVIDIA GeForce RTX 4090.

\section{Experiments}
\label{sec:experiments}

    \begin{table}[t]
    \centering
    \resizebox{\linewidth}{!}{%
        \begin{tabular}{c||cccc}

            \textbf{Method} &
            \textbf{I-AUROC} &
            \textbf{P-AUROC} &
            \textbf{AUPRO@30\%} &
            \textbf{AUPRO@1\%} \\
            
            \hline
            
            AST~\cite{RudWeh2023}             & 0.758 & 0.902 & 0.878 & 0.224 \\
            M3DM~\cite{wang2023multimodal}    & \textbf{0.897} & \textbf{0.977} & \underline{0.882} & \underline{0.331} \\
            Ours                              & \underline{0.881} & \underline{0.974} & \textbf{0.887} & \textbf{0.335} \\
            Ours-M                            & 0.865 & 0.973 & 0.880 & 0.330 \\
            
            \hline
            
        \end{tabular}}
    \caption{\textbf{Eyecandies Results.} Average metrics of 10 classes on the test set. Best results in \textbf{bold}, runner-ups \underline{underlined}.}
    \label{tab:global_comparisons_eyecandies}
\end{table}

    \begin{table*}[ht]
    \centering
    \setlength{\tabcolsep}{2pt}
    \resizebox{\linewidth}{!}{
        \begin{tabular}{c||c|c|c|c}
            \begin{tabular}{c}
            \\
            \hline
            \textbf{Method} \\
            \hline
            BTF~\cite{horwitz2023back} \\
            AST~\cite{RudWeh2023} \\
            M3DM~\cite{wang2023multimodal} \\
            Ours \\
            \hline
            \end{tabular}
            &
            \begin{tabular}{cccc}
                \textit{5-shot} &
                \textit{10-shot} &
                \textit{50-shot} &
                \textit{Full} \\
                \hline
                \multicolumn{4}{c}{\textbf{I-AUROC}} \\
                \hline
                0.671 & \underline{0.695} & 0.806 & 0.865 \\
                0.680 & 0.689 & 0.794 & 0.937 \\
                \textbf{0.822} & \textbf{0.845} & \textbf{0.907} & \underline{0.945} \\
                \underline{0.811} & \textbf{0.845} & \underline{0.906} & \textbf{0.954} \\
                \hline
            \end{tabular}
                 & 
            \begin{tabular}{cccc}
                \textit{5-shot} &
                \textit{10-shot} &
                \textit{50-shot} &
                \textit{Full} \\
                \hline
                \multicolumn{4}{c}{\textbf{P-AUROC}} \\
                \hline
                0.980 & 0.983 & \underline{0.989} & \underline{0.992} \\
                0.950 & 0.946 & 0.974 & 0.976 \\
                \underline{0.984} & \underline{0.986} & \underline{0.989} & \underline{0.992} \\
                \textbf{0.986} & \textbf{0.987} & \textbf{0.991} & \textbf{0.993} \\           
                \hline
            \end{tabular}
                &
            \begin{tabular}{cccc}
                \textit{5-shot} &
                \textit{10-shot} &
                \textit{50-shot} &
                \textit{Full} \\
                \hline
                \multicolumn{4}{c}{\textbf{AUPRO@30\%}} \\
                \hline
                0.920 & 0.928 & 0.947 & 0.959 \\
                0.903 & 0.835 & 0.929 & 0.944 \\
                \underline{0.937} & \underline{0.943} & \underline{0.955} & \underline{0.964} \\
                \textbf{0.949} & \textbf{0.954} & \textbf{0.965} & \textbf{0.971} \\            \hline
            \end{tabular}
                &
            \begin{tabular}{cccc}
                \textit{5-shot} &
                \textit{10-shot} &
                \textit{50-shot} &
                \textit{Full} \\
                \hline
                \multicolumn{4}{c}{\textbf{AUPRO@1\%}} \\
                \hline
                0.288 & 0.308 & 0.356 & 0.383 \\
                0.158 & 0.174 & 0.335 & \underline{0.398} \\
                \underline{0.330} & \underline{0.355} & \underline{0.387} & 0.394 \\
                \textbf{0.382} & \textbf{0.398} & \textbf{0.431} & \textbf{0.455} \\
                \hline
                \end{tabular}
        \end{tabular}}
    \caption{\textbf{Few-shot Anomaly Detection and Segmentation on the MVTec 3D-AD dataset.} Best results in \textbf{bold}, runner-ups \underline{underlined}.}
    \label{tab:ftn_few-shot_mvtec}
\end{table*}
    \begin{figure}
  \centering
  \setlength{\tabcolsep}{1pt}
  \scalebox{0.83}{
  \begin{tabular}{cccccc}
    & \textbf{Bagel} & \textbf{Carrot} & \textbf{Dowel} & \textbf{Peach} & \textbf{Rope} \\
    \rotatebox{90}{\hspace{0.25cm} RGB} & 
        \includegraphics[width=0.175\linewidth]{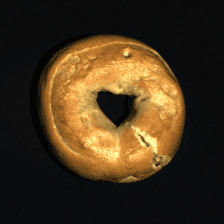} & 
        \includegraphics[width=0.175\linewidth]{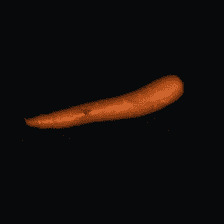} & 
        \includegraphics[width=0.175\linewidth]{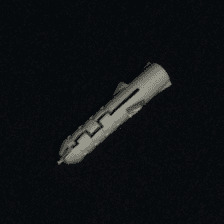} & 
        \includegraphics[width=0.175\linewidth]{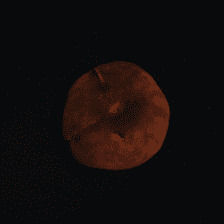} & 
        \includegraphics[width=0.175\linewidth]{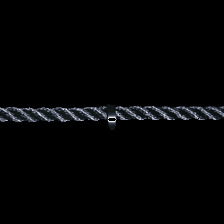} \\
    
    \rotatebox{90}{\hspace{0.35cm} PC} & 
        \includegraphics[width=0.175\linewidth]{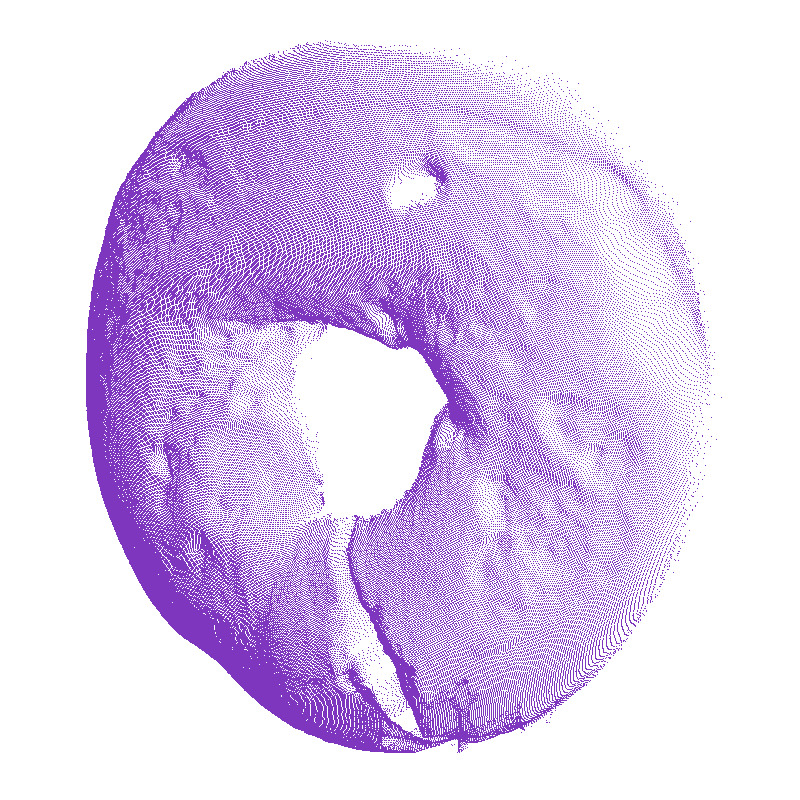} & 
        \includegraphics[width=0.175\linewidth]{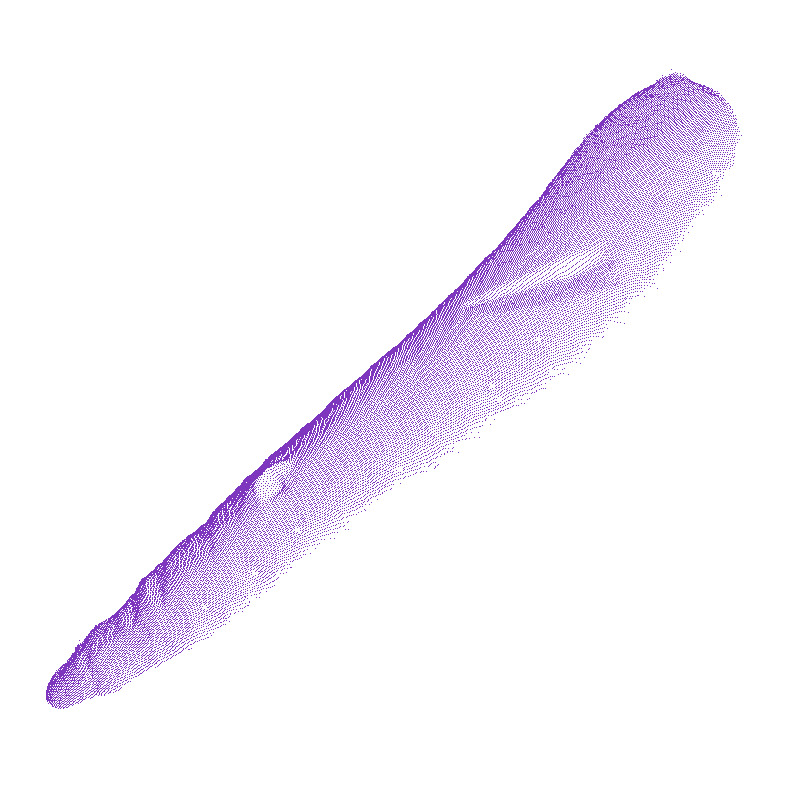} & 
        \includegraphics[width=0.175\linewidth]{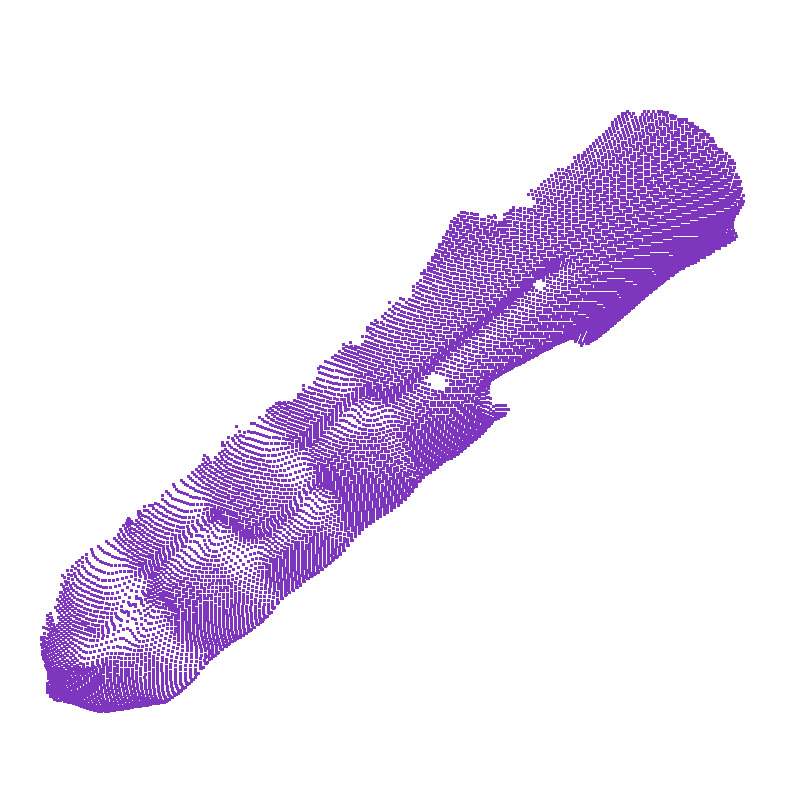} & 
        \includegraphics[width=0.175\linewidth]{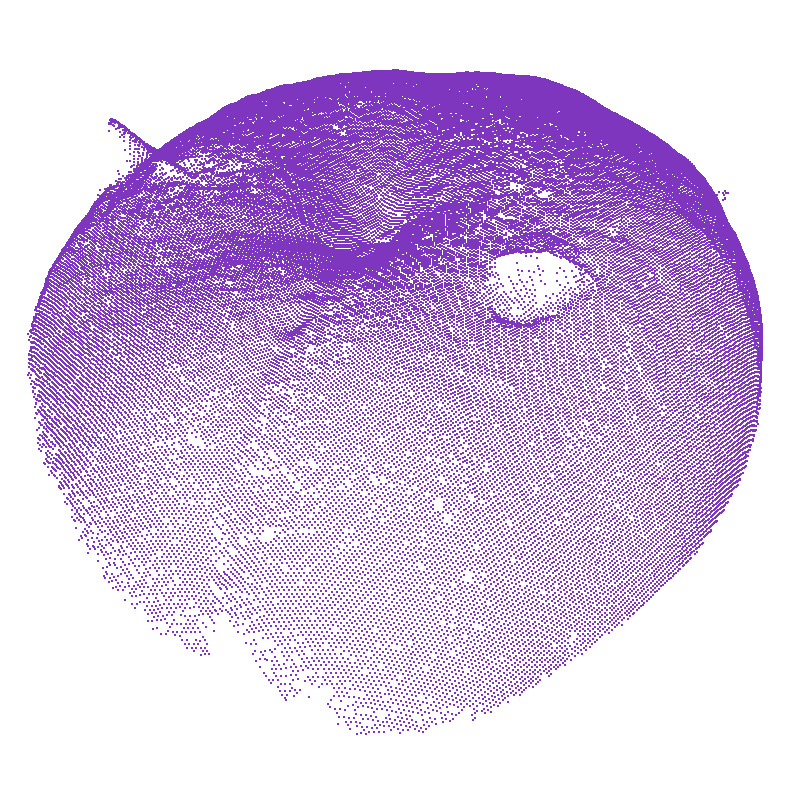} & 
        \includegraphics[width=0.175\linewidth]{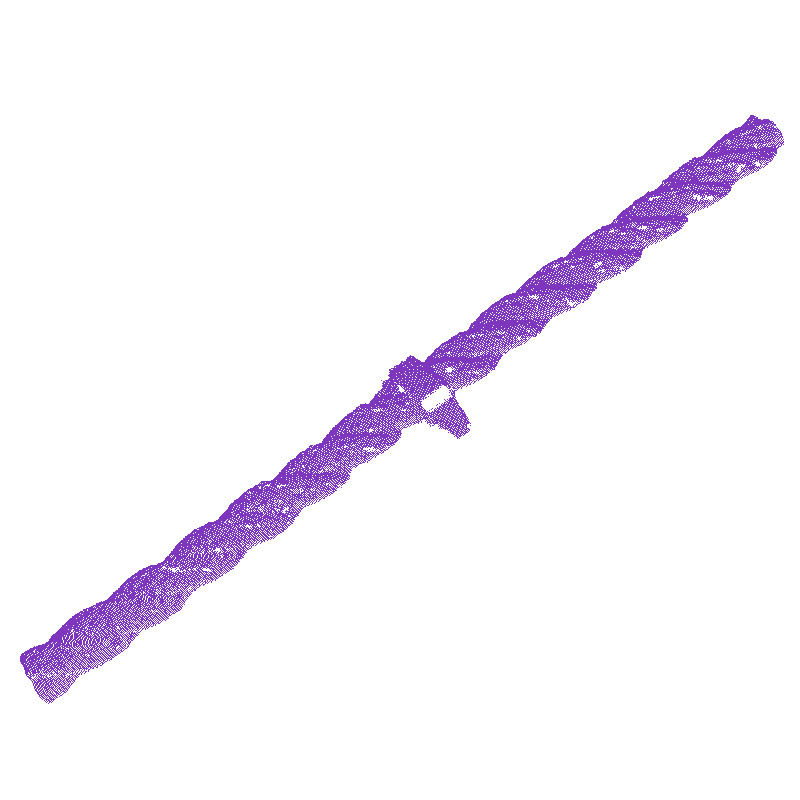} \\
    
    \rotatebox{90}{\hspace{0.35cm} GT} & 
        \includegraphics[width=0.175\linewidth]{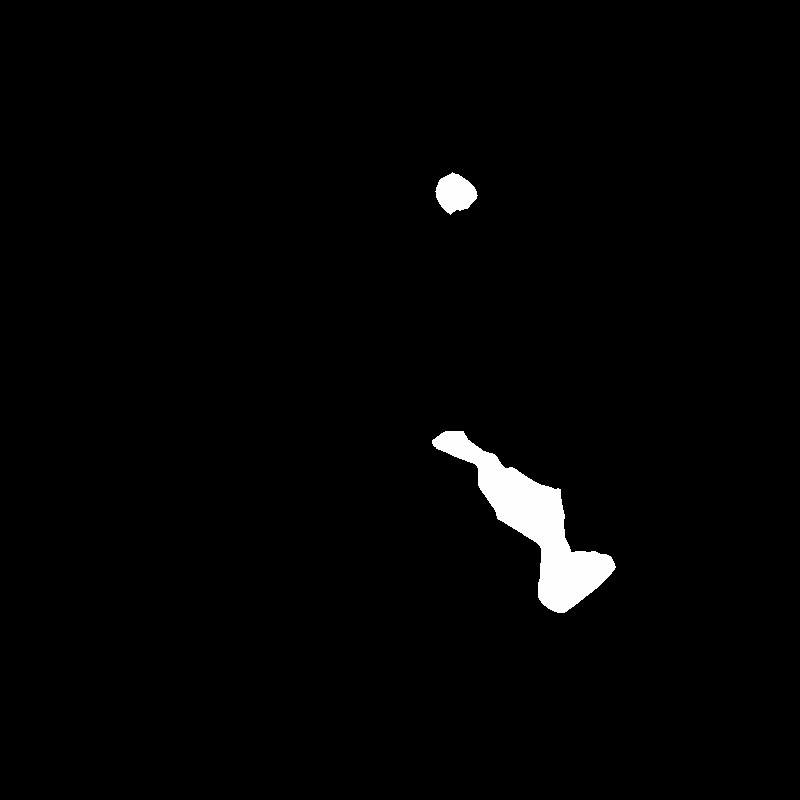} & 
        \includegraphics[width=0.175\linewidth]{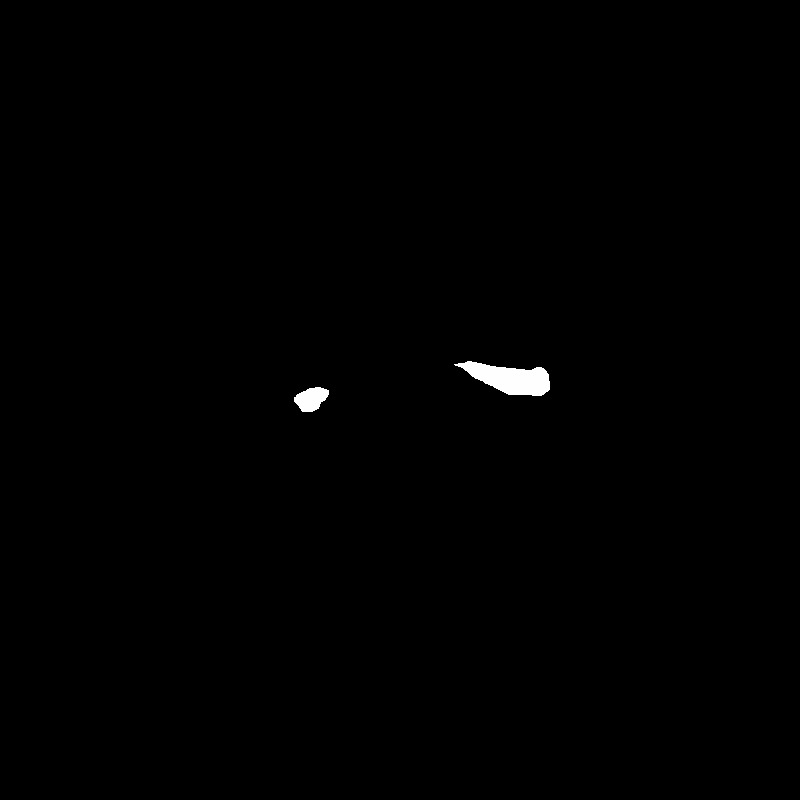} & 
        \includegraphics[width=0.175\linewidth]{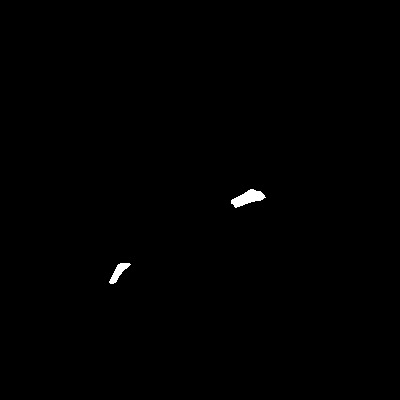} & 
        \includegraphics[width=0.175\linewidth]{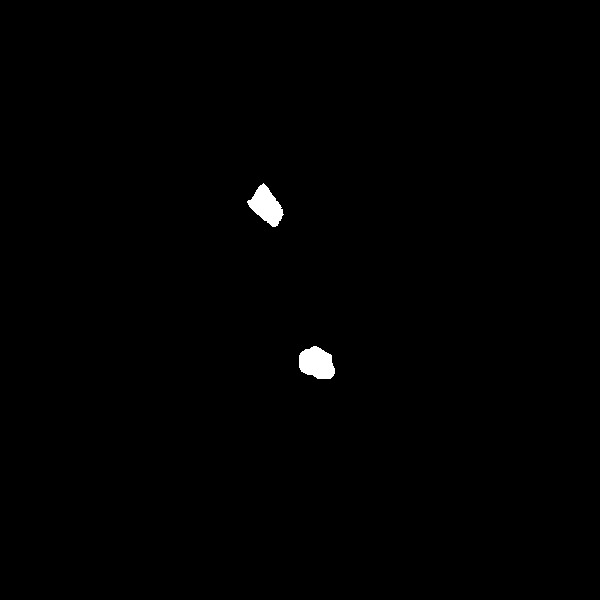} & 
        \includegraphics[width=0.175\linewidth]{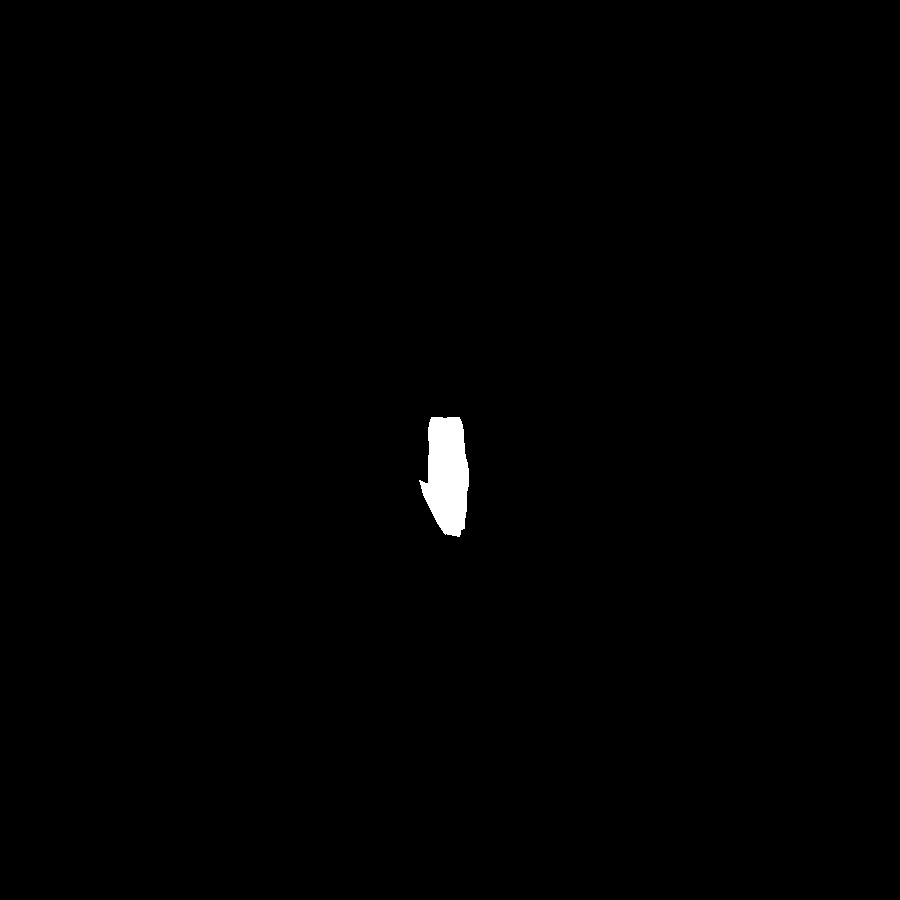} \\
    
    \rotatebox{90}{\hspace{0.1cm} M3DM} & 
        \includegraphics[width=0.175\linewidth]{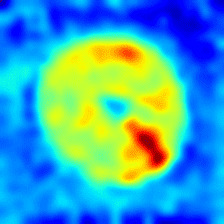} & 
        \includegraphics[width=0.175\linewidth]{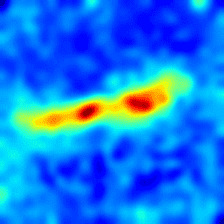} & 
        \includegraphics[width=0.175\linewidth]{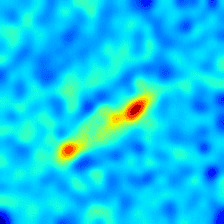} & 
        \includegraphics[width=0.175\linewidth]{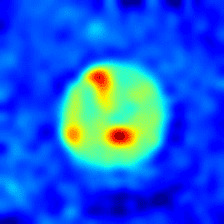} & 
        \includegraphics[width=0.175\linewidth]{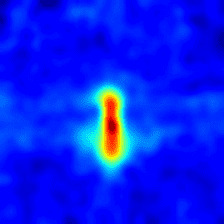} \\

    \rotatebox{90}{\hspace{0.2cm} Ours} & 
        \includegraphics[width=0.175\linewidth]{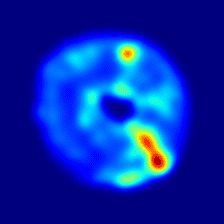} & 
        \includegraphics[width=0.175\linewidth]{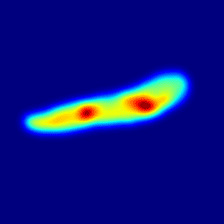} & 
        \includegraphics[width=0.175\linewidth]{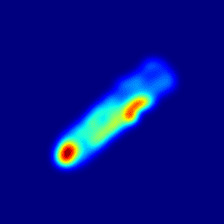} & 
        \includegraphics[width=0.175\linewidth]{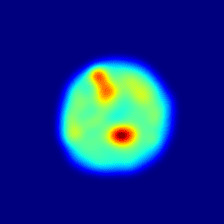} & 
        \includegraphics[width=0.175\linewidth]{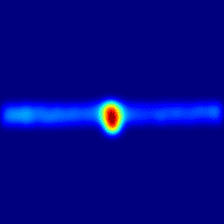} \\
        
  \end{tabular}}
    \caption{\textbf{MVTec-3D AD Qualitative Results.} From top to bottom: input RGB images, point clouds, GT anomaly segmentations, anomaly maps from M3DM, and anomaly maps with our method.}
    \label{fig:qualitatives_mvtec}

\end{figure}

    \noindent
    \textbf{Anomaly Detection and Segmentation.}
        Following the setups of~\cite{wang2023multimodal}, we evaluate our proposal on MVTec 3D-AD and Eyecandies, reporting results in \cref{tab:aupro_iauroc_comparisons_mvtec}, \cref{tab:aupro_fpr-rate_mvtec} and \cref{tab:global_comparisons_eyecandies}.
        Our method achieves the best results in detection and segmentation on MVTec 3D-AD, outperforming the previous state-of-the-art method, M3DM, in all the three mean metrics,  namely I-AUROC, AUPRO@30\% and AUPRO@1\%, as well as in most of the individual categories.   
        Comparison between \cref{tab:aupro_iauroc_comparisons_mvtec} and \cref{tab:aupro_fpr-rate_mvtec} shows how the performance of current AD methods turn out dramatically inferior when the evaluation sets a more challenging bar in terms of tolerable FPR. 
        As mentioned in~\cref{sec:experimental_settings}, we believe that such a challenge better matches the requirements of many real industrial AD applications. Therefore, we posit that the MVTec 3D-AD benchmark is far from saturated, and there exist vast margins of improvements in multimodal AD. 
        %
        As for Eyecandies, \cref{tab:global_comparisons_eyecandies}, we achieve performance comparable to M3DM, with two winning metrics for each method.  Moreover, we highlight that the P-AUROC metric seems almost saturated while also in Eyecandies there is substantial room for improvement in the AUPRO@1\% metric.
        %
        %
        %
        In \cref{fig:qualitatives_mvtec}, we show some qualitative results on the MVTec 3D-AD dataset. Compared to M3DM, our method provides remarkably sharper anomaly maps, well localized relatively to the ground-truth defect segmentation, thereby motivating the larger performance gap in terms of  AUPRO@1\%. 
        More extensive qualitative results are reported in the Supplementary Material.
        
        \begin{table}[t]
    \centering
    \resizebox{\linewidth}{!}{%
        \begin{tabular}{c||cc||cccc}

            \textbf{Method} &
            \textbf{Frame Rate} &
            \textbf{Memory} &
            \textbf{I-AUROC} &
            \textbf{P-AUROC} &
            \textbf{AUPRO@30\%} &
            \textbf{AUPRO@1\%} \\ 

            \hline

            BTF~\cite{horwitz2023back}     & 3.197  & 381.06  & 0.865 & 0.992 & 0.959 & 0.383 \\
            AST~\cite{RudWeh2023}          & 4.966  & 463.94  & 0.937 & 0.976 & 0.944 & 0.398 \\
            M3DM~\cite{wang2023multimodal} & 0.514  & 6526.12 & 0.945 & 0.992 & 0.964 & 0.394 \\
            Ours                           & 21.755 & 437.91  & 0.954 & 0.993 & 0.971 & 0.455 \\
            Ours-M                         & 24.146 & 295.81  & \textbf{0.960} & \textbf{0.994} & \textbf{0.972} & \textbf{0.459} \\
            Ours-S                         & 24.527 & 211.09  & 0.948 & 0.994 & 0.972 & 0.451 \\
            Ours-T                         & \textbf{42.818} & \textbf{48.12} & 0.899 & 0.990 & 0.961 & 0.419 \\
            \hline
            
        \end{tabular}}
    \caption{\textbf{Inference Speed, Memory Footprint and AD Performance on MVTec 3D-AD}. Frame Rate in fps and Memory in MB. }
    \label{tab:time-space_performance}
\end{table}

    \noindent
    \textbf{Few-shot Anomaly Detection and Segmentation.}
        In relevant industrial scenarios, collecting many nominal samples is extremely expensive or even unfeasible. 
        Thus, a desirable property of AD methods is the ability to model the distribution of nominal data by only a few samples. 
        To address this scenario, we define the first benchmark for few-shot multimodal AD based on the MVTec 3D-AD dataset. 
        We randomly select 5, 10, and 50 images from each category as training data. 
        We train the best multimodal methods, BTF~\cite{horwitz2023back}, M3DM~\cite{wang2023multimodal}, and AST~\cite{RudWeh2023} on these samples, and we test them on the entire MVTec 3D-AD test set, reporting the results in \cref{tab:ftn_few-shot_mvtec}.
        As for detection, our method achieves an I-AUROC comparable to M3DM~\cite{wang2023multimodal} while outperforming the other approaches. 
        We obtain the best segmentation performance for all metrics (P-AUROC, AUPRO@1\%, and AUPRO@30\%) in all the few-shot settings, significantly improving the most challenging segmentation metric (+0.052 AUPRO@1\% on 5-shot). 
        These results show that our framework enables learning general crossmodal relationships even from a few nominal samples. 

    \noindent
    \textbf{Frame Rate and Memory Occupancy.}
    \label{sec:timem}
        Computational efficiency is key to industrial AD. 
        Thus, we investigate the memory footprint and inference speed w.r.t. AD performance for the best multimodal approaches,  BTF~\cite{horwitz2023back}, M3DM~\cite{wang2023multimodal}, and AST~\cite{RudWeh2023}, as well as our method. 
        In addition, we report the performance of our framework by pruning the feature extractors at various levels using the technique described in~\cref{sec:pruning}.
        The results are reported in \cref{tab:time-space_performance}.
        %
        We compute inference speed in frames per second on the same machine equipped with an NVIDIA 4090 and Pytorch 1.13, reporting the average across all the test samples of MVTec 3D-AD.
        For each method, we include the time for each step of its \textit{inference} pipeline, from input pre-processing to the computation of anomaly scores, synchronizing all GPU threads before estimating the total inference time. We do not include training-only steps such as the memory bank creation. Regarding memory occupancy during inference, we consider network parameters, activations, and memory banks.         
        As expected, memory-bank methods (BTF~\cite{horwitz2023back} and M3DM~\cite{wang2023multimodal}) exhibit the lowest frame rate and the highest memory footprint.
        AST~\cite{RudWeh2023} requires only 26 MB more than our model, as it is based on two feed-forward networks. 
        However, it is still relatively slow ($4.966$ fps) since it is based on Normalizing Flow~\cite{papamakarios2021normalizing}.
        Our method has the highest frame rate ($21.755$ fps) and the lowest memory occupancy ($437.91$ MB) while outperforming competitors across all metrics.
        The pruned models Ours-M, Ours-S, and Ours-T are even more efficient with a marginal sacrifice in accuracy. 
        For instance, Ours-S occupies half of the memory of our full model and yet achieves state-of-the-art results on MVTec 3D-AD on all metrics. 
        Remarkably, Ours-T obtains state-of-the-art anomaly segmentation performance according to the most challenging metric (AUPRO@1\%=$0.419$) while running in real-time ($48.12$ fps).

    \noindent
    \textbf{Aggregation Analysis.}
        We investigate on the impact of the proposed product-based aggregation discussed in \cref{sec:anomaly_score}. 
        In \cref{tab:ftn_modality-aggregation_ablation}, we report the results obtained on MVTec 3D-AD by using the anomaly maps before aggregation, $\Psi_{2D}$ and $\Psi_{3D}$, or combined using different functions, such as pixel-wise sum $\Psi_{2D} + \Psi_{3D}$,  pixel-wise maximum $\max(\Psi_{2D}, \Psi_{3D})$, and  pixel-wise product $\Psi_{2D} \cdot \Psi_{3D}$.
        It is possible to note how the product performs best in both detection and segmentation.
        Indeed, considering as anomalous only points in which both $\Psi_{2D}$ and $\Psi_{3D}$ have high scores, enables discarding false positives that may occur when nominal relationships between RGB and 3D features are not unique, as discussed in \cref{sec:crossmodal_mapping}.
        \begin{table}[t]
    \centering
    \resizebox{\linewidth}{!}{%
        \begin{tabular}{c||cccc}

            \textbf{Anomaly Map} &
            \textbf{I-AUROC} &
            \textbf{P-AUROC} &
            \textbf{AUPRO@30\%} &
            \textbf{AUPRO@1\%} \\

            \hline

            $\Psi_{2D}$                    & 0.895 & 0.985 & 0.950 & 0.401 \\
            $\Psi_{3D}$                    & 0.885 & 0.987 & 0.956 & 0.403 \\
            $\Psi_{2D}$ + $\Psi_{3D}$      & 0.939 & 0.988 & 0.959 & 0.430 \\
            $\max(\Psi_{2D}$, $\Psi_{3D})$ & 0.895 & 0.985 & 0.950 & 0.400 \\
            $\Psi_{2D}\cdot\Psi_{3D}$      & \textbf{0.954} &  \textbf{0.993} & \textbf{0.971} & \textbf{0.455} \\

            \hline
            
        \end{tabular}}
    \caption{\textbf{Analysis of Aggregation Functions.} Results on MVTec 3D-AD. Best results in \textbf{bold}.}
    \label{tab:ftn_modality-aggregation_ablation}
\end{table}
        \begin{table}[t]
    \centering
    \resizebox{\linewidth}{!}{%
        \begin{tabular}{c||c||cccc}
            \textbf{Modality} & 
            \textbf{Anomaly Map} & 
            \textbf{I-AUROC} & 
            \textbf{P-AUROC} & 
            \textbf{AUPRO@30\%} & 
            \textbf{AUPRO@1\%} \\
            
            \hline
            
            Intra & $\Phi_{2D}$ & 0.860 & 0.980 & 0.932 & 0.361\\
            Intra & $\Phi_{3D}$ & 0.816 & 0.970 & 0.900 & 0.348\\
            Intra & $\Phi_{2D} \cdot \Phi_{3D}$ & 0.898 & 0.989 & 0.963 & 0.426\\

            \hline 
            
            Cross & $\Psi_{2D}$ & 0.865 & 0.982 & 0.944 & 0.382 \\
            Cross & $\Psi_{3D}$ & 0.885 & 0.985 & 0.952 & 0.391 \\
            Cross & $\Psi_{2D} \cdot \Psi_{3D}$ & \textbf{0.944} & \textbf{0.993} & \textbf{0.970} & \textbf{0.450} \\


            
            \hline
        
        \end{tabular}}
    \caption{\textbf{Crossmodal vs Intramodal.} Results on MVTec 3D-AD. Best results in \textbf{bold}. Networks are trained for 50 epochs.}
    \label{tab:ftn_crossmodality_ablation}
\end{table}

\begin{figure}[t]
    \centering
    \setlength{\tabcolsep}{1pt}
    \scalebox{0.5}{
    \begin{tabular}{cccccccccc}
        & 
        {\small RGB} & 
        {\small PC} & 
        {\small $E_{2D}$ } & 
        {\small  $E_{3D\rightarrow2D}$} & 
        {\small $\Psi_{2D}$} & 
        {\small $E_{3D}$ } & 
        {\small  $E_{2D\rightarrow3D}$} & 
        {\small $\Psi_{3D}$} & 
        {\small $\Psi_{2D} \cdot \Psi_{3D}$} \\
        
        \rotatebox{90}{\small \hspace{0.2cm} Nominal} &
        \includegraphics[width=0.2\linewidth]{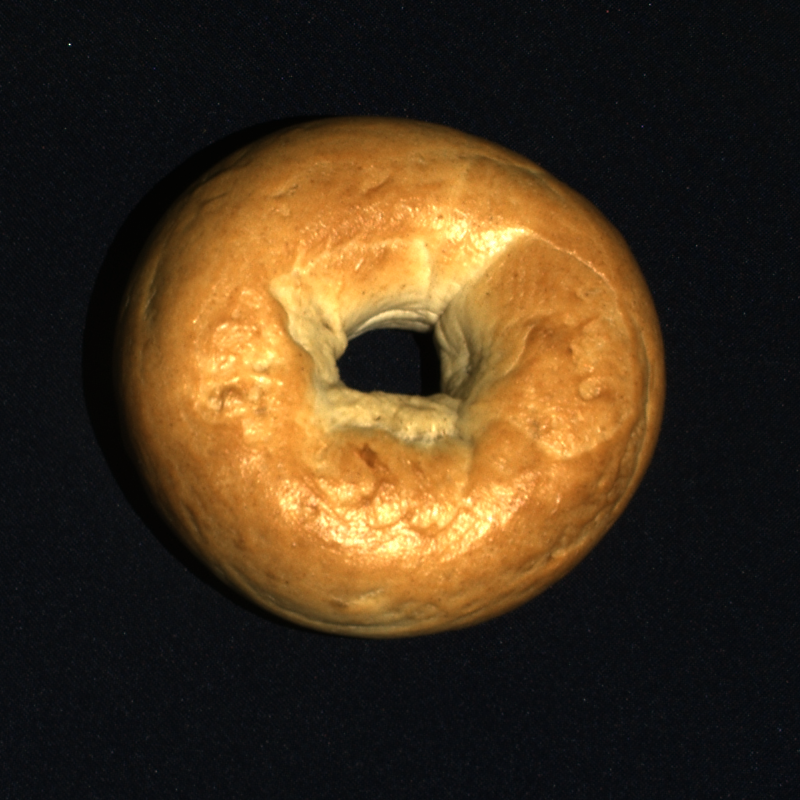} &
        \includegraphics[width=0.2\linewidth]{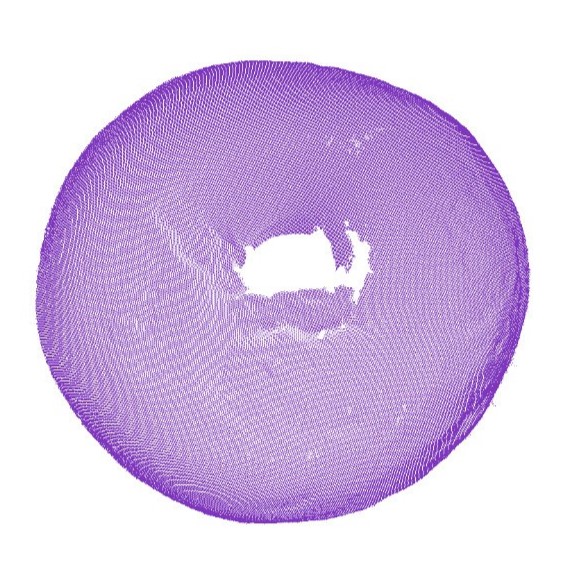} &
        \includegraphics[width=0.2\linewidth]{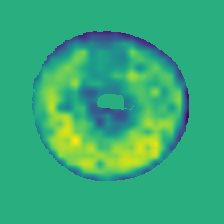} &
        \includegraphics[width=0.2\linewidth]{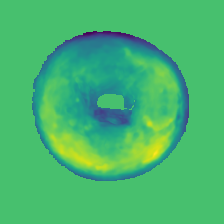} &
        \includegraphics[width=0.2\linewidth]{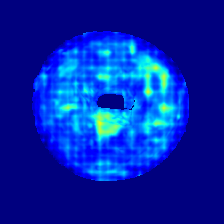} &
        \includegraphics[width=0.2\linewidth]{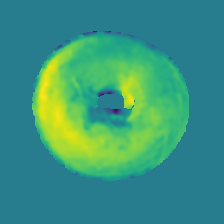} &
        \includegraphics[width=0.2\linewidth]{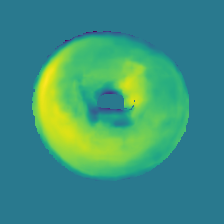} &
        \includegraphics[width=0.2\linewidth]{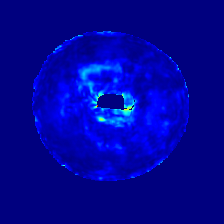} &
        \includegraphics[width=0.2\linewidth]{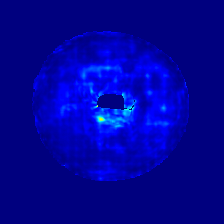} \\
        
        \rotatebox{90}{\small \hspace{0.4cm} Anomalous} &
        \includegraphics[width=0.2\linewidth]{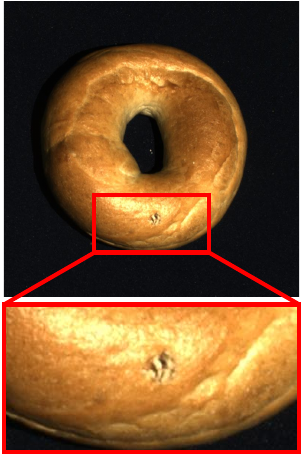} &
        \includegraphics[width=0.2\linewidth]{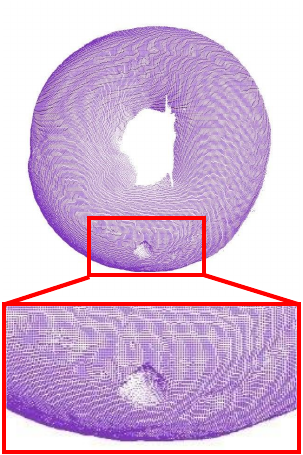} &
        \includegraphics[width=0.2\linewidth]{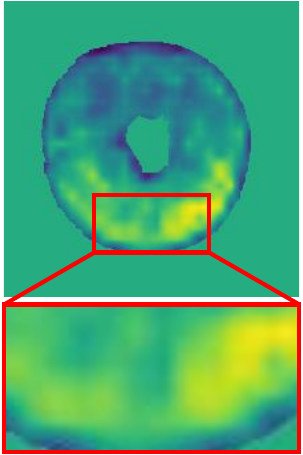} &
        \includegraphics[width=0.2\linewidth]{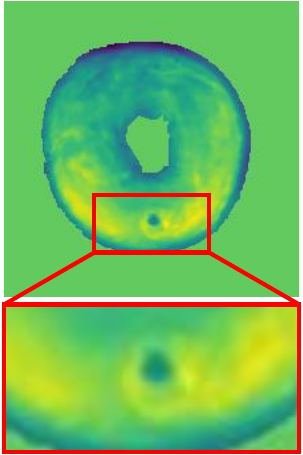} &
        \includegraphics[width=0.2\linewidth]{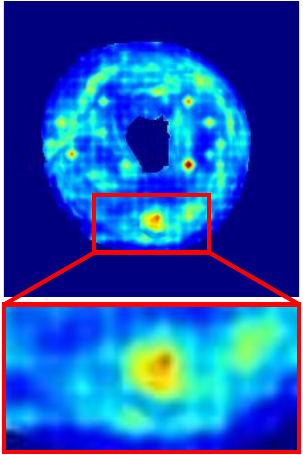} &
        \includegraphics[width=0.2\linewidth]{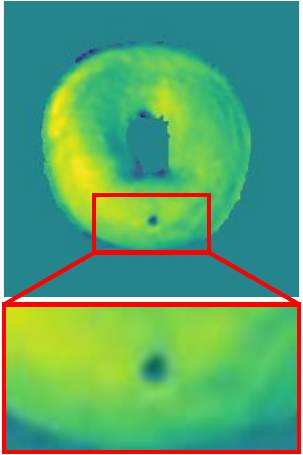} &
        \includegraphics[width=0.2\linewidth]{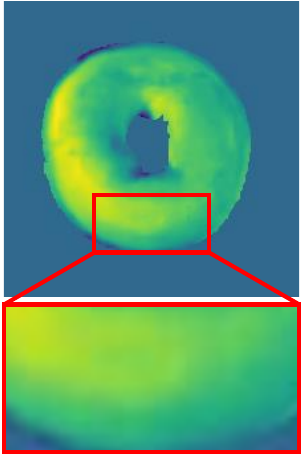} &
        \includegraphics[width=0.2\linewidth]{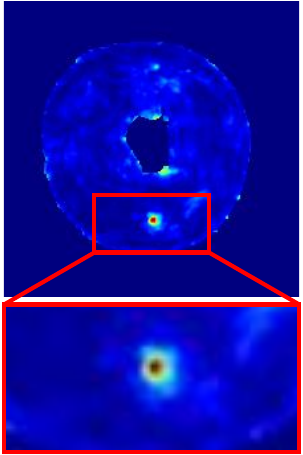} &
        \includegraphics[width=0.2\linewidth]{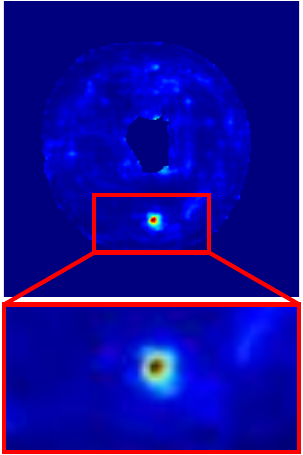} \\
    \end{tabular}}
    \caption{\textbf{Features Visualization}. Channels average of feature maps before and after crossmodal mapping (\textit{viridis} colormap). 
    }
    \label{fig:feature_vis}
\end{figure}

    \noindent
    \textbf{Features Visualization.}
        In ~\cref{fig:feature_vis} we show the spatially aligned 2D and 3D feature maps before ($E_{2D}, E_{3D}$) and after ($E_{3D\rightarrow2D}, E_{2D\rightarrow3D}$) crossmodal mappings, as well as the 2D ($\Psi_{2D}$), 3D ($\Psi_{3D}$) and final anomaly maps ($\Psi_{2D} \cdot \Psi_{3D}$), for a nominal (top) and an anomalous (bottom) test sample of MVTec 3D-AD. 
        Comparing the two rows, we note that while $E_{2D}$ feature maps of the nominal and anomalous samples look similar, $E_{3D}$ better highlights the \emph{hole} anomaly. 
        We can relate this to a 3D-only anomaly depicted in~\cref{fig:sketch}. 
        After mapping, the hole is visible in $E_{3D \rightarrow 2D}$ yet not in $E_{2D \rightarrow 3D}$. 
        This visualization agrees with the rationale discussed in \cref{sec:crossmodal_mapping} and \cref{fig:sketch}: nominal 2D features are mapped into nominal 3D features, anomalous 3D features into anomalous 2D features. 
        Thus, by computing the discrepancy between mapped and original features, we obtain 2D and 3D anomaly maps, both highlighting the defect. 
        Remarkably, though the hole occupies a very small portion of the image, it is detected accurately.
        
    \noindent
    \textbf{Crossmodal Mapping vs Intramodal  Reconstruction.}
        The authors of DFR~\cite{yang2020dfr} argued that learning a reconstruction network in feature space from nominal samples makes it possible to detect anomalies in RGB images by analyzing the reconstruction error. 
        As our method may be thought of as performing a \emph{Crossmodal} reconstruction in feature space, we investigate the impact of learning Crossmodal vs. Intra-modal feature mapping functions.  
        \cref{tab:ftn_crossmodality_ablation} compares the results of our approach (Cross) to those obtained by modifying the input layers of both our mapping networks so as to learn to reconstruct features within the same modality (Intra).  
        The results obtained by reconstructing each modality independently show that our proposed crossmodal feature mapping sets forth a more effective modality-specific learning objective w.r.t. intra-modal feature reconstruction (rows 1~vs.~4, 2~vs.~5). 
        This yields better results also by the aggregated maps obtained by pixel-wise product (rows 3~vs.~6).
        

    
\section{Conclusions and Limitations}
\label{sec:conclusion}
    We have developed an effective and efficient multimodal AD framework based on the core idea of mapping features extracted by Transformer architectures across modalities. 
    This novel paradigm outperforms previous resource-intensive methods on the MVTec 3D-AD benchmark while delivering substantially faster inference speed.
    Additionally, we have proposed a layer-pruning strategy for frozen Transformer encoders that can vastly reduce the memory footprint and yield even faster inference without compromising  AD performance.
    Lastly, we outperform competitors in the challenging few-shot scenario, achieving state-of-the-art performance on the proposed multimodal few-shot AD benchmark.
    A limitation of our approach lies in its  \emph{multimodal-only} nature, \ie, our paradigm cannot be applied to 2D AD or 3D AD, as it mandates data from both modalities at training and test times.
    %
    
    \noindent
    \textbf{Acknowledgements} We gratefully acknowledge the support of SACMI Imola.    
{
    \clearpage
    \small
    \bibliographystyle{ieeenat_fullname}
    \bibliography{main}
}

\onecolumn
\setcounter{section}{0}
\setcounter{figure}{0}
\renewcommand{\thesection}{\Alph{section}}
    
    \section*{Overview}
    This supplementary material includes additional experimental results.  
    In particular, we report:
    \begin{itemize}
        \item A more detailed analysis on the dynamic of the PRO (Per-Region Overlap) curve, alongside comparisons dealing with different integration thresholds;
         \item An ablation study concerning the architecture of the Feature Mapping networks, \ie the core components in our method;
         \item An ablation study regarding the backbone employed as 2D Feature Extractor;
         \item Additional quantitative and qualitative results dealing with both  MVTec 3D-AD and Eyecandies.
         
    \end{itemize}

 \section{Analysis of the PRO curve}
 \label{sec:an_pro_curve}
\begin{figure}[h]
        \centering
        \begin{minipage}{0.45\textwidth}
            \centering
            \includegraphics[width=\linewidth]{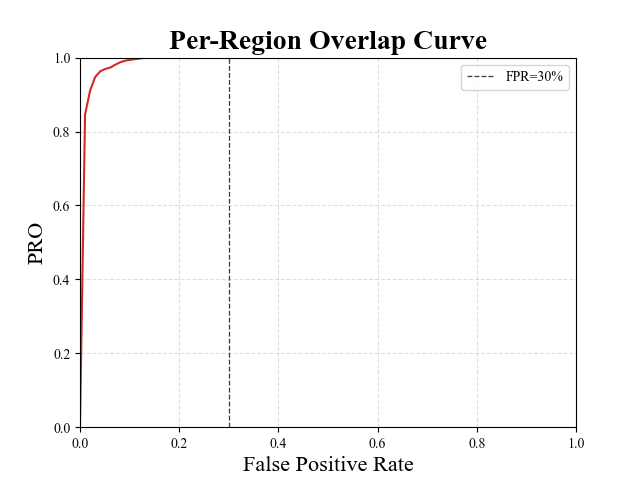}    
            \caption{
            \textbf{PRO curve - Whole FPR Range.}
             Per-Region Overlap curve obtained by our method on class \emph{Foam} of MVTec 3D-AD. The dotted line shows the AUPRO@30\% threshold.
            }
            \label{fig:pro_curve}
        \end{minipage}
        \hspace{0.5cm}
        \begin{minipage}{0.45\textwidth}
            \centering
                \includegraphics[width=\linewidth]{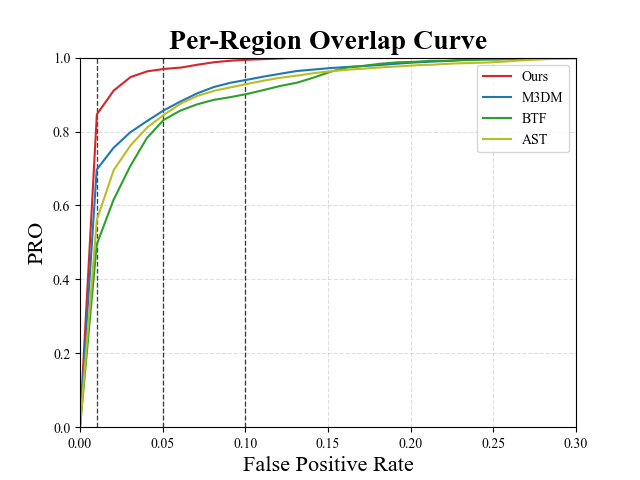}
                
            \caption{
            \textbf{PRO curve - Lower FPRs.}
             Per-Region Overlap curve obtained by all Multmodal AD methods on class \emph{Foam} of MVTec 3D-AD. Focus on the [0-0.3] FPR range.
            }
            \label{fig:pro_curve_bis}
        \end{minipage}
        \end{figure}

     The chart in \cref{fig:pro_curve} reports the Per-Region Overlap curve provided by our method on class \emph{Foam} of the MVTec 3D-AD dataset. 
     The chart shows how most of the dynamic of the curve is concentrated way underneath the $0.3$ integration threshold used to define the popular  AUPRO@30\% metric. This is also highlighted in \cref{fig:pro_curve_bis}, which compares the different Multimodal AD methods focusing on lower FPRs. 

\begin{figure*}[t]
        \centering
            \includegraphics[width=0.32\linewidth]{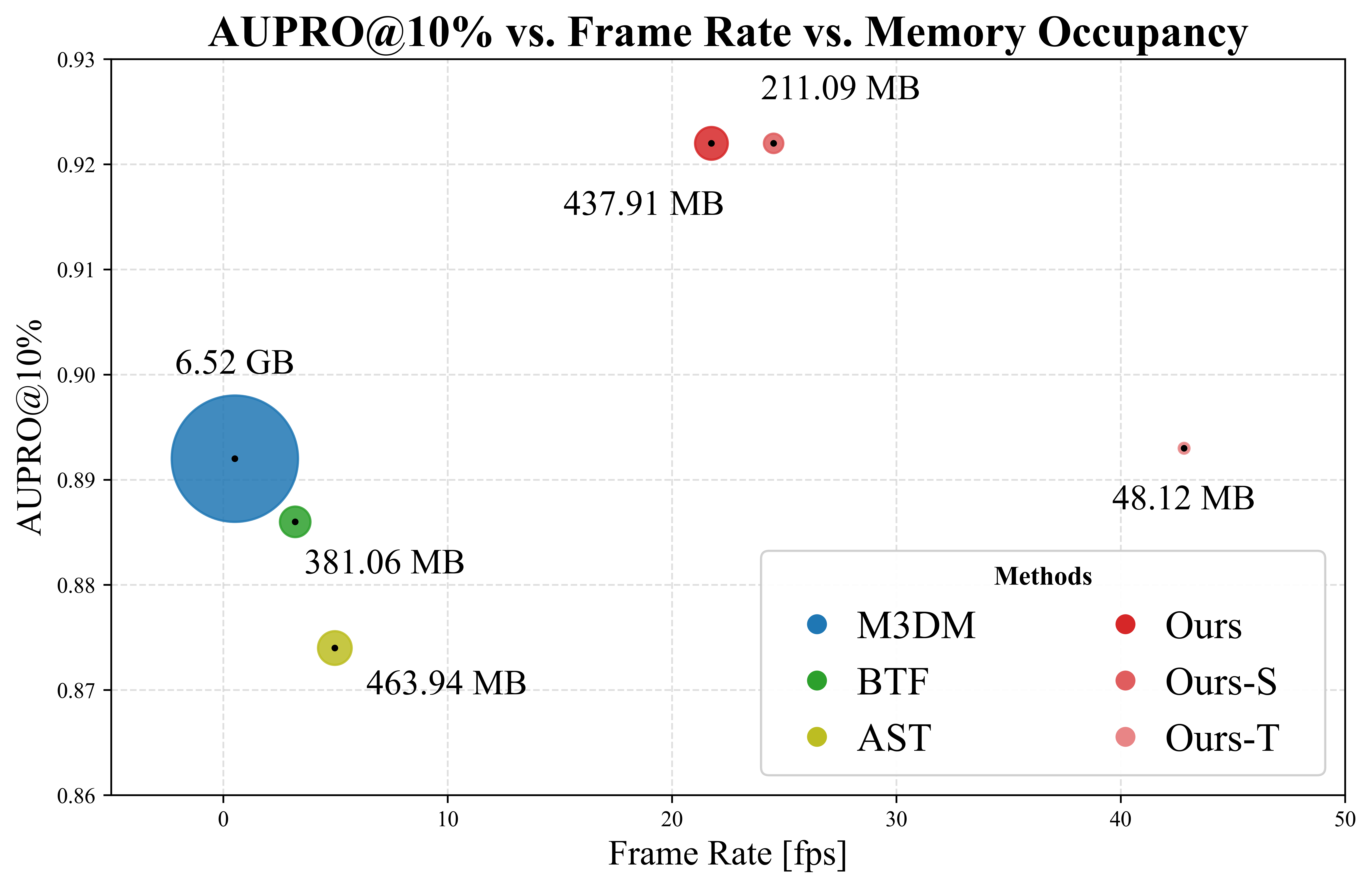}
            \hfill
            \includegraphics[width=0.32\linewidth]{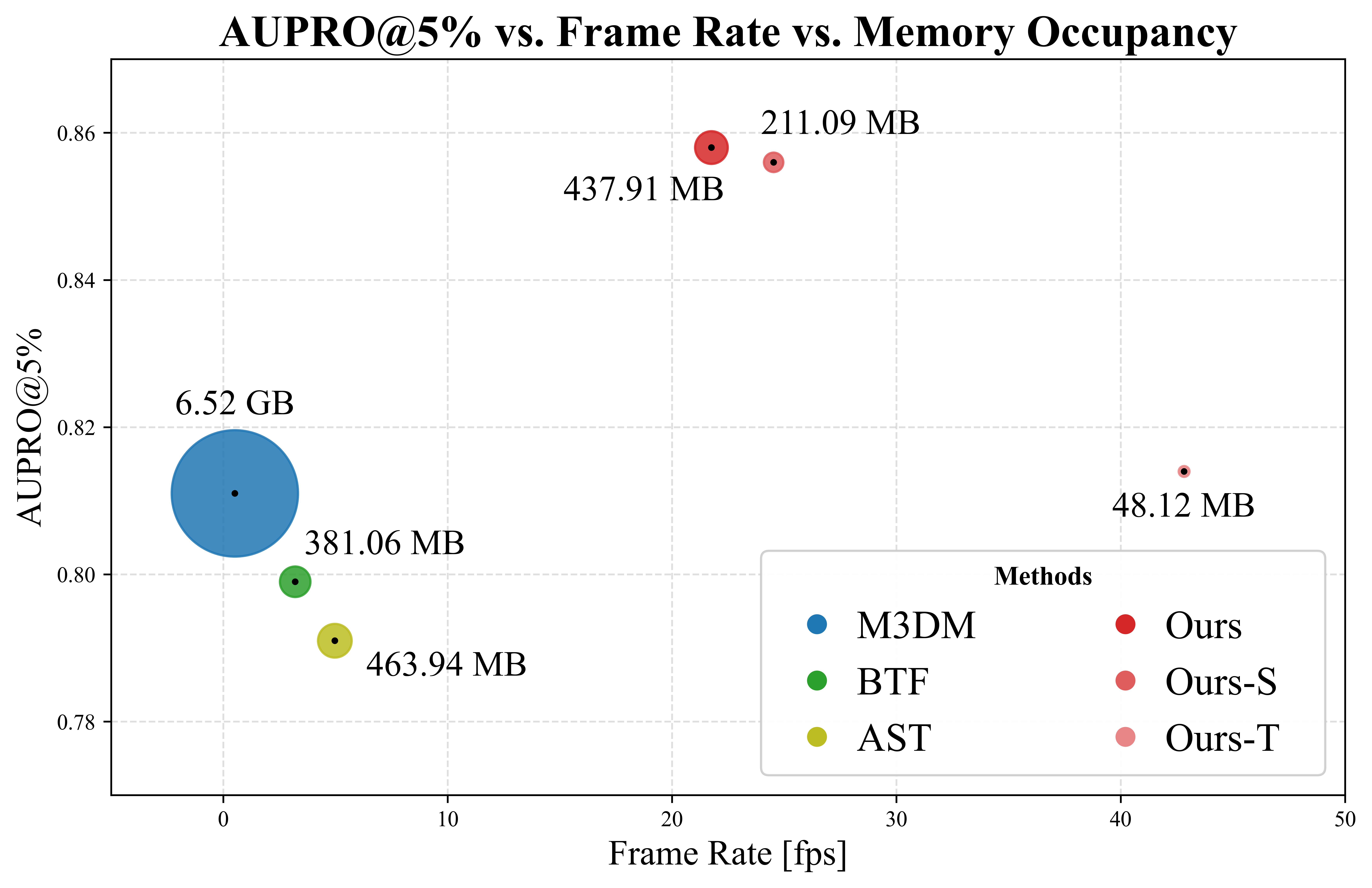}
            \hfill
            \includegraphics[width=0.32\linewidth]{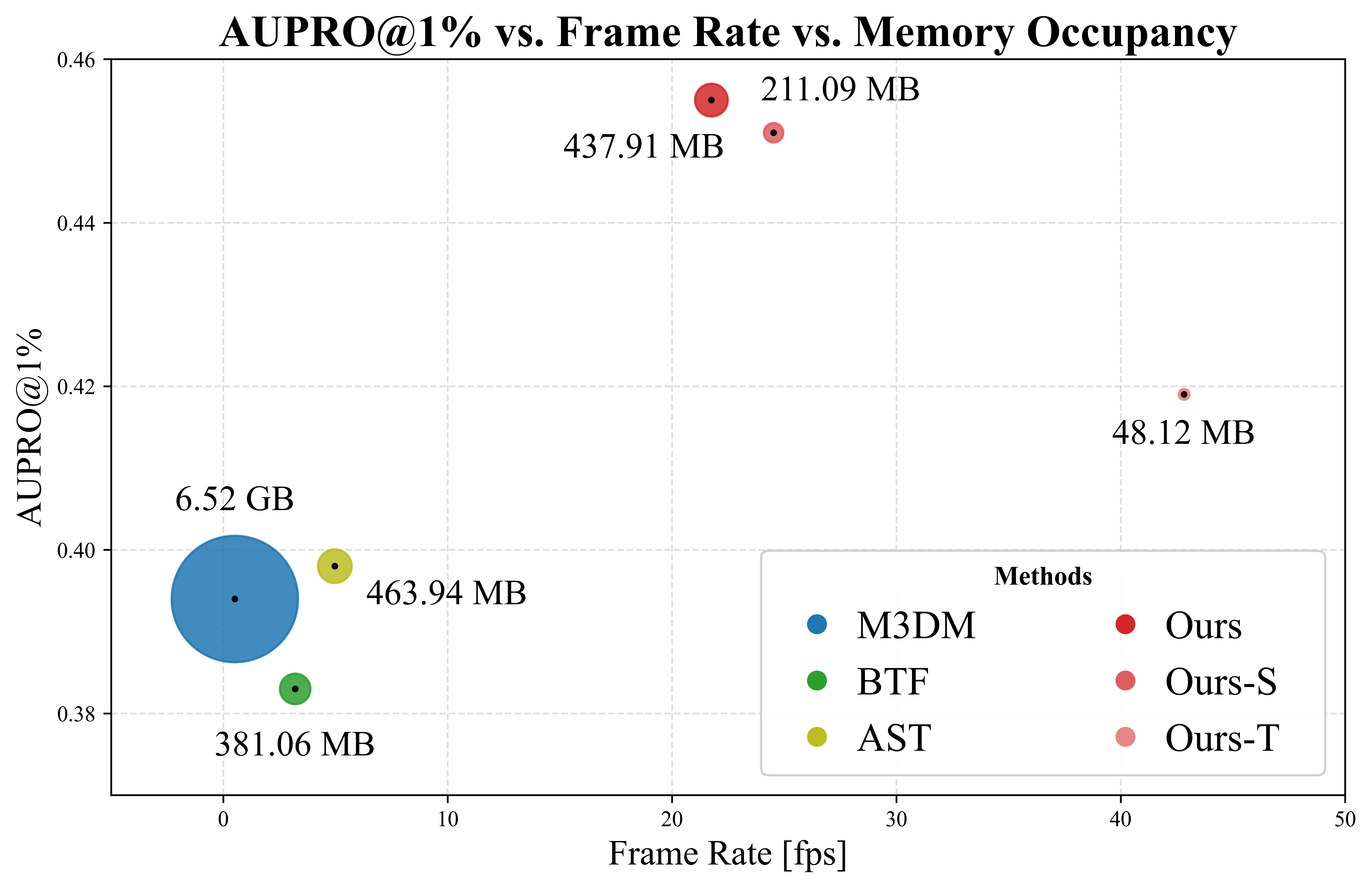}
            
        \caption{\textbf{Performance, speed and memory occupancy of Multimodal Anomaly Detection methods.}
        The chart reports anomaly segmentation performance on MVTec 3D-AD according to different AUPRO variants (from left to right: AUPRO@10\%, AUPRO@5\%, AUPRO@1\% )  vs. inference speed (Frame Rate on an NVIDIA 4090 GPU). 
        The size of the symbols is proportional to memory occupancy at inference time.         }
        \label{fig:teaser_suppl}
    \end{figure*}

    Thus, as discussed in the main paper, on one hand choosing FPR=0.3  as integration threshold may not match the requirements of a number of  industrial applications, on the other, it tends to wash out the performance differences between the methods, which, indeed, behave much more differently at lower, \ie, more challenging FPRs. 
    Hence, we deem it worth considering also more demanding variants of the AUPRO metric, such as, in particular, those obtained with integration thresholds 0.1, 0.05, and 0.01, referred to as AUPRO@10\%, AUPRO@5\% and AUPRO@1\%, respectively.
    As illustrated in \cref{fig:teaser_suppl}, our proposal consistently provides better performance (\ie, higher AUPRO) than previous Multimodal AD methods across all the considered variants of the AUPRO metric while running much faster and requiring way less memory. 
    In particular, the performance gap is higher for the more challenging variants of the AUPRO.

    
    \section{Feature Mapping Networks}
        We investigate the use of alternative network architectures to implement the Feature Mapping functions, namely: (i) MLP Encoder-Decoder,  (ii) MLP Projection, \ie the architecture described in the main paper, and (iii) Convolutional Encoder-Decoder. 
        
        The MLP Encoder-Decoder architecture comprises an encoding stage and a decoding stage, each consisting of two layers, along with an extra bottleneck layer between these two stages.
        The input layer in the encoding stage has a number of neurons equal to the dimensionality of the input feature space, while the last layer in the decoding stage has a number of neurons equal to the dimensionality of the output feature space.
        Between each pair of successive layers, but for the bottleneck layer, the number of neurons is either halved (in the encoding stage) or doubled (in the decoding stage).
        Accordingly, in our setup, we have $[768, 384, 192, 192, 384, 1152]$ neurons in each layer for $\mathcal{M}_{2D \rightarrow 3D}$, and $[1152, 576, 288, 288, 576, 768]$ neurons in each layer for $\mathcal{M}_{3D \rightarrow 2D}$.
        In both networks, all but the last layer employ GeLU activations.
        \begin{table*}[b]
    \centering
    \resizebox{0.65\textwidth}{!}{%
        \begin{tabular}{c||cccccccccc||c}

            \textbf{Metric} &
            \textit{Bagel} &
            \textit{Cable Gland} &
            \textit{Carrot} &
            \textit{Cookie} &
            \textit{Dowel} &
            \textit{Foam} &
            \textit{Peach} &
            \textit{Potato} &
            \textit{Rope} &
            \textit{Tire} &
            \textbf{Mean} \\ 

            \hline 

            & \multicolumn{10}{c}{\textbf{MLP Encoder-Decoder}} & \\

            \hline

            I-AUROC & \underline{0.993} & 0.858 & \textbf{0.992} & 0.988 & \underline{0.985} & 0.911 & \underline{0.959} & 0.866 & \underline{0.986} & 0.864 & 0.940 \\

            \hline
            
            AUPRO@30\% & \textbf{0.979} & 0.959 & \textbf{0.982} & \underline{0.940} & \underline{0.946} & 0.960 & \underline{0.980} & \underline{0.982} & \underline{0.972} & \textbf{0.981} & \underline{0.968} \\
            AUPRO@10\% & \textbf{0.938} & 0.882 & \underline{0.946} & \underline{0.890} & \underline{0.843} & 0.883 & \underline{0.941} & \underline{0.946} & \underline{0.918} & \underline{0.942} & 0.913 \\
            AUPRO@5\% & \underline{0.879} & 0.791 & \underline{0.893} & \underline{0.830} & \underline{0.749} & 0.797 & \underline{0.883} & 0.892 & \underline{0.853} & \underline{0.884} & 0.845 \\
            AUPRO@1\% & \underline{0.467} & 0.385 & \underline{0.487} & \textbf{0.455} & \underline{0.385} & 0.395 & \underline{0.466} & 0.480 & 0.451 & \underline{0.466} & 0.444 \\
            
            \hline

            Frame Rate (fps) &  &  &  &  &  &  &  &  &  &  & \textbf{25.769} \\
            Memory (MB)      &  &  &  &  &  &  &  &  &  &  & \textbf{369.856} \\
            
            \hline

            & \multicolumn{10}{c}{\textbf{MLP Projection (main paper)}} & \\

            \hline
            
            I-AUROC & 0.990 & \textbf{0.894} & 0.986 & \underline{0.989} & 0.980 & \underline{0.916} & 0.951 & \textbf{0.916} & \underline{0.986} & \underline{0.886} & \underline{0.949} \\

            \hline
            
            AUPRO@30\% & \textbf{0.979} & \underline{0.963} & \textbf{0.982} & \underline{0.940} & 0.944 & \underline{0.961} & \underline{0.980} & \textbf{0.983} & \underline{0.972} & \underline{0.980} & \underline{0.968} \\
            AUPRO@10\% & \underline{0.937} & \underline{0.892} & \textbf{0.947} & \underline{0.890} & 0.838 & \underline{0.885} & 0.940 & \textbf{0.948} & \underline{0.918} & 0.941 & \underline{0.914} \\
            AUPRO@5\% & 0.878 & \underline{0.806} & \textbf{0.894} & \underline{0.830} & 0.742 & \underline{0.799} & 0.882 & \textbf{0.897} & \underline{0.853} & 0.882 & \underline{0.846} \\
            AUPRO@1\% & \textbf{0.469} & \underline{0.402} & 0.486 & 0.450 & 0.380 & \underline{0.397} & 0.463 & \textbf{0.490} & \underline{0.453} & 0.463 & \underline{0.445} \\

            \hline

            Frame Rate (fps) &  &  &  &  &  &  &  &  &  &  & \underline{21.755} \\
            Memory (MB)      &  &  &  &  &  &  &  &  &  &  & \underline{437.911} \\

            \hline 
            
            & \multicolumn{10}{c}{\textbf{Convolutional Encoder-Decoder}} & \\

            \hline

            I-AUROC & \textbf{0.997} & \underline{0.866} & \underline{0.990} & \textbf{0.993} & \textbf{0.989} & \textbf{0.927} & \textbf{0.979} & \underline{0.897} & \textbf{0.990} & \textbf{0.918} & \textbf{0.955} \\

            \hline
            
            AUPRO@30\% & \textbf{0.979} & \textbf{0.965} & \textbf{0.982} & \textbf{0.941} & \textbf{0.948} & \underline{0.969} & \textbf{0.982} & \textbf{0.983} & \textbf{0.977} & \textbf{0.981} & \textbf{0.971} \\
            AUPRO@10\% & \textbf{0.938} & \textbf{0.897} & \textbf{0.947} & \textbf{0.893} & \textbf{0.847} & \textbf{0.906} & \textbf{0.945} & \textbf{0.948} & \textbf{0.931} & \textbf{0.944} & \textbf{0.920} \\
            AUPRO@5\% & \textbf{0.880} & \textbf{0.813} & \textbf{0.894} & \textbf{0.834} & \textbf{0.756} & \textbf{0.820} & \textbf{0.891} & \underline{0.896} & \textbf{0.872} & \textbf{0.889} & \textbf{0.855} \\
            AUPRO@1\% & \textbf{0.469} & \textbf{0.409} & \textbf{0.488} & \underline{0.453} & \textbf{0.393} & \textbf{0.409} & \textbf{0.477} & \underline{0.488} & \textbf{0.467} & \textbf{0.473} & \textbf{0.453} \\

            \hline

            Frame Rate (fps) &  &  &  &  &  &  &  &  &  &  & 9.906 \\
            Memory (MB)      &  &  &  &  &  &  &  &  &  &  & 2780.690 \\

            \hline 
            
        \end{tabular}}
    \caption{Results on MVTec 3D-AD, Models trained for 50 epochs. Best results in \textbf{bold}, runner-ups \underline{underlined}.}
    \label{tab:ftn_architecture_ablation}
\end{table*}
        \begin{table*}[b]
    \centering
    \resizebox{0.65\textwidth}{!}{%
        \begin{tabular}{c||cccccccccc||c}

            \textbf{Metric} &
            \textit{Bagel} &
            \textit{Cable Gland} &
            \textit{Carrot} &
            \textit{Cookie} &
            \textit{Dowel} &
            \textit{Foam} &
            \textit{Peach} &
            \textit{Potato} &
            \textit{Rope} &
            \textit{Tire} &
            \textbf{Mean} \\ 

            \hline

            & \multicolumn{10}{c}{$\Psi_{2D}$} & \\

            \hline
            
            I-AUROC & 0.937 & 0.864 & \underline{0.984} & 0.951 & \textbf{0.984} & 0.789 & 0.915 & 0.736 & \underline{0.968} & 0.825 & 0.895 \\

            \hline
            
            AUPRO@30\% & 0.960 & 0.966 & 0.979 & 0.884 & 0.911 & 0.916 & \underline{0.981} & 0.974 & 0.958 & 0.971 & 0.950 \\
            AUPRO@10\% & 0.896 & 0.906 & 0.937 & 0.813 & 0.741 & 0.783 & \underline{0.942} & 0.922 & 0.878 & 0.913 & 0.873 \\
            AUPRO@5\% & 0.819 & 0.834 & 0.874 & 0.738 & 0.624 & 0.675 & \underline{0.884} & 0.844 & 0.789 & 0.841 & 0.792 \\
            AUPRO@1\% & 0.410 & 0.427 & 0.456 & 0.371 & 0.311 & 0.326 & \underline{0.468} & 0.410 & 0.401 & 0.429 & 0.401 \\
            
            \hline

            & \multicolumn{10}{c}{$\Psi_{3D}$} & \\

            \hline
            
            I-AUROC & 0.948 & 0.770 & 0.968 & 0.981 & 0.937 & \textbf{0.893} & 0.694 & \underline{0.909} & 0.939 & 0.812 & 0.885 \\

            \hline
            
            AUPRO@30\% & 0.967 & 0.922 & \underline{0.981} & \underline{0.926} & \underline{0.919} & \underline{0.965} & 0.965 & \underline{0.981} & \underline{0.963} & 0.976 & 0.956 \\
            AUPRO@10\% & 0.903 & 0.782 & \underline{0.943} & \underline{0.871} & \underline{0.764} & \underline{0.899} & 0.894 & \underline{0.943} & \underline{0.892} & 0.928 & 0.882 \\
            AUPRO@5\% & 0.817 & 0.664 & \underline{0.887} & \underline{0.806} & \underline{0.661} & \underline{0.812} & 0.793 & \underline{0.887} & \underline{0.818} & 0.858 & 0.800 \\
            AUPRO@1\% & 0.402 & 0.302 & \underline{0.474} & 0.443 & \underline{0.341} & \underline{0.389} & 0.338 & \underline{0.474} & \underline{0.431} & 0.437 & 0.403 \\

            \hline

            & \multicolumn{10}{c}{$\Psi_{2D} + \Psi_{3D}$} & \\

            \hline
            
            I-AUROC   & \underline{0.980} & \textbf{0.893} & \textbf{0.991} & \textbf{0.996} & 0.980 & 0.844 & \textbf{0.970} & 0.876 & 0.966 & \underline{0.894} & \underline{0.939} \\
            
            \hline
            
            AUPRO@30\% & \underline{0.969} & \underline{0.968} & 0.980 & 0.904 & 0.914 & 0.958 & \textbf{0.982} & 0.977 & 0.961 & \underline{0.977} & \underline{0.959} \\
            AUPRO@10\% & \underline{0.917} & \underline{0.912} & 0.941 & 0.853 & 0.749 & 0.877 & \textbf{0.945} & 0.932 & 0.886 & \underline{0.931} & \underline{0.894} \\
            AUPRO@5\%  & \underline{0.852} & \textbf{0.844} & 0.882 & 0.799 & 0.638 & 0.784 & \textbf{0.890} & 0.864 & 0.806 & \underline{0.869} & \underline{0.823} \\
            AUPRO@1\%  & \underline{0.448} & \textbf{0.439} & 0.468 & \underline{0.462} & 0.323 & 0.384 & \textbf{0.478} & 0.439 & 0.424 & \underline{0.456} & \underline{0.432} \\

            \hline
            
            & \multicolumn{10}{c}{$\max(\Psi_{2D}, \Psi_{3D}$)} & \\
            
            \hline
            
            I-AUROC   & 0.937 & 0.865 & \underline{0.984} & 0.951 & \underline{0.983} & 0.789 & 0.915 & 0.736 & \underline{0.968} & 0.825 & 0.895 \\

            \hline
            
            AUPRO@30\% & 0.960 & 0.966 & 0.979 & 0.884 & 0.911 & 0.916 & \underline{0.981} & 0.974 & 0.958 & 0.971 & 0.950 \\
            AUPRO@10\% & 0.896 & 0.906 & 0.937 & 0.813 & 0.741 & 0.783 & \underline{0.942} & 0.922 & 0.878 & 0.913 & 0.873 \\
            AUPRO@5\%  & 0.819 & 0.834 & 0.874 & 0.738 & 0.624 & 0.675 & \underline{0.884} & 0.844 & 0.789 & 0.841 & 0.792 \\
            AUPRO@1\%  & 0.410 & 0.428 & 0.456 & 0.371 & 0.311 & 0.326 & \underline{0.468} & 0.410 & 0.401 & 0.429 & 0.401 \\

            \hline
            
            & \multicolumn{10}{c}{$\Psi_{2D} \cdot \Psi_{3D}$} & \\

            \hline
            
            I-AUROC   & \textbf{0.994} & \underline{0.888} & \underline{0.984} & \underline{0.993} & 0.980 & \underline{0.888} & \underline{0.941} & \textbf{0.943} & \textbf{0.980} & \textbf{0.953} & \textbf{0.954} \\

            \hline
            
            AUPRO@30\% & \textbf{0.979} & \textbf{0.972} & \textbf{0.982} & \textbf{0.945} & \textbf{0.950} & \textbf{0.968} & 0.980 & \textbf{0.982} & \textbf{0.975} & \textbf{0.981} & \textbf{0.971} \\
            AUPRO@10\% & \textbf{0.937} & \textbf{0.917} & \textbf{0.947} & \textbf{0.897} & \textbf{0.855} & \textbf{0.906} & \underline{0.942} & \textbf{0.947} & \textbf{0.926} & \textbf{0.944} & \textbf{0.922} \\
            AUPRO@5\%  & \textbf{0.877} & \underline{0.843} & \textbf{0.894} & \textbf{0.840} & \textbf{0.765} & \textbf{0.828} & \underline{0.884} & \textbf{0.894} & \textbf{0.865} & \textbf{0.889} & \textbf{0.858} \\
            AUPRO@1\%  & \textbf{0.459} & \underline{0.431} & \textbf{0.485} & \textbf{0.469} & \textbf{0.394} & \textbf{0.413} & \underline{0.468} & \textbf{0.487} & \textbf{0.464} & \textbf{0.476} & \textbf{0.455} \\
            
            \hline
            
        \end{tabular}}
    \caption{\textbf{Aggregation analysis.} Best results in \textbf{bold}, runner-ups \underline{underlined}.}
    \label{tab:ftn_modality_ablation}
\end{table*}


        As to MLP Projection architecture, we refer to shallow MLPs consisting of three layers, with  GeLU activations but in the last one.  The input layer has a number of neurons equal to the dimensionality of the input feature space, while the last layer has a number of neurons equal to the dimensionality of the output feature space.
        The intermediate layer has a number of neurons equal to the mean between the dimensionality of the input and output features.
        Thus, as also reported in the main paper, in our setup the three layers in $\mathcal{M}_{2D \rightarrow 3D}$ have $768$, $960$ and $1152$ neurons each, while the three layers of $\mathcal{M}_{3D \rightarrow 2D}$ have $1152$, $960$ and $768$ neurons each.

        Finally, unlike the previous two architectures which ingest individual feature vectors, 
        the Convolutional Encoder-Decoder receives input tensors of spatial size  $H \times W$ (with  $D_{2D}$ and $D_{3D}$ channels for $\mathcal{M}_{2D \rightarrow 3D}$ and $\mathcal{M}_{3D \rightarrow 2D}$, respectively). The architecture follows a UNet-like structure without skip-connections, with two 3x3 convolutional layers followed by 2x2 max-pooling in the encoder stage  and  one 3x3 conv followed by a 2x2 transpose convolution in the decoding stage. All layers except the last one employ ReLU activations. The number of channels is kept equal to the input one up to the last layer, where it is modified so as to match the dimensionality of output feature space (\ie from  $D_{2D}$ and $D_{3D}$ for $\mathcal{M}_{2D \rightarrow 3D}$  and from $D_{3D}$ and $D_{2D}$ for $\mathcal{M}_{3D \rightarrow 2D}$.  


        For this new set of experiments, we follow the same training protocol as defined in the main paper.
        The results on  MVTec 3D-AD are reported in \cref{tab:ftn_architecture_ablation}, and show that the Convolutional Encoder-Decoder architecture provides slightly superior performance. However, despite its enhanced performance, it operates at a significantly slower inference rate, namely $9.906$ fps, in contrast to the $21.755$ fps achieved by our base model which is based on the MLP Projection architecture. Furthermore, the Convolutional Architecture requires six times more memory compared to our base model, \eg, $2780.690$ MB compared to $437.911$ MB. Thus, we are led to prefer the performance vs efficiency (both speed and memory) trade-off provided by the  MLP Projection architecture. 

        \section{Feature Extractors}
            The ever-increasing availability of frozen Transformer-based RGB feature extractors trained on large data corpora has motivated us to explore alternatives to DINO ViT-B/8, such as, in particular, the ViT-B/16 used in SAM~\cite{kirillov2023segany}, the ViT-B/16 used in CLIP~\cite{radford2021learning}, and the ViT-B/14 used in DINO-v2~\cite{oquab2023dinov2}. 
            Results obtained on MVTec 3D-AD with the different 2D Feature Extractors are reported in \cref{tab:ftn_backbone_ablation}. Interestingly, DINO and DINO-v2 exhibit much better performance than other feature extractors, which hints at - and may foster further investigation on - the benefits of foundation models trained via self-supervised contrastive learning in industrial AD.
    
        \begin{table}[t]
    \centering
    \resizebox{0.5\linewidth}{!}{%
        \begin{tabular}{c||ccccc}
            
            \textbf{$\mathcal{F}_{2D}$} & 
            \textbf{I-AUROC} & 
            \textbf{P-AUROC} & 
            \textbf{AUPRO@30\%} & 
            \textbf{AUPRO@1\%} \\
            
            \hline
            
            DINO~\cite{caron2021emerging}   & 0.949 & \textbf{0.992} & \textbf{0.968} & \textbf{0.445}\\
            SAM~\cite{kirillov2023segany}   & 0.792 & 0.973 & 0.906 &  0.311\\
            CLIP~\cite{radford2021learning} & 0.833 & 0.984 & 0.942 & 0.346\\
            DINO-v2~\cite{oquab2023dinov2}  & \textbf{0.958} & \textbf{0.992} & 0.964 & 0.437 \\
            
            \hline     
        \end{tabular}}
    \caption{\textbf{2D Feature Extractor Alternatives.} Results on MVTec 3D-AD. Best results in \textbf{bold}. Networks are trained for 50 epochs.}
    \label{tab:ftn_backbone_ablation}
\end{table}

        \section{Additional Quantitative Results}
        
        
            In this section, we report the class-wise anomaly detection and segmentation results for some of the experiments discussed in the main paper,  considering also the additional FPR thresholds to compute the AUPRO introduced in \cref{sec:an_pro_curve}.
            \begin{table*}[!h]
    \centering
    \resizebox{0.65\textwidth}{!}{%
        \begin{tabular}{c||cccccccccc||c}

            \textbf{Metric} &
            \textit{Bagel} &
            \textit{Cable Gland} &
            \textit{Carrot} &
            \textit{Cookie} &
            \textit{Dowel} &
            \textit{Foam} &
            \textit{Peach} &
            \textit{Potato} &
            \textit{Rope} &
            \textit{Tire} &
            \textbf{Mean} \\ 

            \hline

            & \multicolumn{10}{c}{\textbf{Ours}} & \\

            \hline

            I-AUROC & \textbf{0.994} & \textbf{0.888} & \textbf{0.984} & \textbf{0.993} & 0.980 & 0.888 & \underline{0.941} & \underline{0.943} & 0.980 & \textbf{0.953} & \underline{0.954} \\

            \hline
            
            AUPRO@30\% & \underline{0.979} & \textbf{0.972} & \textbf{0.982} & 0.945 & 0.950 & \underline{0.968} & \underline{0.980} & \underline{0.982} & \underline{0.975} & \underline{0.981} & \underline{0.971} \\
            AUPRO@10\% & \underline{0.937} & \textbf{0.917} & \textbf{0.947} & 0.897 & 0.855 & \underline{0.906} & \underline{0.942} & \underline{0.947} & 0.926 & \underline{0.944} & \underline{0.922} \\
            AUPRO@5\% & \underline{0.877} & \textbf{0.843} & \underline{0.894} & 0.840 & 0.765 & \underline{0.828} & \underline{0.884} & 0.894 & 0.865 & \underline{0.889} & \underline{0.858} \\
            AUPRO@1\% & 0.459 & \textbf{0.431} & 0.485 & 0.469 & 0.394 & \underline{0.413} & \underline{0.468} & 0.487 & 0.464 & \underline{0.476} & \underline{0.455} \\
               
            \hline

            & \multicolumn{10}{c}{\textbf{Ours-M}} & \\

            \hline
            
            I-AUROC & 0.988 & 0.875 & \textbf{0.984} & \underline{0.992} & \textbf{0.997} & \textbf{0.924} & \textbf{0.964} & \textbf{0.949} & 0.979 & \underline{0.950} & \textbf{0.960} \\            

            \hline
            
            AUPRO@30\% & \textbf{0.980} & \underline{0.966} & \textbf{0.982} & \underline{0.947} & \underline{0.959} & 0.967 & \textbf{0.982} & \textbf{0.983} & \textbf{0.976} & \textbf{0.982} & \textbf{0.972} \\
            AUPRO@10\% & \textbf{0.941} & \underline{0.901} & \textbf{0.947} & 0.899 & \underline{0.880} & 0.901 & \textbf{0.945} & \textbf{0.949} & \textbf{0.930} & \textbf{0.947} & \textbf{0.924} \\
            AUPRO@5\% & \textbf{0.884} & \underline{0.817} & \textbf{0.895} & 0.842 & \underline{0.798} & 0.823 & \textbf{0.890} & \textbf{0.898} & \textbf{0.872} & \textbf{0.893} & \textbf{0.861} \\
            AUPRO@1\% & \textbf{0.480} & \underline{0.398} & \underline{0.490} & 0.467 & \textbf{0.413} & 0.408 & \textbf{0.481} & \textbf{0.494} & \textbf{0.468} & \textbf{0.488} & \textbf{0.459} \\

            \hline
            
            & \multicolumn{10}{c}{\textbf{Ours-S}} & \\

            \hline

            I-AUROC & 0.983 & \underline{0.878} & \underline{0.973} & \underline{0.992} & \underline{0.987} & \underline{0.913} & 0.900 & 0.936 & \underline{0.981} & 0.941 & 0.948 \\

            \hline
            
            AUPRO@30\% & 0.978 & 0.960 & \textbf{0.982} & \textbf{0.948} & \textbf{0.960} & \textbf{0.972} & 0.977 & \textbf{0.983} & \textbf{0.976} & \underline{0.981} & \textbf{0.972} \\
            AUPRO@10\% & 0.936 & 0.882 & \textbf{0.947} & \underline{0.900} & \textbf{0.884} & \textbf{0.918} & 0.932 & \textbf{0.949} & \underline{0.929} & 0.943 & \underline{0.922} \\
            AUPRO@5\% & 0.874 & 0.782 & \underline{0.894} & \underline{0.843} & \textbf{0.800} & \textbf{0.845} & 0.864 & \textbf{0.898} & \underline{0.870} & 0.886 & 0.856 \\
            AUPRO@1\% & \underline{0.461} & 0.379 & \textbf{0.492} & \underline{0.479} & \underline{0.411} & \textbf{0.429} & 0.430 & \textbf{0.494} & \underline{0.467} & 0.472 & 0.451 \\            

            \hline

            & \multicolumn{10}{c}{\textbf{Ours-T}} & \\

            \hline

            I-AUROC & 0.948 & 0.784 & 0.946 & 0.985 & 0.946 & 0.855 & 0.815 & 0.932 & \textbf{0.989} & 0.794 & 0.899 \\

            \hline
            
            AUPRO@30\% & 0.977 & 0.903 & \underline{0.981} & 0.950 & 0.945 & 0.956 & 0.973 & \textbf{0.983} & 0.973 & 0.973 & 0.961 \\
            AUPRO@10\% & 0.932 & 0.736 & \underline{0.944} & \textbf{0.901} & 0.838 & 0.873 & 0.919 & \textbf{0.949} & 0.920 & 0.918 & 0.893 \\
            AUPRO@5\% & 0.867 & 0.612 & 0.889 & \textbf{0.844} & 0.729 & 0.773 & 0.839 & \underline{0.897} & 0.856 & 0.838 & 0.814 \\
            AUPRO@1\% & 0.449 & 0.267 & 0.487 & \textbf{0.487} & 0.364 & 0.369 & 0.395 & \underline{0.491} & 0.462 & 0.421 & 0.419 \\
            
            \hline
            
        \end{tabular}
        }
    \caption{\textbf{Layers Pruning analysis.} Best results in \textbf{bold}, runner-ups \underline{underlined}.}
    \label{tab:ftn_layers_ablation}
\end{table*}

            In particular, \cref{tab:ftn_modality_ablation} provides a detailed view of the results for the \emph{Aggregation} function introduced in Sec.~3.3 of the main paper. 
            As already highlighted in the evaluation summarized in Tab.~6 and discussed in Sec.~5 of the main paper, the product aggregation achieves the best results across most of the classes except for one class, \ie, \emph{Peach}, which shows higher results using the sum aggregation. 
            These results further support our choice of relying on the product function, which realizes a logical AND between the discrepancies found in the individual modalities, as preferred aggregation approach. 

            In addition, \cref{tab:ftn_layers_ablation} reports the detailed results for the \emph{Layers Pruning} technique. As described in Sec. 3.4 of the main paper, to obtain lighter versions of our framework, we prune both feature extractors after the 1st, 4th, and 8th layer to obtain \emph{Tiny}, \emph{Small}, and \emph{Medium} architectures, referred to as Ours-T, Ours-S and Ours-M.
            Thus, \cref{tab:ftn_layers_ablation}  extends the evaluation summarized in Tab.~5 and discussed in Sec.~5 of the main paper.
            It is worth noticing how Ours-M achieves the best results in both detection and segmentation.
            We also highlight that Ours obtains the second-best results in all average metrics.

            For the sake of completeness, we also report in \cref{tab:pauroc_comparisons_mvtec} the P-AUROC results on the MVTec 3D-AD dataset.
            As already anticipated in Sec.~5 of the main paper, this metric is mostly saturated since every method reaches the same very high results for each class.
            \begin{table*}[t]
    \centering
    \resizebox{0.6\textwidth}{!}{%
        \begin{tabular}{c||cccccccccc||c}

            \textbf{Method} &
            \textit{Bagel} &
            \textit{Cable Gland} &
            \textit{Carrot} &
            \textit{Cookie} &
            \textit{Dowel} &
            \textit{Foam} &
            \textit{Peach} &
            \textit{Potato} &
            \textit{Rope} &
            \textit{Tire} &
            \textbf{Mean} \\ 

            \hline

            BTF~\cite{horwitz2023back}     & 0.996 & 0.992 & 0.997 & 0.994 & 0.981 & 0.974 & 0.996 & 0.998 & 0.994 & 0.995 & 0.992 \\
            AST~\cite{RudWeh2023}          & - & - & - & - & - & - & - & - & - & - & 0.976 \\
            M3DM~\cite{wang2023multimodal} & 0.995 & 0.993 & 0.997 & 0.985 & 0.985 & 0.984 & 0.996 & 0.994 & 0.997 & 0.996 & 0.992 \\
            Ours                           & 0.997 & 0.992 & 0.999 & 0.972 & 0.987 & 0.993 & 0.998 & 0.999 & 0.998 & 0.998 & 0.993 \\
            
            \hline
            
        \end{tabular}}
    \caption{P-AUROC on MVTec 3D-AD dataset in comparison with state-of-the-art models. 
    }
    \label{tab:pauroc_comparisons_mvtec}
\end{table*}
      
            As regards the Eyecandies dataset, we provide a detailed view of the results for each class in \cref{tab:aupro_iauroc_comparisons_eyecandies}, also considering different FPR thresholds.
            It is worth highlighting that the original results provided by M3DM~\cite{wang2023multimodal} were obtained by training on a subset of the train set of Eyecandies, mostly due to the limitations caused by the memory bank resource requirements.
            To achieve more comparable results, we retrained M3DM~\cite{wang2023multimodal} on the full training set and reevaluated the benchmark, denoted as M3DM* in \cref{tab:aupro_iauroc_comparisons_eyecandies}.


            Generally, we note that features from deeper layers deliver higher contextualizations, thus enabling our cross-modal mapping to perform anomaly detection better, for the reasons highlighted in Sec. 3 of the main paper.
            However, some literature findings suggest that, in self-supervised learning, features from slightly shallower layers may turn out more task agnostic, i.e. exhibit a better ability to generalize to a wider range of downstream tasks. Thus, we argue that the above considerations may explain the slightly different performance between Ours and Ours-M in the considered datasets.
            Overall, we suggest the simplest and most general approach of keeping the whole Transformer-based feature extractors (i.e. Ours) as the default choice in our framework.
            
            \begin{table*}[!h]
    \centering
    \resizebox{0.75\textwidth}{!}{%
        \begin{tabular}{c||c||cccccccccc||c}
            &
            \textbf{Method} &
            \textit{Can. C.} &
            \textit{Cho. C.} &
            \textit{Cho. P.} &
            \textit{Conf.} &
            \textit{Gum. B.} &
            \textit{Haz. T.} &
            \textit{Lic. S.} &
            \textit{Lollip.} &
            \textit{Marsh.} &
            \textit{Pep. C.} &
            \textbf{Mean} \\

            \hline
            
            \multirow{6}{*}{\rotatebox{0}{\textbf{I-AUROC}}}

            & RGB-D~\cite{bonfiglioli2022eyecandies}    & 0.529 & 0.861 & 0.739 & 0.752 & 0.594 & 0.498 & 0.679 & 0.651 & 0.838 & 0.750 & 0.689 \\
            & RGB-cD-n~\cite{bonfiglioli2022eyecandies} & 0.596 & 0.843 & 0.819 & 0.846 & 0.833 & 0.550 & 0.750 & 0.846 & 0.940 & 0.848 & 0.787 \\
            & M3DM~\cite{wang2023multimodal}            & 0.624 & \textbf{0.958} & \textbf{0.958} & \textbf{1.000} & \textbf{0.886} & \underline{0.758} & \textbf{0.949} & 0.836 & \textbf{1.000} & \textbf{1.000} & \textbf{0.897} \\
            & M3DM*~\cite{wang2023multimodal}           & 0.597 & \underline{0.954} & 0.931 & \underline{0.990} & \underline{0.883} & 0.666 & \underline{0.923} & \underline{0.888} & 0.995 & \textbf{1.000} & \underline{0.882}\\
            & AST~\cite{RudWeh2023}                     & 0.574 & 0.747 & 0.747 & 0.889 & 0.596 & 0.617 & 0.816 & 0.841 & 0.987 & \underline{0.987} & 0.780 \\
            & Ours                                      & \textbf{0.680} & 0.931 & \underline{0.952} & 0.880 & 0.865 & \textbf{0.782} & 0.917 & 0.840 & \underline{0.998} & 0.962 & 0.881 \\
            & Ours-M                                    & \underline{0.645} & 0.936 & 0.914 & 0.901 & 0.845 & 0.747 & 0.877 & \textbf{0.904} & 0.992 & 0.885 & 0.865 \\
            
            \hline 
            
            \multirow{6}{*}{\rotatebox{0}{\textbf{P-AUROC}}} 
            & RGB-D~\cite{bonfiglioli2022eyecandies}    & 0.973 & 0.927 & 0.958 & 0.945 & 0.929 & 0.806 & 0.827 & 0.977 & 0.931 & 0.928 & 0.920 \\
            & RGB-cD-n~\cite{bonfiglioli2022eyecandies} & 0.980 & 0.979 & \textbf{0.982} & 0.978 & 0.951 & 0.853 & 0.971 & 0.978 & 0.985 & 0.967 & 0.962 \\
            & M3DM~\cite{wang2023multimodal}            & 0.974 & \textbf{0.987} & 0.962 & \textbf{0.998} & \underline{0.966} & \underline{0.941} & \underline{0.973} & \underline{0.984} & \textbf{0.996} & 0.985 & \textbf{0.977} \\
            & M3DM*~\cite{wang2023multimodal}           & 0.968 & \underline{0.986} & \underline{0.964} & \textbf{0.998} & \textbf{0.976} & 0.928 & \textbf{0.976} & \textbf{0.988} & \textbf{0.996} & \textbf{0.995} & \textbf{0.977} \\
            & AST~\cite{RudWeh2023}                     & 0.763 & 0.960 & 0.911 & 0.969 & 0.788 & 0.837 & 0.918 & 0.924 & 0.983 & 0.968 & 0.902 \\
            & Ours                                      & \underline{0.983} & 0.982 & \underline{0.964} & \underline{0.989} & 0.949 & \textbf{0.946} & 0.969 & 0.980 & \underline{0.995} & \underline{0.987} & \underline{0.974} \\
            & Ours-M                                    & \textbf{0.985} & 0.984 & 0.961 & 0.986 & 0.958 & 0.937 & 0.968 & 0.981 & 0.994 & 0.978 & 0.973 \\

            \hline

            \multirow{4}{*}{\rotatebox{0}{\textbf{AUPRO@30\%}}} 
            & M3DM~\cite{wang2023multimodal}  & 0.906 & \textbf{0.923} & 0.803 & \textbf{0.983} & 0.855 & 0.688 & \textbf{0.880} & 0.906 & \textbf{0.966} & \underline{0.955} & \underline{0.882} \\
            & M3DM*~\cite{wang2023multimodal} & 0.889 & \underline{0.921} & \underline{0.808} & \underline{0.982} & \textbf{0.889} & 0.675 & \underline{0.872} & 0.901 & \underline{0.964} & \textbf{0.973} & \textbf{0.887} \\
            & AST~\cite{RudWeh2023}           & 0.514   & 0.835   & 0.714   & 0.905 & 0.587   & 0.590 & 0.736    & 0.769 & 0.918       & 0.878                 & 0.744 \\
            & Ours                            & \underline{0.942} & 0.902 & \textbf{0.831} & 0.965 & \underline{0.875} & \underline{0.762} & 0.791 & \textbf{0.913} & 0.939 & 0.949 & \textbf{0.887} \\
            & Ours-M                          & \textbf{0.943} & 0.892 & 0.795 & 0.962 & 0.871 & \textbf{0.779} & 0.767 & \underline{0.909} & 0.944 & 0.935 & 0.880 \\

            \hline

            \multirow{3}{*}{\rotatebox{0}{\textbf{AUPRO@10\%}}} 
            & M3DM*~\cite{wang2023multimodal} & 0.677 & \textbf{0.836} & \underline{0.698} & \textbf{0.947} & \textbf{0.754} & 0.410 & \textbf{0.732} & 0.712 & \textbf{0.913} & \textbf{0.924} & 0.760 \\
            & AST~\cite{RudWeh2023}           & 0.285 & 0.709 & 0.545 & 0.770 & 0.404 & 0.350 & 0.584 & 0.544 & 0.770 & 0.744 & 0.570 \\
            & Ours                            & \underline{0.827} & \underline{0.815} & \textbf{0.731} & \underline{0.896} & 0.741 & \textbf{0.550} & 0.663 & \textbf{0.739} & 0.893 & \underline{0.868} & \textbf{0.772} \\
            & Ours-M                          & \textbf{0.829} & 0.814 & 0.683 & 0.886 & \underline{0.742} & \underline{0.564} & \underline{0.666} & \underline{0.728} & \underline{0.898} & 0.830 & \underline{0.764} \\

            \hline

            \multirow{3}{*}{\rotatebox{0}{\textbf{AUPRO@5\%}}} 
            & M3DM*~\cite{wang2023multimodal} & 0.479 & \textbf{0.759} & \underline{0.626} & \textbf{0.894} & 0.655 & 0.300 & \textbf{0.634} & \textbf{0.562} & \textbf{0.849} & \textbf{0.861} & 0.661 \\
            & AST~\cite{RudWeh2023}           & 0.173 & 0.592 & 0.421 & 0.635 & 0.288 & 0.242 & 0.461 & 0.378 & 0.634 & 0.617 & 0.444 \\
            & Ours                            & \textbf{0.662} & \underline{0.750} & \textbf{0.653} & \underline{0.801} & \textbf{0.657} & \underline{0.427} & 0.609 & \underline{0.552} & 0.838 & \underline{0.796} & \textbf{0.675} \\
            & Ours-M                          & \underline{0.661} & 0.747 & 0.611 & 0.792 & \underline{0.665} & \textbf{0.446} & \underline{0.619} & 0.518 & \underline{0.840} & 0.751 & \underline{0.665} \\

            \hline

            \multirow{3}{*}{\rotatebox{0}{\textbf{AUPRO@1\%}}} 
            & M3DM*~\cite{wang2023multimodal} & 0.166 & 0.388 & 0.329 & \textbf{0.486} & 0.315 & 0.131 & 0.323 & \textbf{0.258} & \textbf{0.462} & \textbf{0.454} & \underline{0.331} \\
            & AST~\cite{RudWeh2023}           & 0.035 & 0.230 & 0.129 & 0.234 & 0.092 & 0.069 & 0.139 & 0.090 & 0.255 & 0.224 & 0.149 \\
            & Ours                            & \textbf{0.229} & \textbf{0.397} & \textbf{0.345} & 0.389 & \textbf{0.353} & \underline{0.188} & \underline{0.333} & \underline{0.236} & \underline{0.455} & \underline{0.428} & \textbf{0.335} \\
            & Ours-M                          & \underline{0.223} & \underline{0.389} & \underline{0.333} & \underline{0.395} & \underline{0.348} & \textbf{0.206} & \textbf{0.342} & 0.225 & 0.452 & 0.385 & 0.330 \\

            \hline
        \end{tabular}}
    \caption{\textbf{Various metrics on the Eyecandies dataset for several multimodal AD methods.} Best results in \textbf{bold}, runner-ups \underline{underlined}.}
    \label{tab:aupro_iauroc_comparisons_eyecandies}
\end{table*}

    \section{Additional Qualitative Results}
        \label{sec:eyecandies}
        
\begin{figure*}[h]
  \centering
  \setlength{\tabcolsep}{1pt}
  \begin{tabular}{cc}
    
      \begin{tabular}{cccc}
        \textbf{RGB} & \textbf{PC} & \textbf{GT} & $\Psi$ \\
        
            \includegraphics[width=0.1\linewidth]{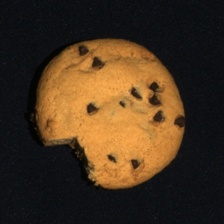} & 
            \includegraphics[width=0.1\linewidth]{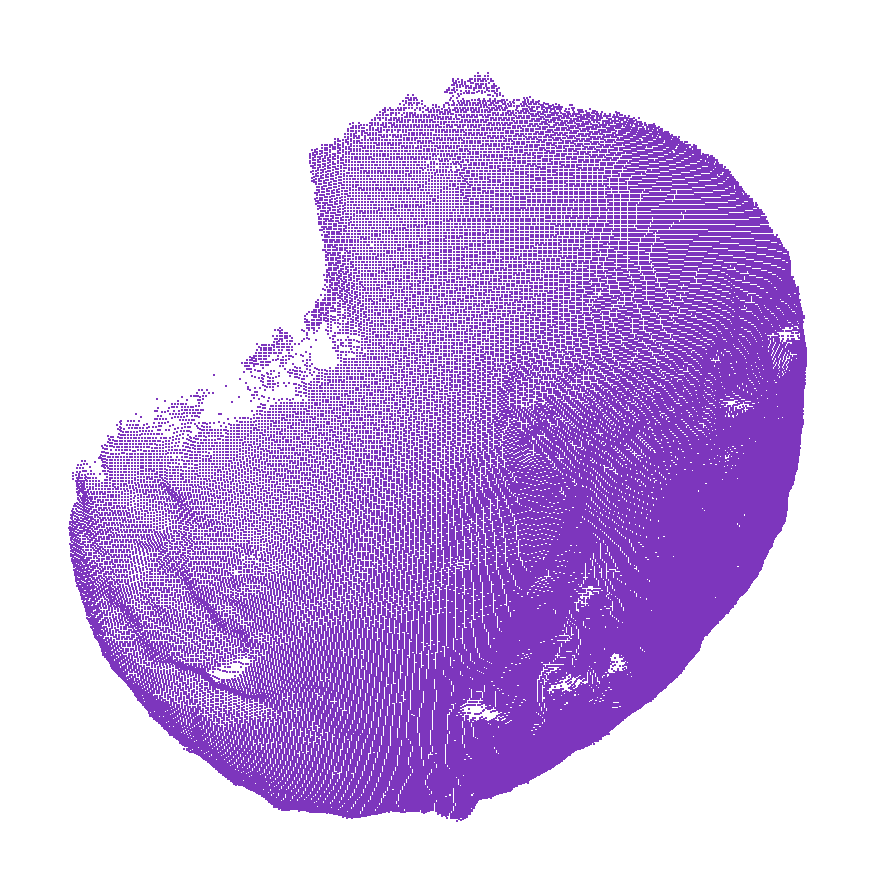} & 
            \includegraphics[width=0.1\linewidth]{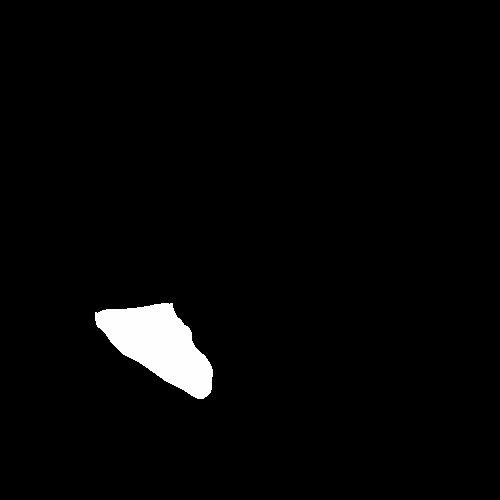} & 
            \includegraphics[width=0.1\linewidth]{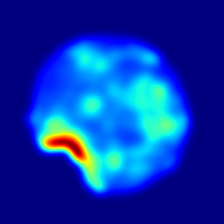} \\
        
            \includegraphics[width=0.1\linewidth]{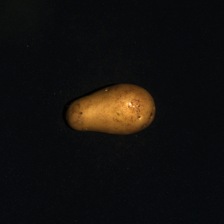} & 
            \includegraphics[width=0.1\linewidth]{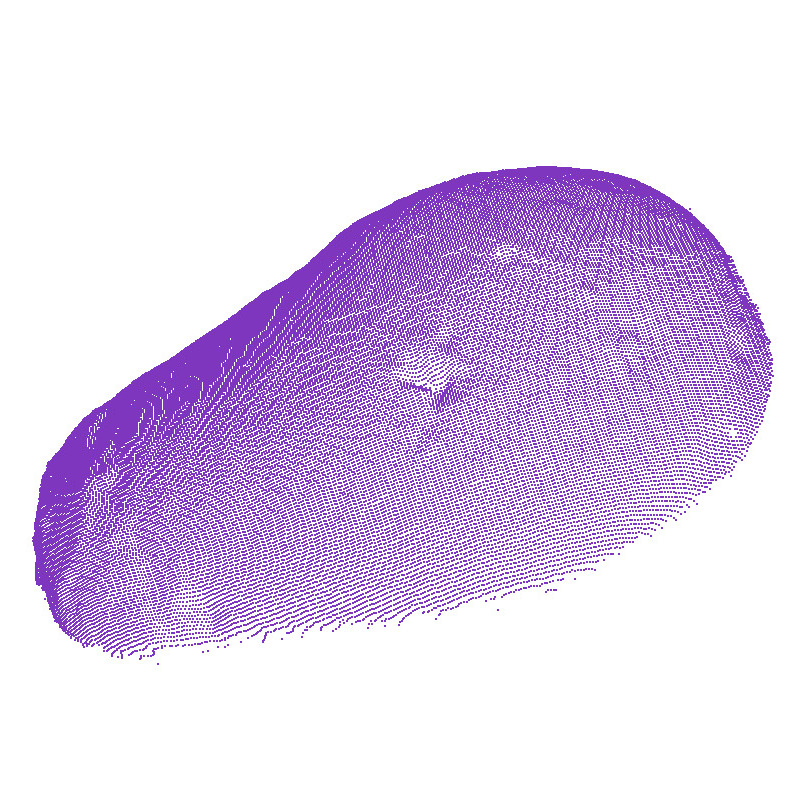} &
            \includegraphics[width=0.1\linewidth]{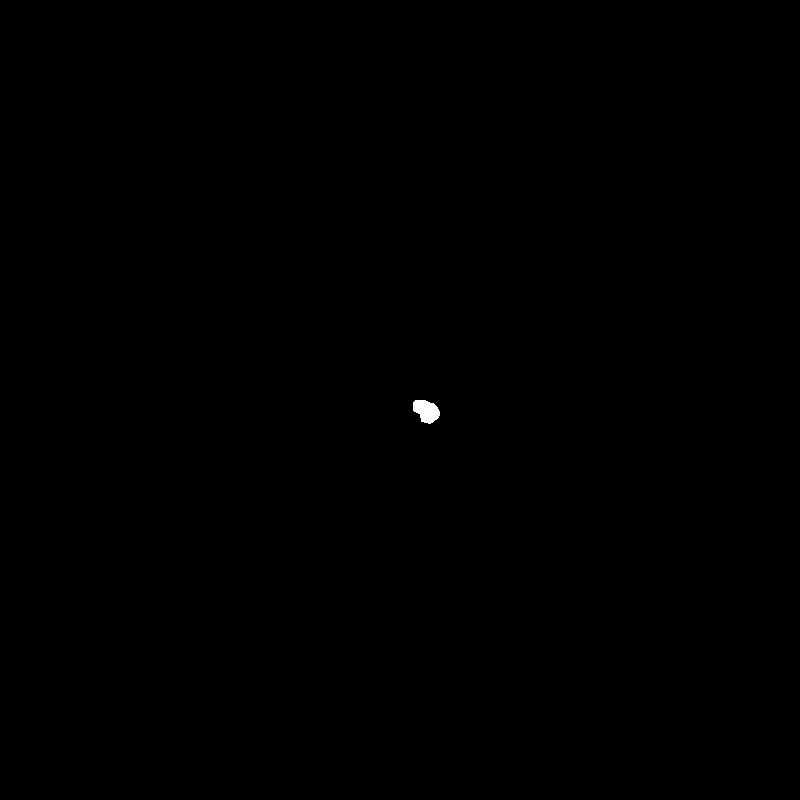} & 
            \includegraphics[width=0.1\linewidth]{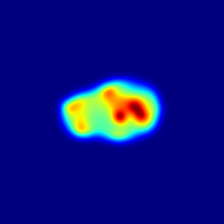} \\
            \multicolumn{4}{c}{MVTec 3D-AD}
    \end{tabular}
    & \hspace{1cm}
    \begin{tabular}{cccc}
     \textbf{RGB} & \textbf{PC} & \textbf{GT} & $\Psi$ \\
            \includegraphics[width=0.1\linewidth]{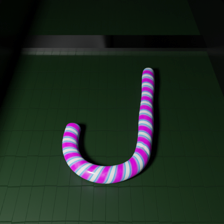} & 
            \includegraphics[width=0.1\linewidth]{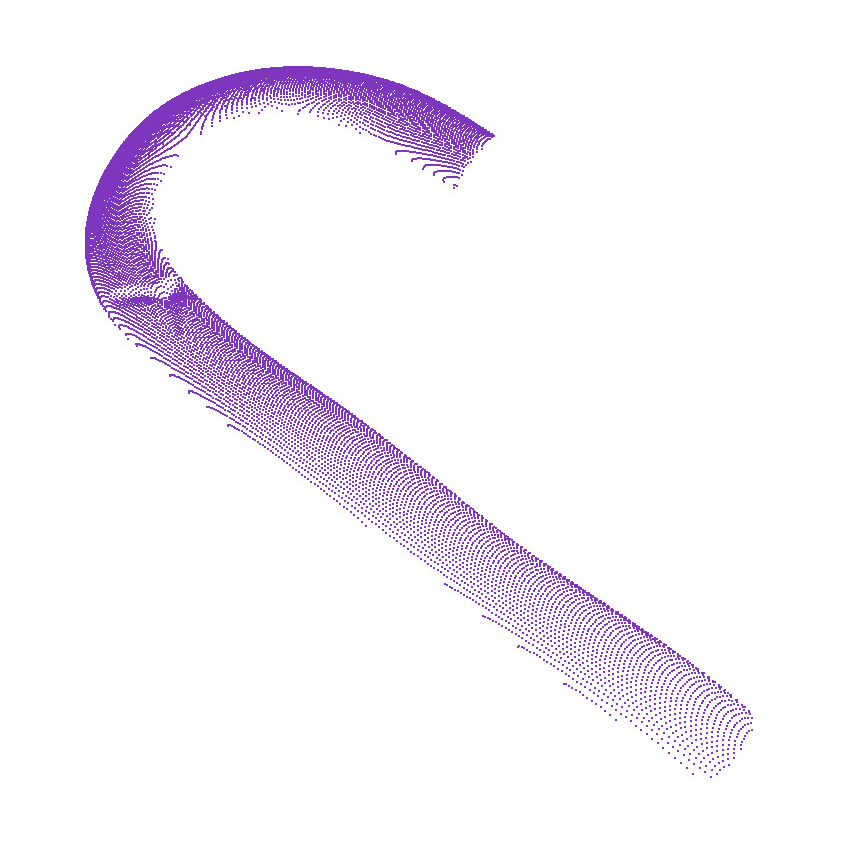} & 
            \includegraphics[width=0.1\linewidth]{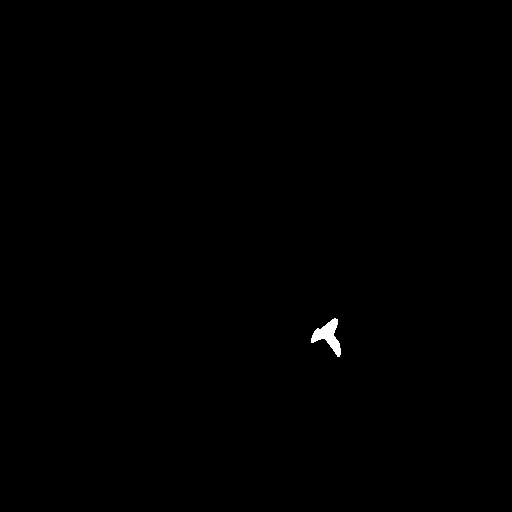} & 
            \includegraphics[width=0.1\linewidth]{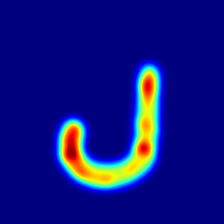}\\

            \includegraphics[width=0.1\linewidth]{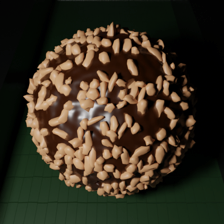} &
            \includegraphics[width=0.1\linewidth]{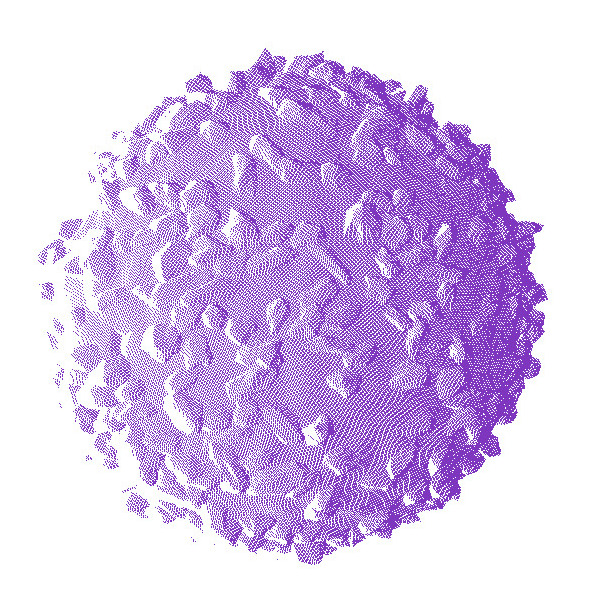} & 
            \includegraphics[width=0.1\linewidth]{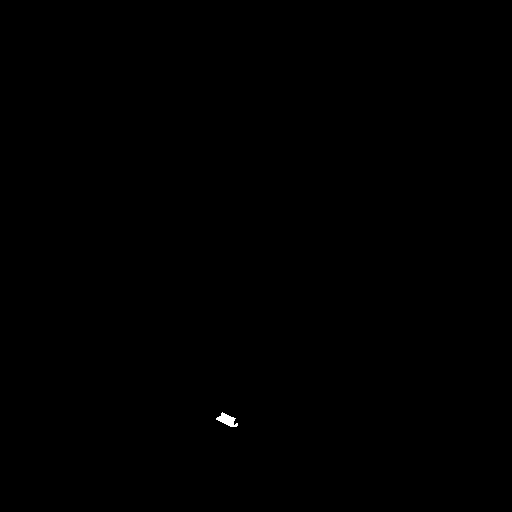} &
            \includegraphics[width=0.1\linewidth]{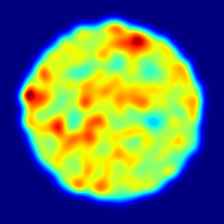} \\
            \multicolumn{4}{c}{Eyecandies}
        \end{tabular}      
      \end{tabular}
  \caption{
    \textbf{Failure cases. Results on MVTec 3D-AD (left) and Eyecandies (right).}
    }
  \label{fig:failure}
\end{figure*}

        In \cref{fig:failure}, we highlight some failure cases of this approach.
        For instance, in the first left row, we note that our method cannot detect the missing left part of the cookie. Nevertheless, we predict higher anomaly scores for the area adjacent to the defect.
        In the second left row, the potato presents a tiny defect on its body, while the anomaly map --- although covering the defect correctly --- predicts a much broader anomaly.
        In the first and second right rows, the candy cane and the hazelnut truffle present high-frequency 2D or 3D patterns that produce higher anomaly scores compared to the real defects.  
        
\begin{figure*}
  \centering
  \setlength{\tabcolsep}{1pt}
      \begin{tabular}{cccccccccccc}
        & \textbf{Bagel} & \textbf{Cable Gl.} & \textbf{Carrot} & \textbf{Cookie} & \textbf{Dowel} & \textbf{Foam} & \textbf{Peach} & \textbf{Potato} & \textbf{Rope} & \textbf{Tire} \\
        
        \rotatebox{90}{\hspace{0.35cm} RGB} & 
            \includegraphics[width=0.1\linewidth]{images/bagel/rgb_007.jpg} & 
            \includegraphics[width=0.1\linewidth]{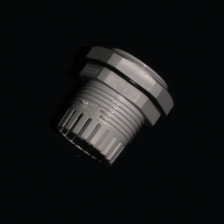} & 
            \includegraphics[width=0.1\linewidth]{images/carrot/rgb_015.jpg} & 
            \includegraphics[width=0.1\linewidth]{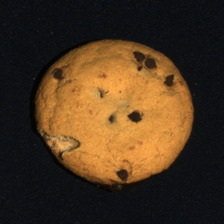} & 
            \includegraphics[width=0.1\linewidth]{images/dowel/rgb_017.jpg} &
            \includegraphics[width=0.1\linewidth]{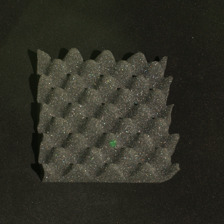} & 
            \includegraphics[width=0.1\linewidth]{images/peach/rgb_023.jpg} & 
            \includegraphics[width=0.1\linewidth]{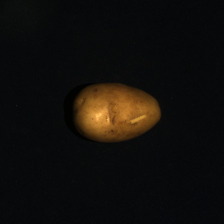} & 
            \includegraphics[width=0.1\linewidth]{images/rope/rgb_010_bg.jpg} & 
            \includegraphics[width=0.1\linewidth]{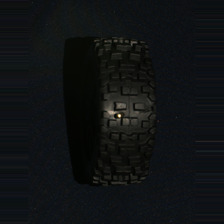} \\
        
        \rotatebox{90}{\hspace{0.45cm} PC} & 
            \includegraphics[width=0.1\linewidth]{images/bagel/pc_007.jpg} & 
            \includegraphics[width=0.1\linewidth]{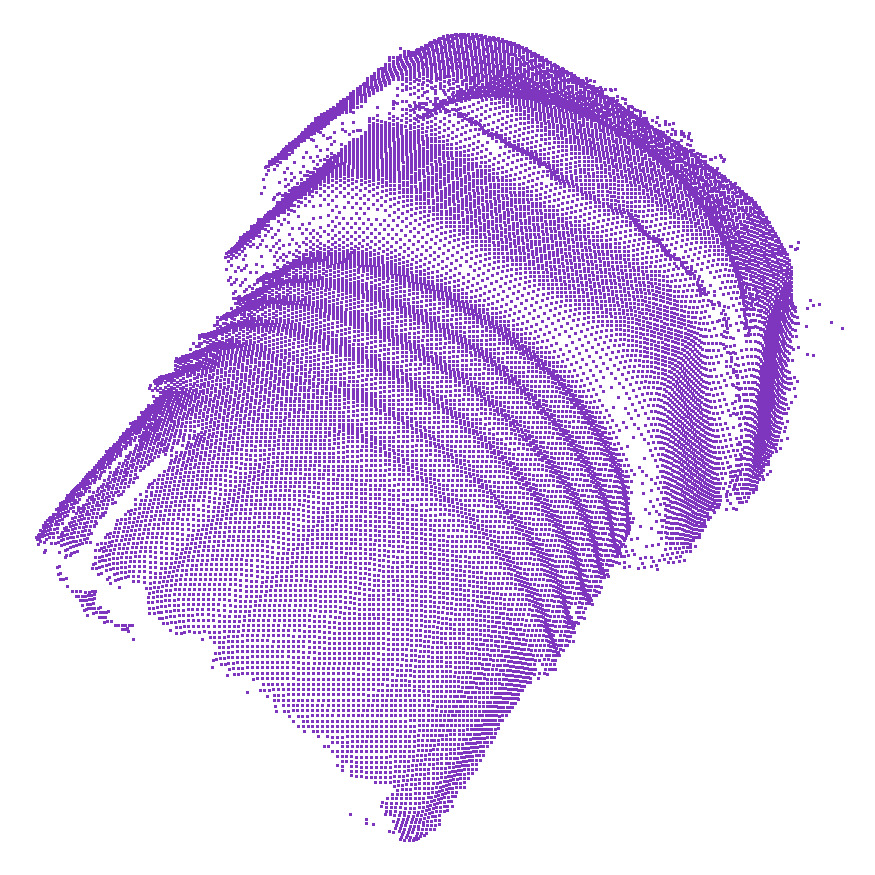} & 
            \includegraphics[width=0.1\linewidth]{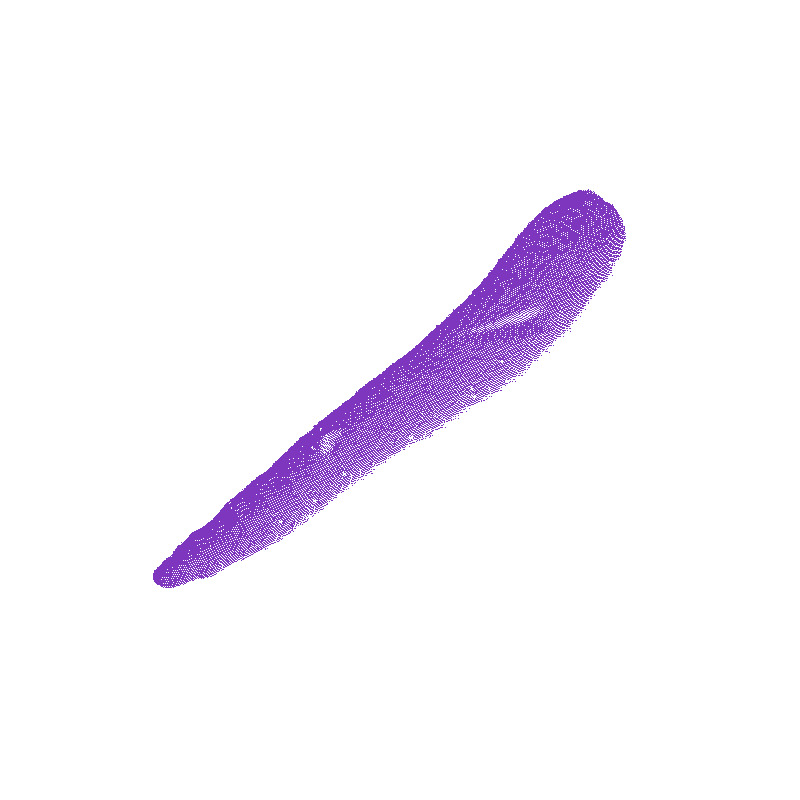} & 
            \includegraphics[width=0.1\linewidth]{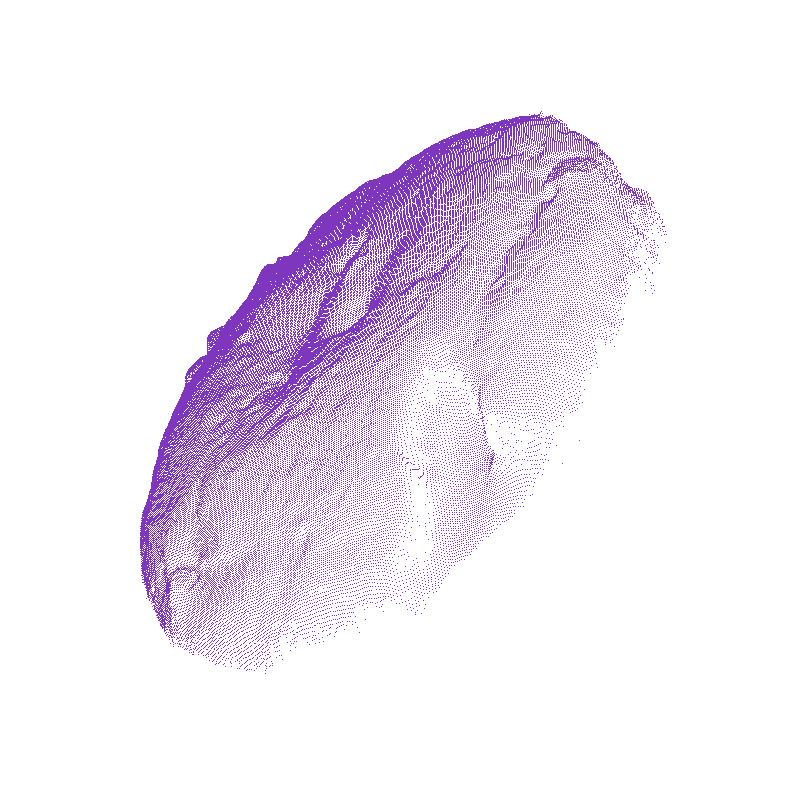} & 
            \includegraphics[width=0.1\linewidth]{images/dowel/pc_017.jpg} &
            \includegraphics[width=0.1\linewidth]{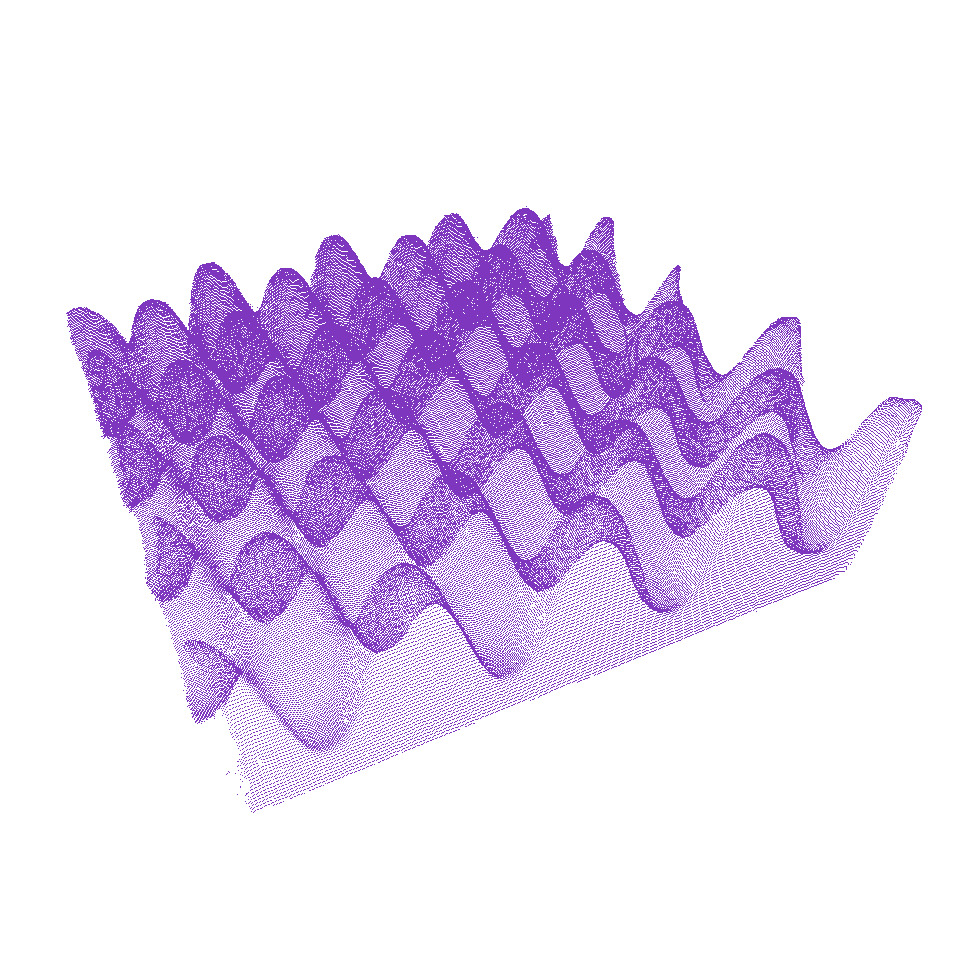} & 
            \includegraphics[width=0.1\linewidth]{images/peach/pc_023.jpg} & 
            \includegraphics[width=0.1\linewidth]{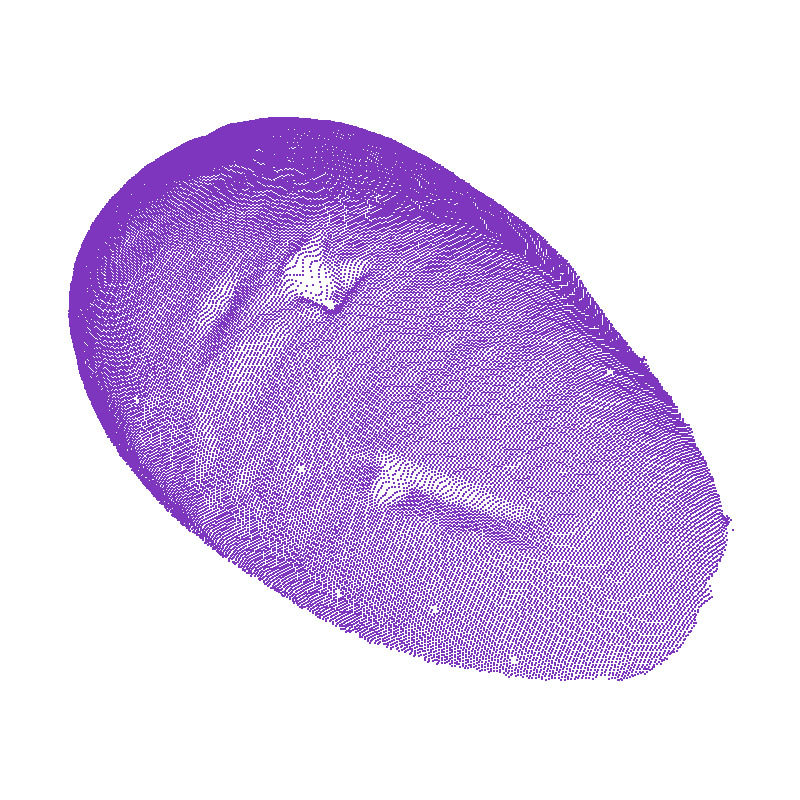} & 
            \includegraphics[width=0.1\linewidth]{images/rope/pc_010.jpg} & 
            \includegraphics[width=0.1\linewidth]{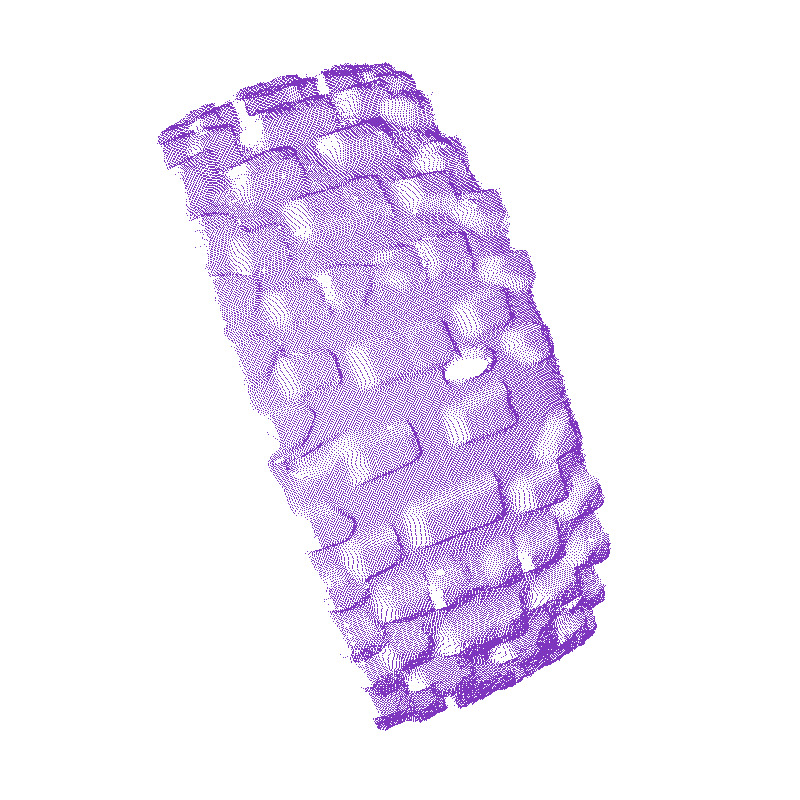} \\
        
        \rotatebox{90}{\hspace{0.45cm} GT} & 
            \includegraphics[width=0.1\linewidth]{images/bagel/gt_007.jpg} & 
            \includegraphics[width=0.1\linewidth]{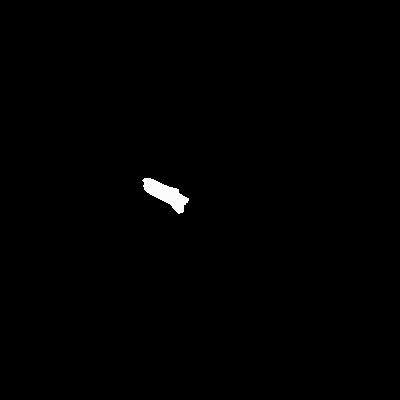} & 
            \includegraphics[width=0.1\linewidth]{images/carrot/gt_015.jpg} & 
            \includegraphics[width=0.1\linewidth]{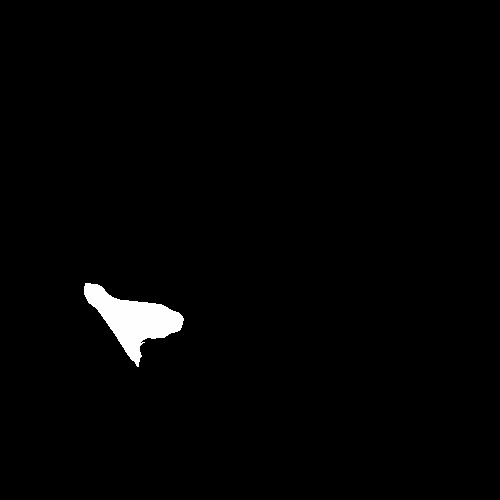} & 
            \includegraphics[width=0.1\linewidth]{images/dowel/gt_017.jpg} &
            \includegraphics[width=0.1\linewidth]{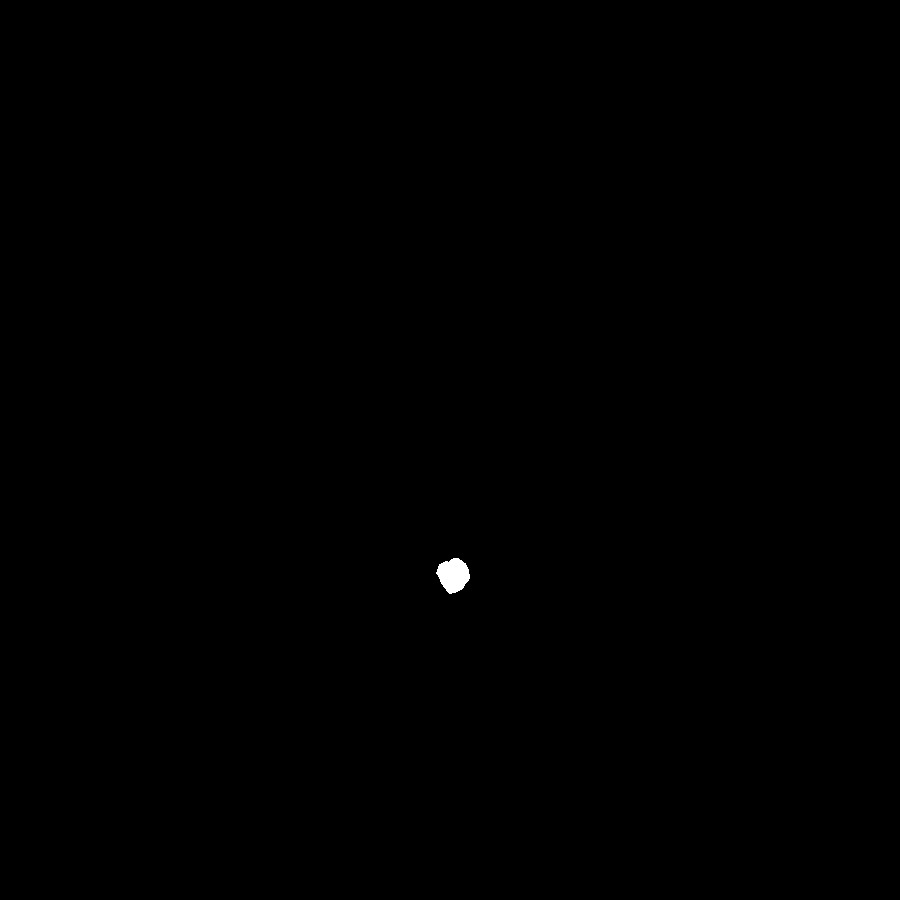} & 
            \includegraphics[width=0.1\linewidth]{images/peach/gt_023.jpg} & 
            \includegraphics[width=0.1\linewidth]{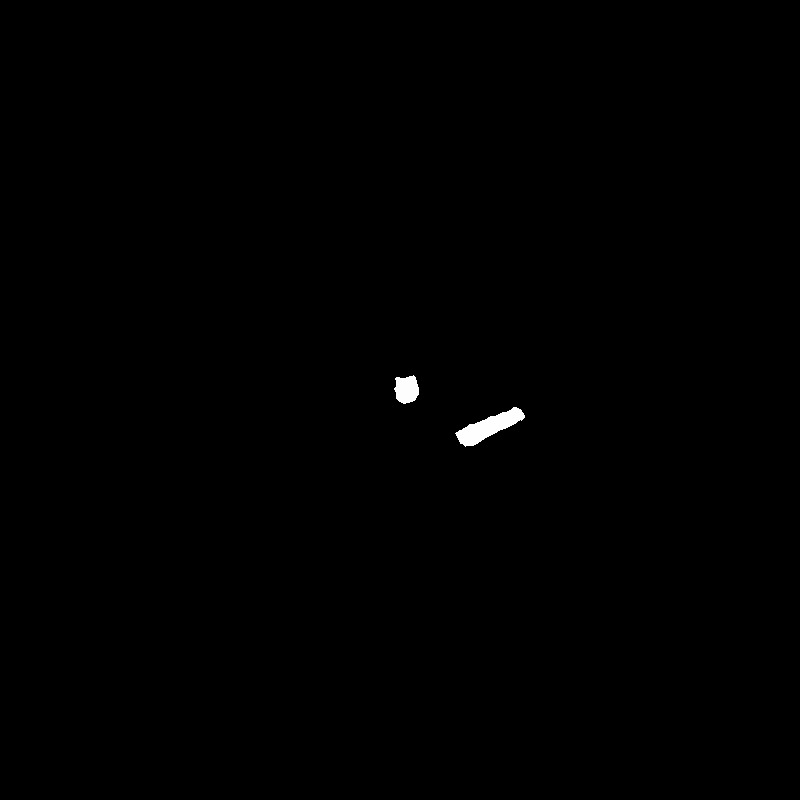} & 
            \includegraphics[width=0.1\linewidth]{images/rope/gt_010_pad.jpg} & 
            \includegraphics[width=0.1\linewidth]{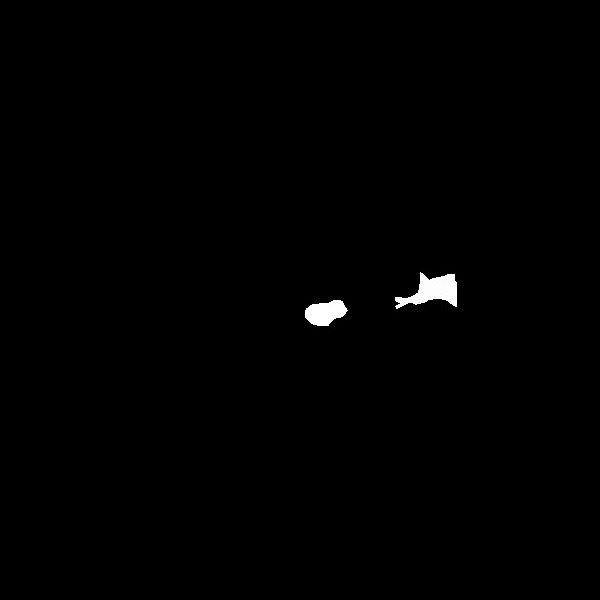} \\
        
        \rotatebox{90}{\hspace{0.30cm} M3DM} & 
            \includegraphics[width=0.1\linewidth]{images/bagel/m3dm_007.jpg} & 
            \includegraphics[width=0.1\linewidth]{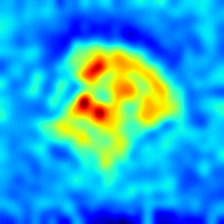} & 
            \includegraphics[width=0.1\linewidth]{images/carrot/m3dm_015.jpg} & 
            \includegraphics[width=0.1\linewidth]{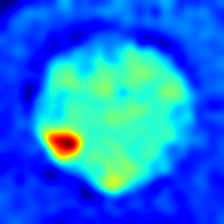} & 
            \includegraphics[width=0.1\linewidth]{images/dowel/m3dm_017.jpg} &
            \includegraphics[width=0.1\linewidth]{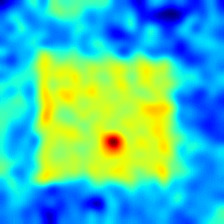} & 
            \includegraphics[width=0.1\linewidth]{images/peach/m3dm_023.jpg} & 
            \includegraphics[width=0.1\linewidth]{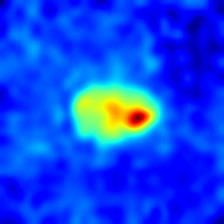} & 
            \includegraphics[width=0.1\linewidth]{images/rope/m3dm_010.jpg} & 
            \includegraphics[width=0.1\linewidth]{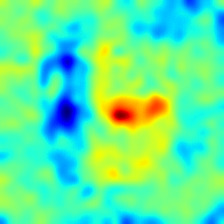} \\
            
        \rotatebox{90}{\hspace{0.35cm} Ours} & 
            \includegraphics[width=0.1\linewidth]{images/bagel/cos_007.jpg} & 
            \includegraphics[width=0.1\linewidth]{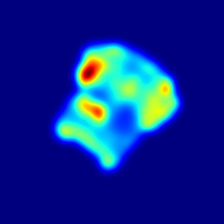} & 
            \includegraphics[width=0.1\linewidth]{images/carrot/cos_015.jpg} & 
            \includegraphics[width=0.1\linewidth]{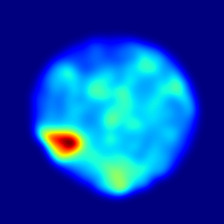} & 
            \includegraphics[width=0.1\linewidth]{images/dowel/cos_017.jpg} &
            \includegraphics[width=0.1\linewidth]{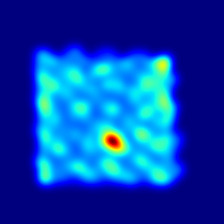} & 
            \includegraphics[width=0.1\linewidth]{images/peach/cos_023.jpg} & 
            \includegraphics[width=0.1\linewidth]{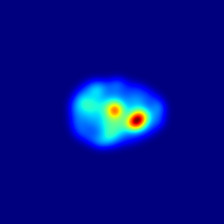} & 
            \includegraphics[width=0.1\linewidth]{images/rope/cos_010.jpg} & 
            \includegraphics[width=0.1\linewidth]{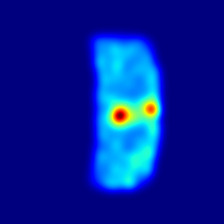} \\

        \\

        \\

        & \textbf{Bagel} & \textbf{Cable Gl.} & \textbf{Carrot} & \textbf{Cookie} & \textbf{Dowel} & \textbf{Foam} & \textbf{Peach} & \textbf{Potato} & \textbf{Rope} & \textbf{Tire} \\

        \rotatebox{90}{\hspace{0.35cm} RGB} & 
            \includegraphics[width=0.1\linewidth]{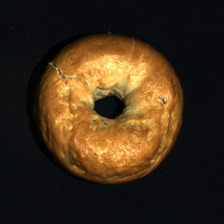} & 
            \includegraphics[width=0.1\linewidth]{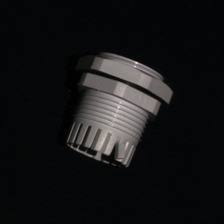} & 
            \includegraphics[width=0.1\linewidth]{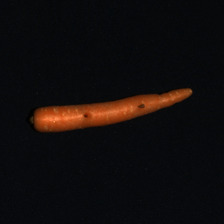} & 
            \includegraphics[width=0.1\linewidth]{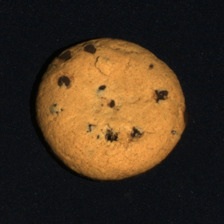} & 
            \includegraphics[width=0.1\linewidth]{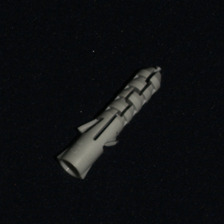} &
            \includegraphics[width=0.1\linewidth]{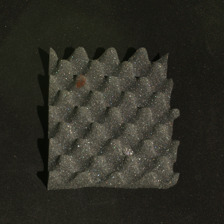} & 
            \includegraphics[width=0.1\linewidth]{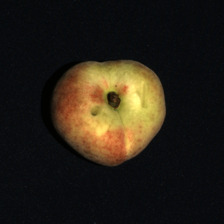} & 
            \includegraphics[width=0.1\linewidth]{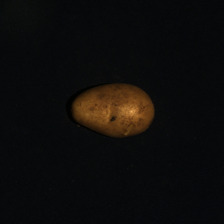} & 
            \includegraphics[width=0.1\linewidth]{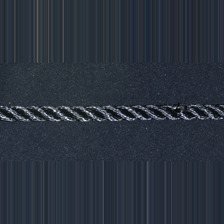} & 
            \includegraphics[width=0.1\linewidth]{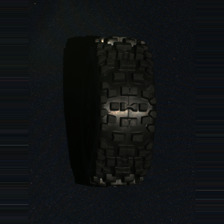} \\
        
        \rotatebox{90}{\hspace{0.45cm} PC} & 
            \includegraphics[width=0.1\linewidth]{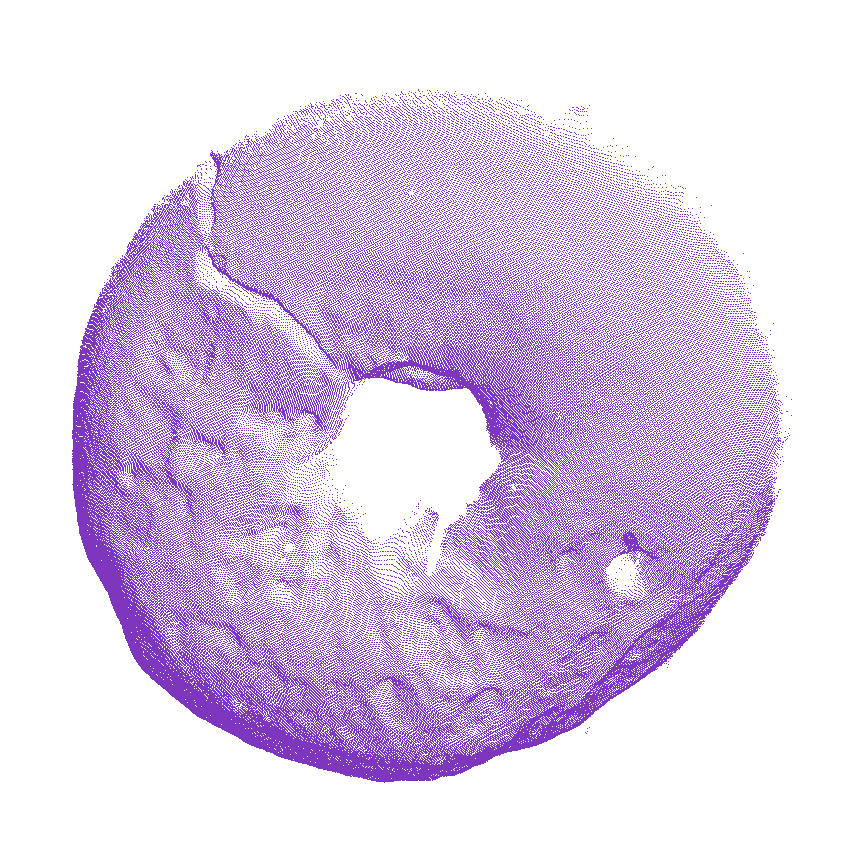} & 
            \includegraphics[width=0.1\linewidth]{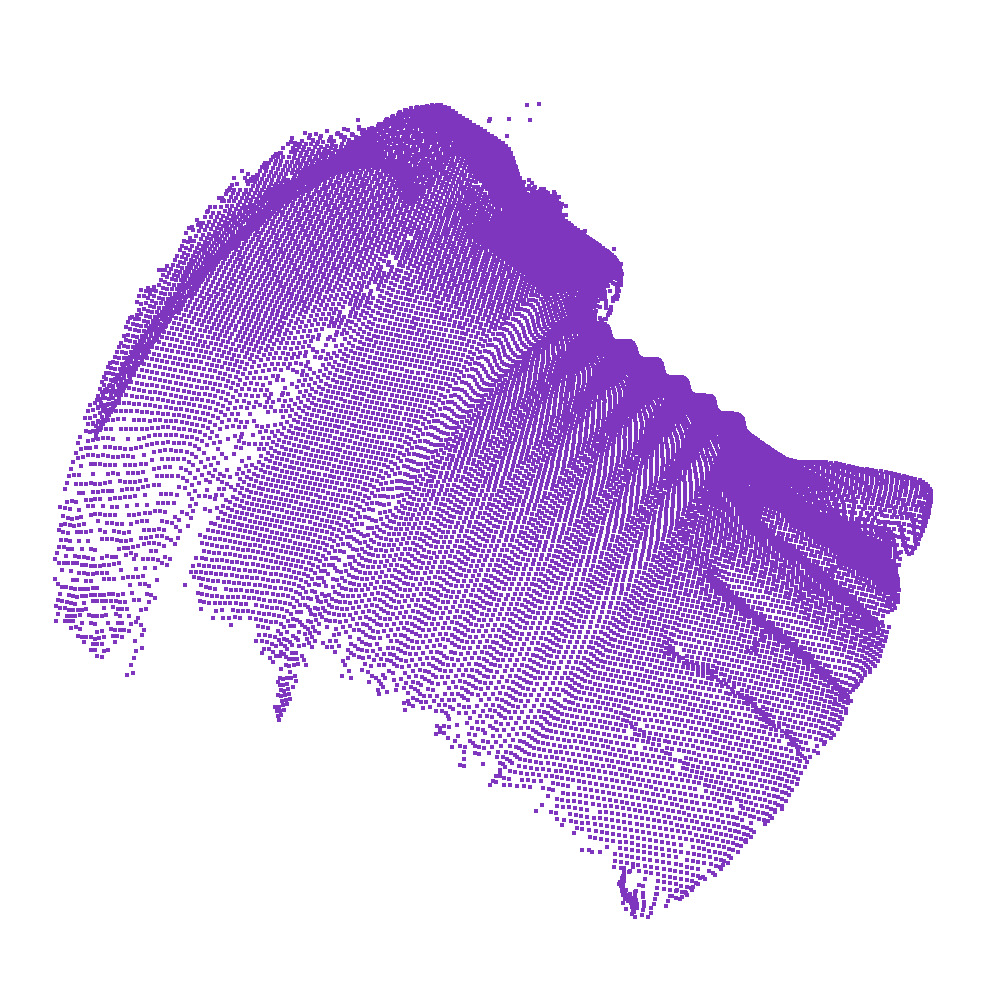} & 
            \includegraphics[width=0.1\linewidth]{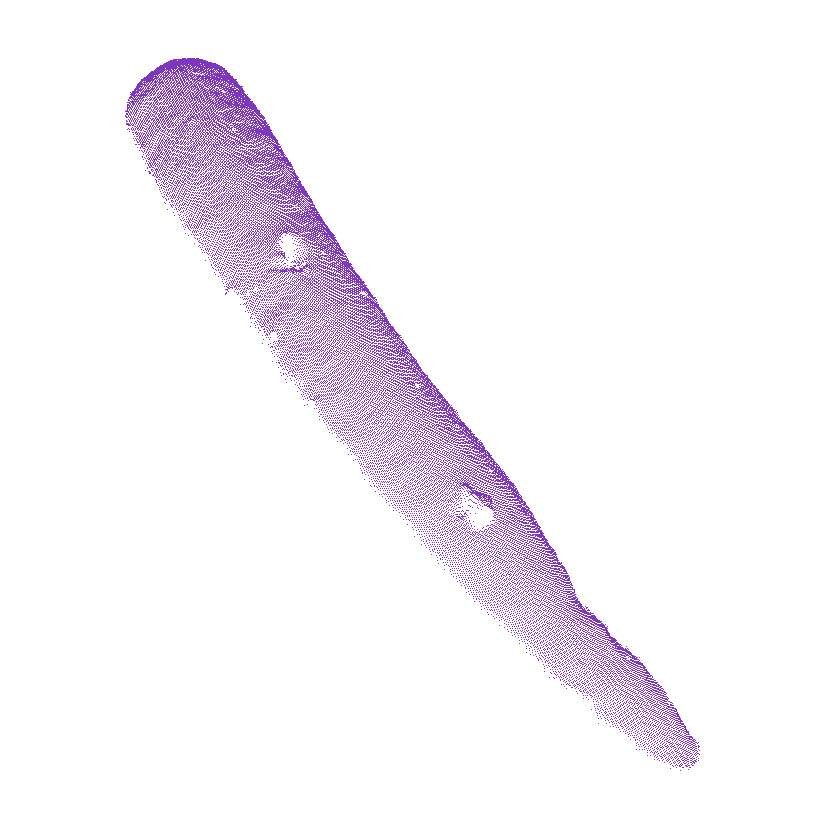} & 
            \includegraphics[width=0.1\linewidth]{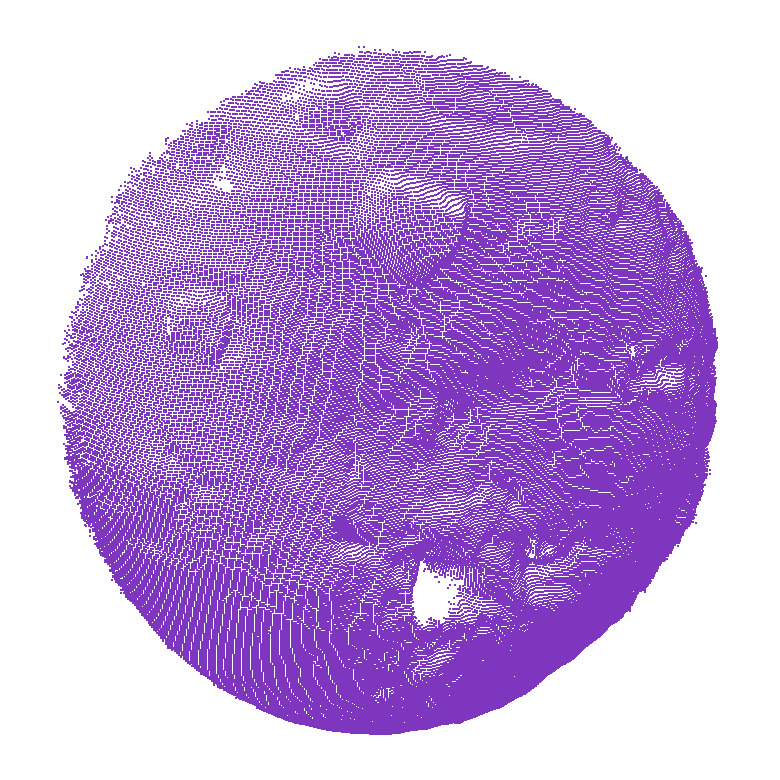} & 
            \includegraphics[width=0.1\linewidth]{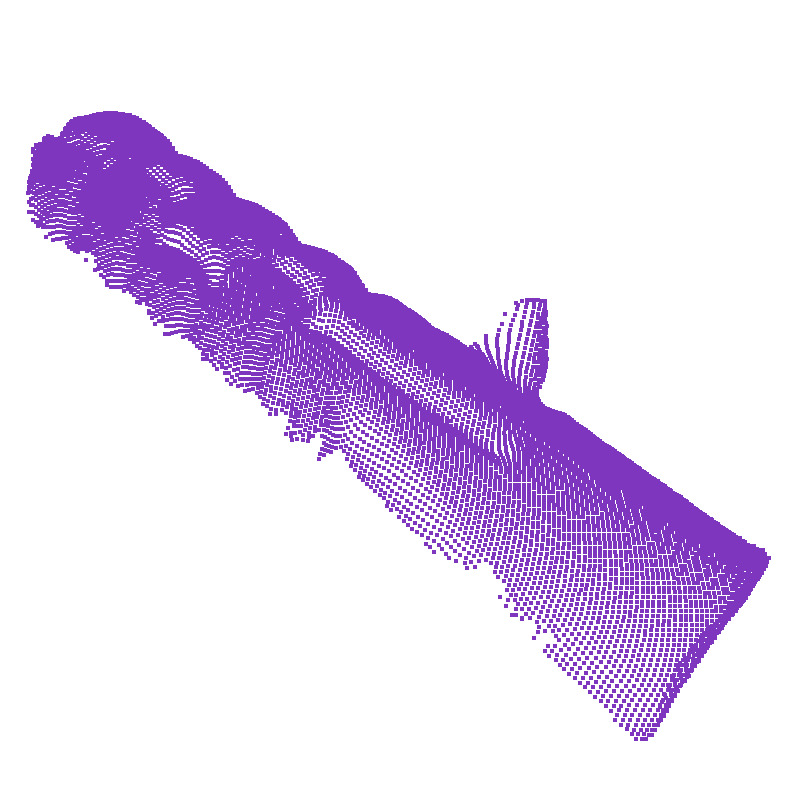} &
            \includegraphics[width=0.1\linewidth]{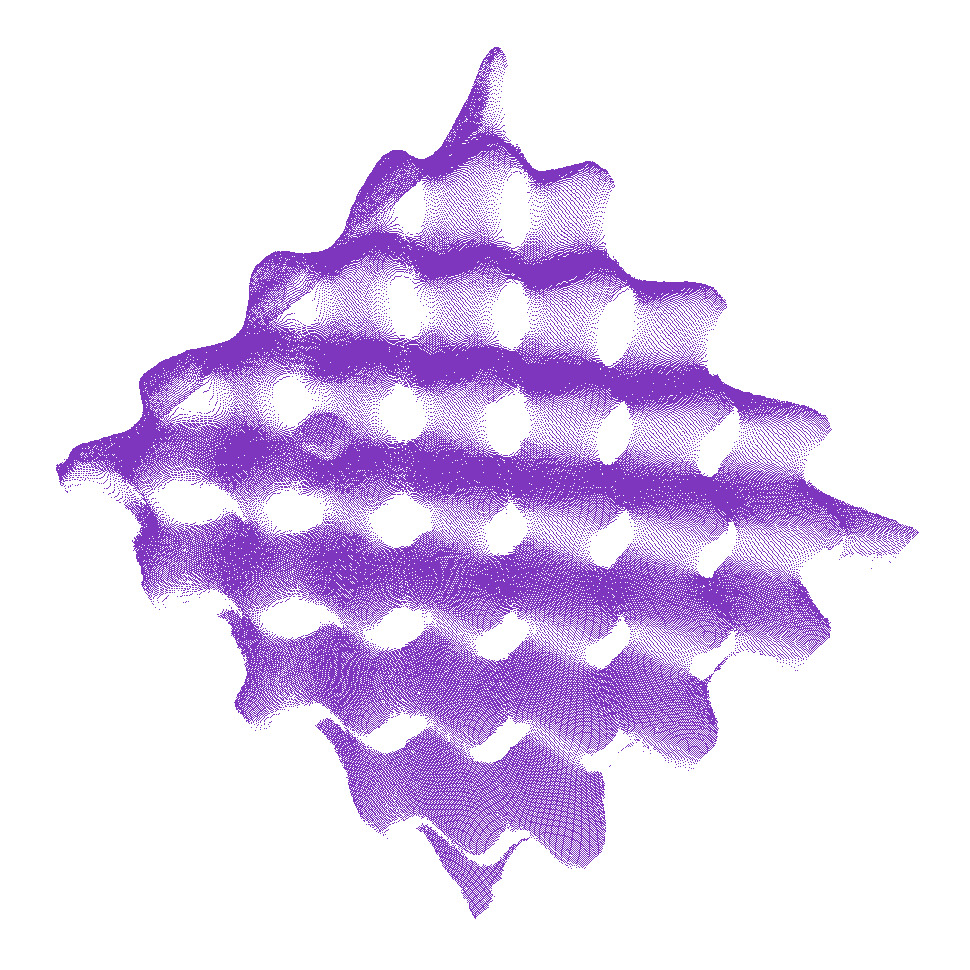} & 
            \includegraphics[width=0.1\linewidth]{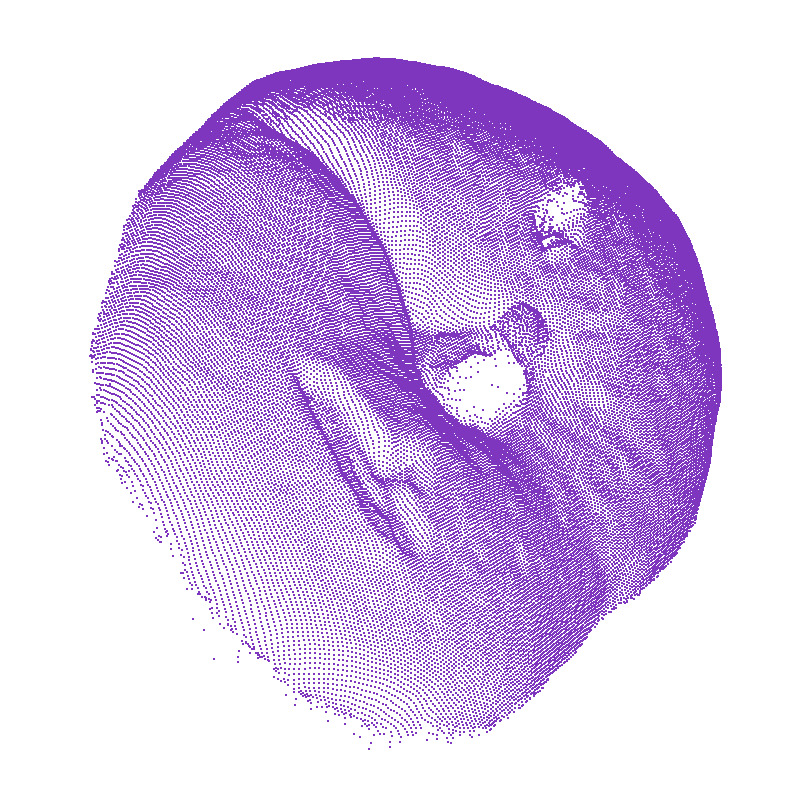} & 
            \includegraphics[width=0.1\linewidth]{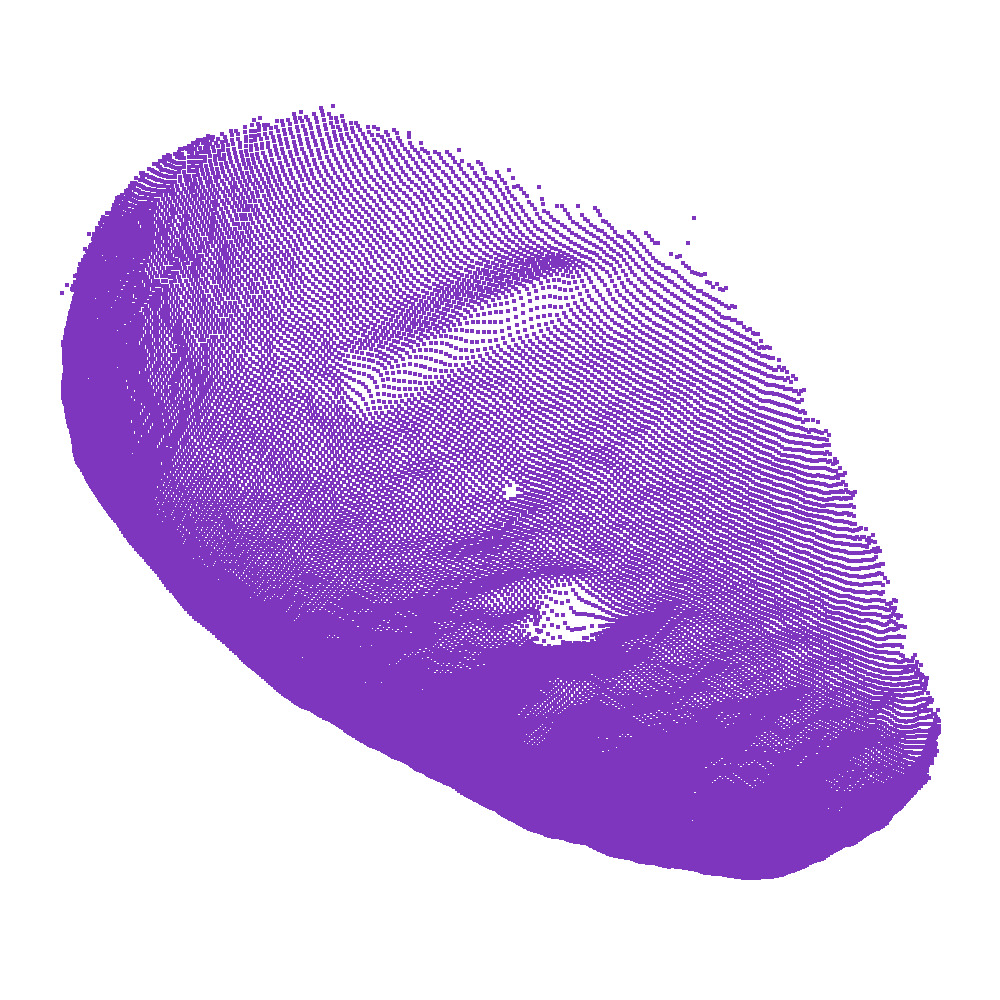} & 
            \includegraphics[width=0.1\linewidth]{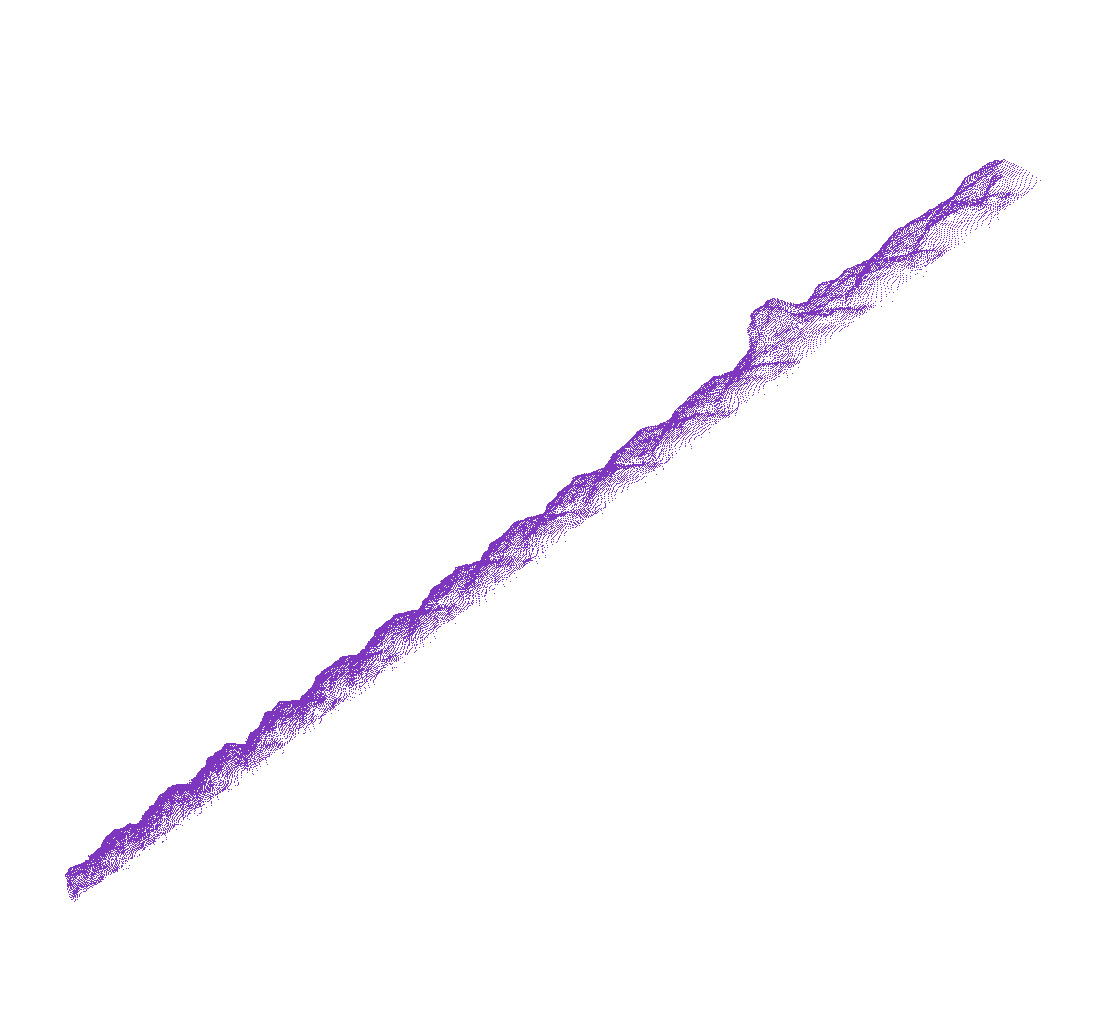} & 
            \includegraphics[width=0.1\linewidth]{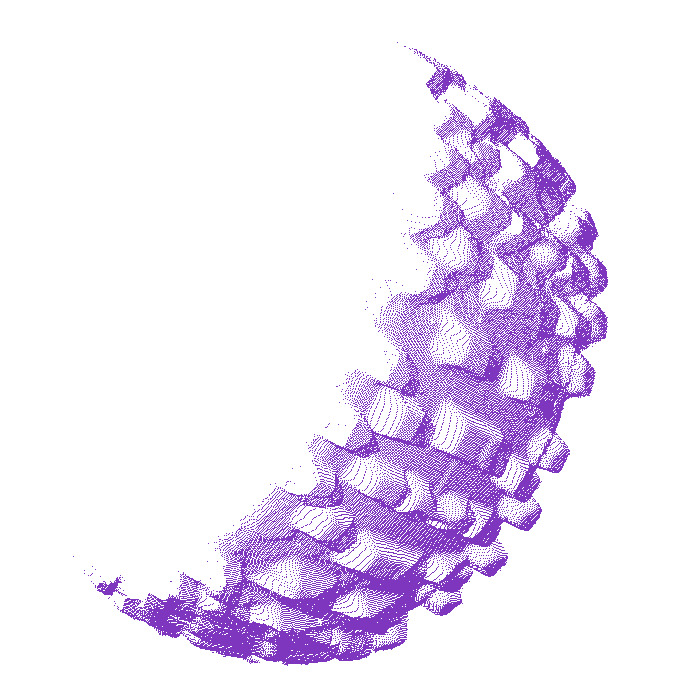} \\
        
        \rotatebox{90}{\hspace{0.45cm} GT} & 
            \includegraphics[width=0.1\linewidth]{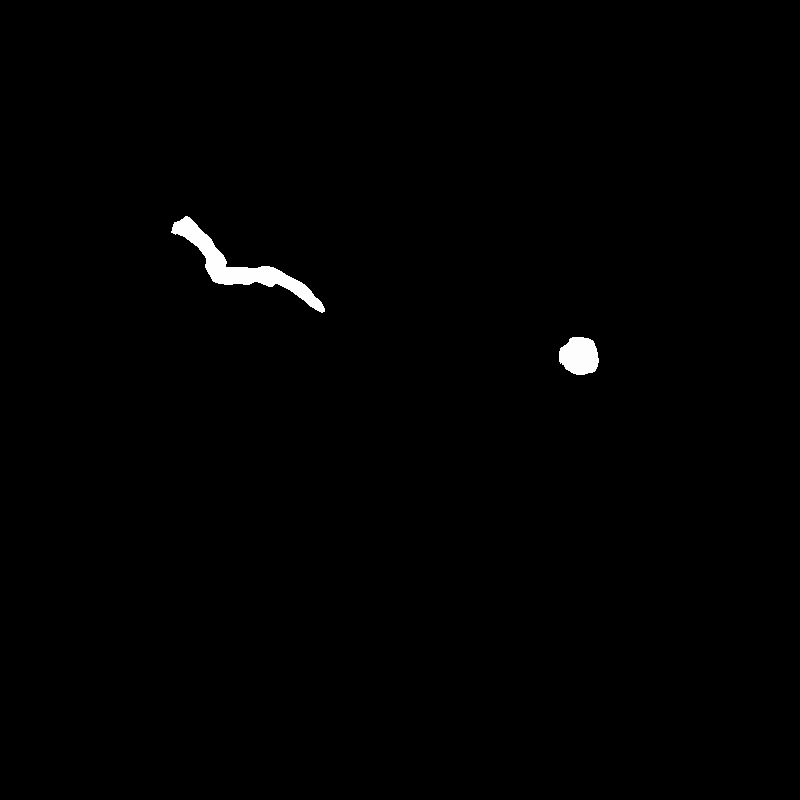} & 
            \includegraphics[width=0.1\linewidth]{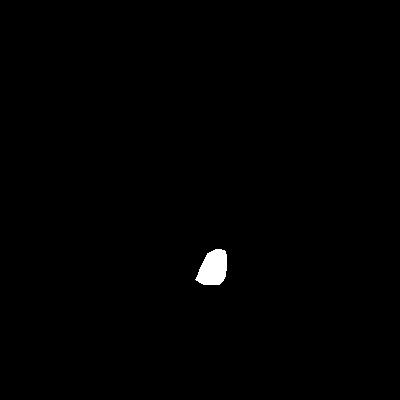} & 
            \includegraphics[width=0.1\linewidth]{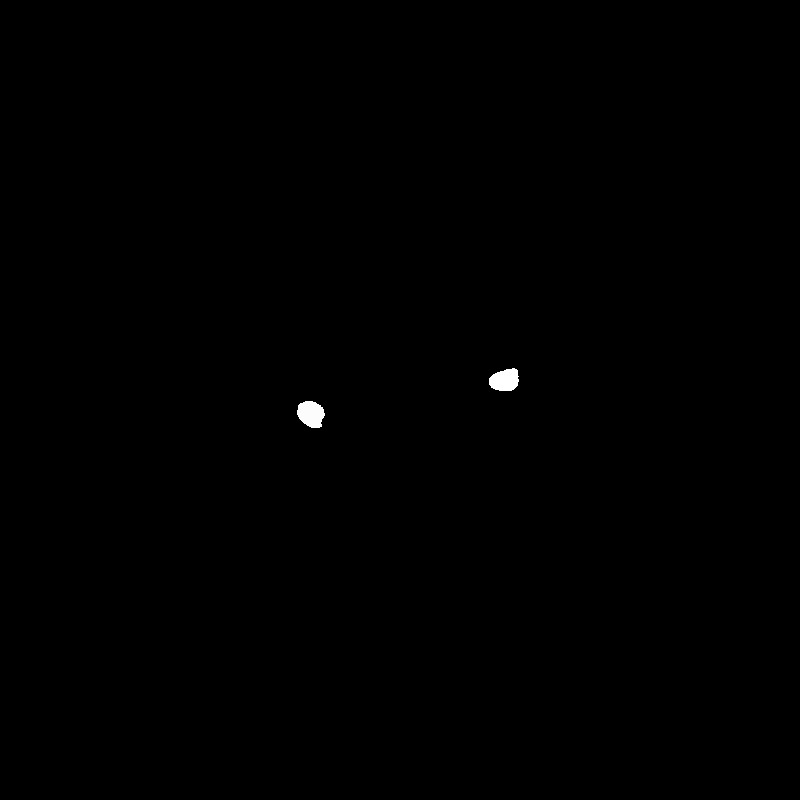} & 
            \includegraphics[width=0.1\linewidth]{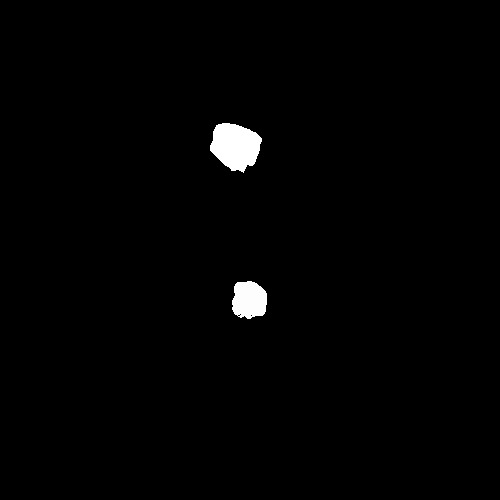} & 
            \includegraphics[width=0.1\linewidth]{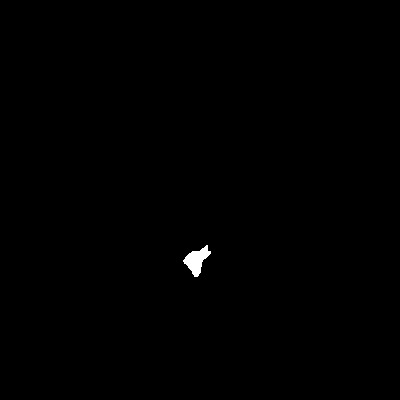} &
            \includegraphics[width=0.1\linewidth]{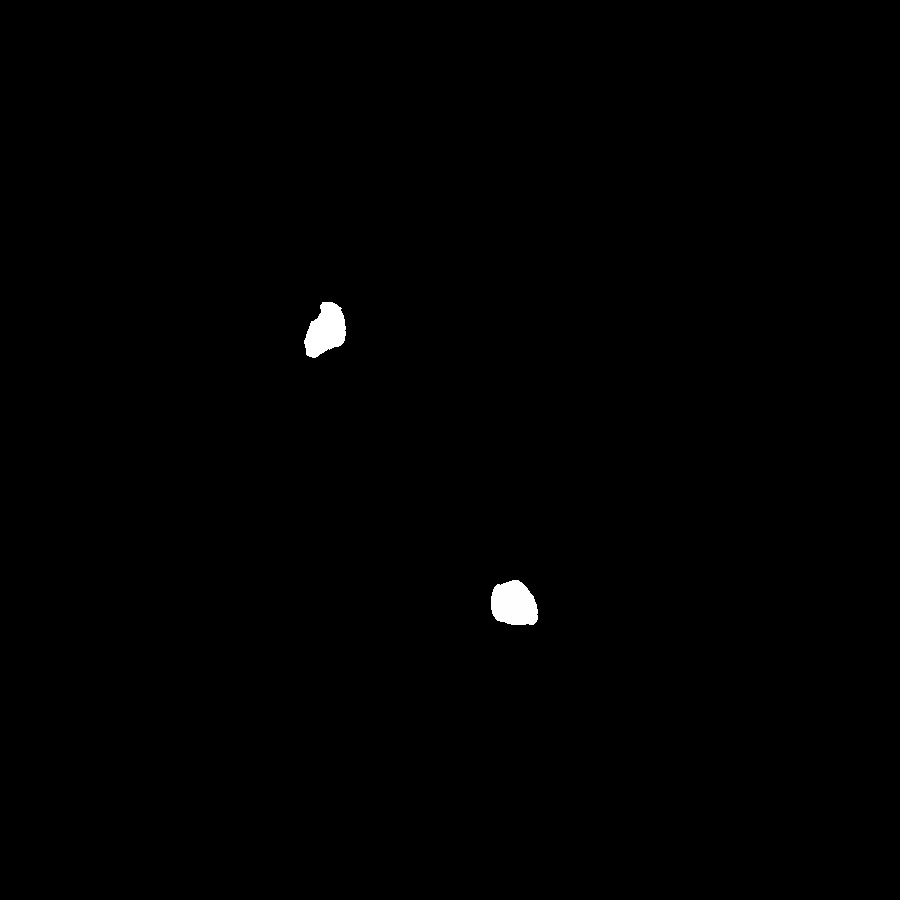} & 
            \includegraphics[width=0.1\linewidth]{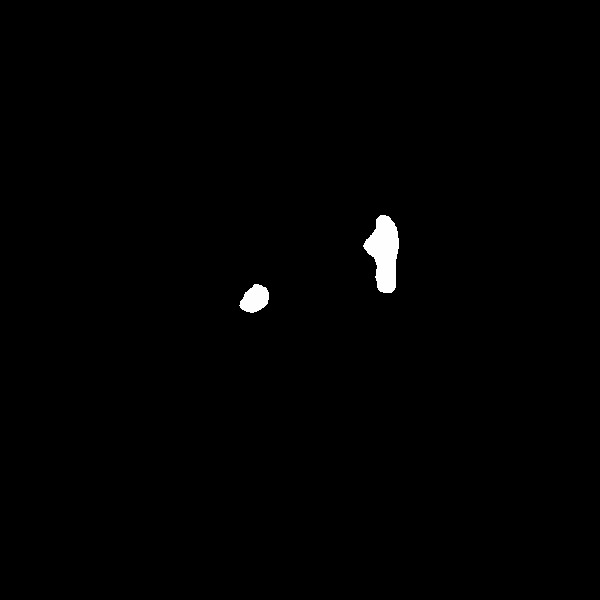} & 
            \includegraphics[width=0.1\linewidth]{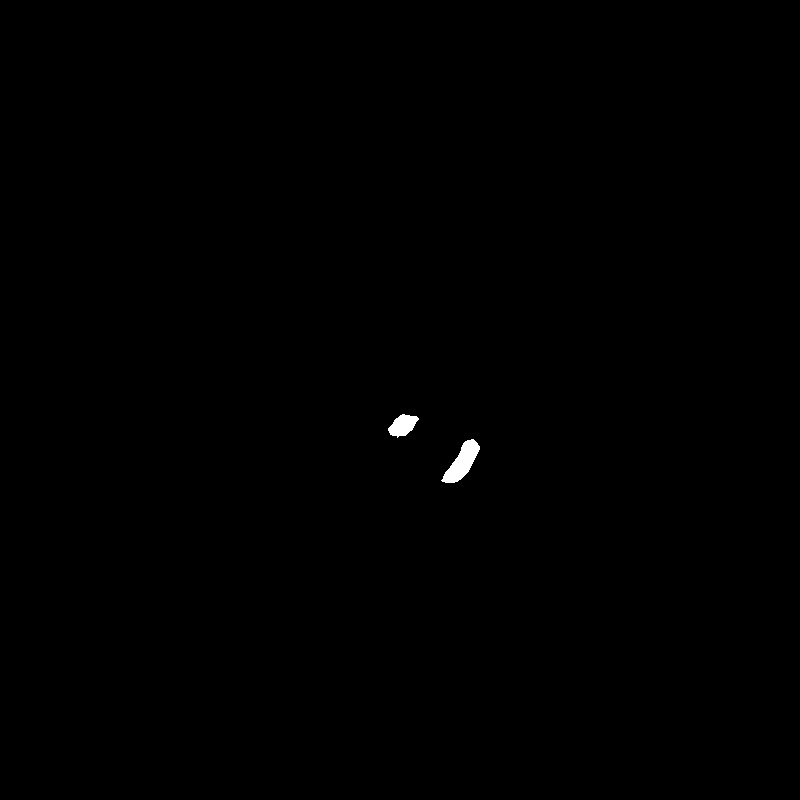} & 
            \includegraphics[width=0.1\linewidth]{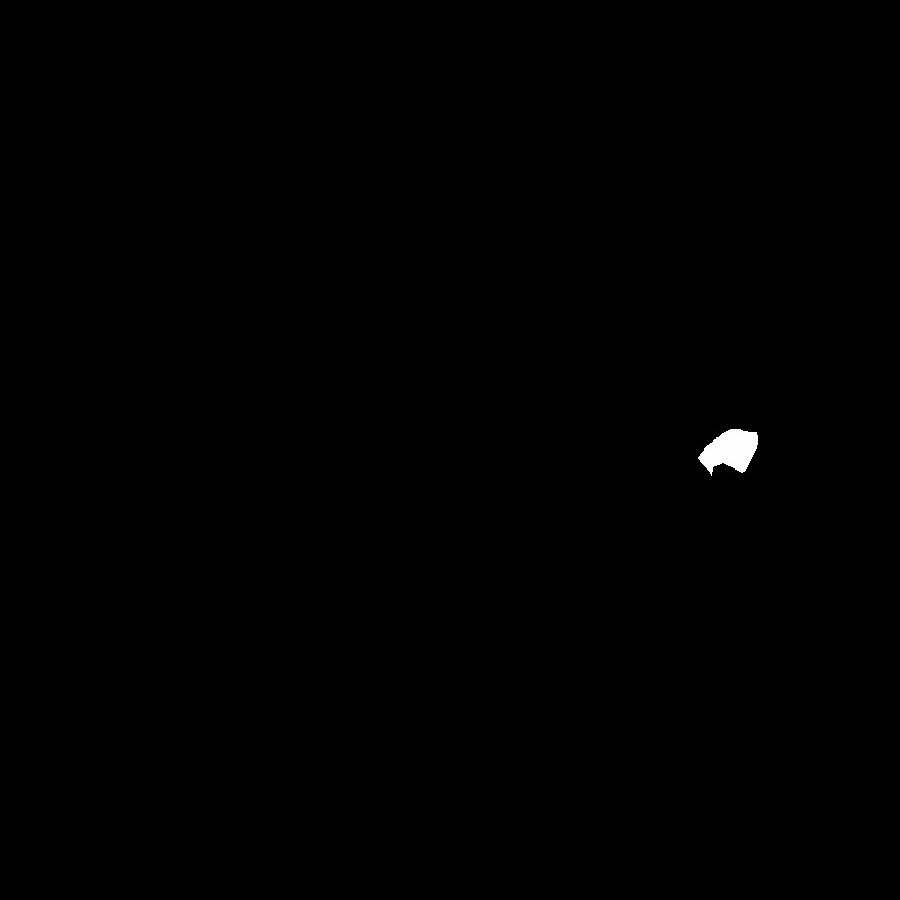} & 
            \includegraphics[width=0.1\linewidth]{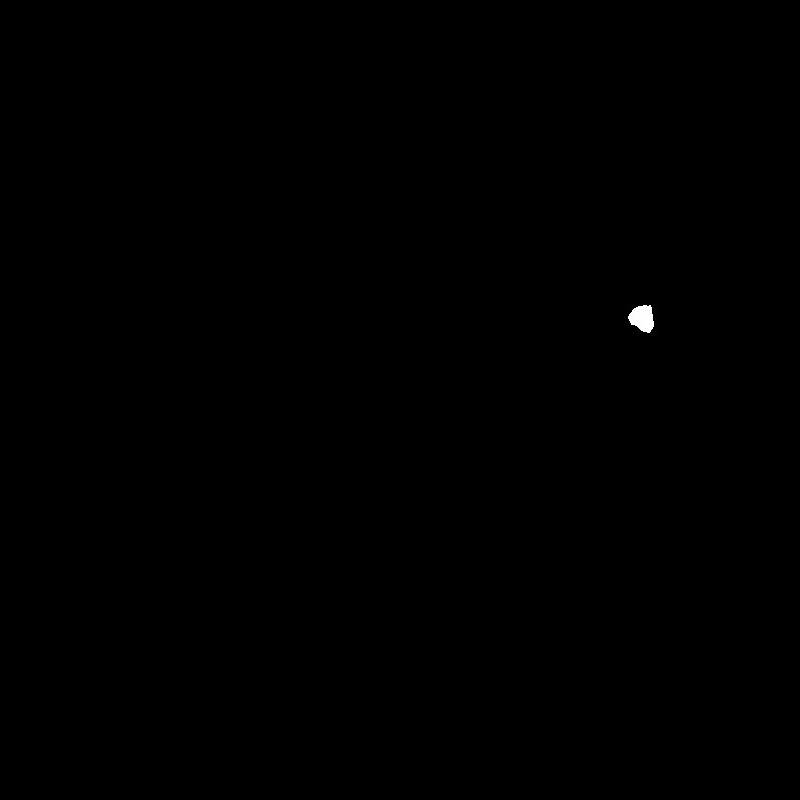} \\
        
        \rotatebox{90}{\hspace{0.30cm} M3DM} & 
            \includegraphics[width=0.1\linewidth]{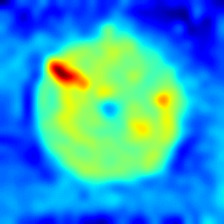} & 
            \includegraphics[width=0.1\linewidth]{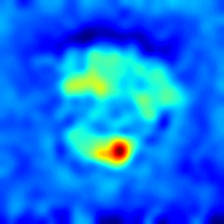} & 
            \includegraphics[width=0.1\linewidth]{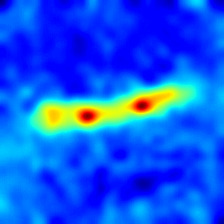} & 
            \includegraphics[width=0.1\linewidth]{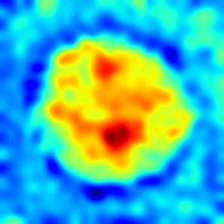} & 
            \includegraphics[width=0.1\linewidth]{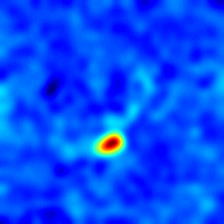} &
            \includegraphics[width=0.1\linewidth]{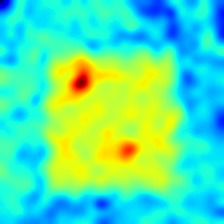} & 
            \includegraphics[width=0.1\linewidth]{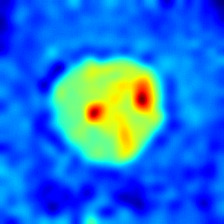} & 
            \includegraphics[width=0.1\linewidth]{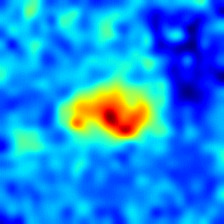} & 
            \includegraphics[width=0.1\linewidth]{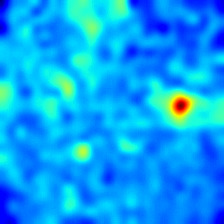} & 
            \includegraphics[width=0.1\linewidth]{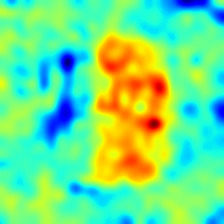} \\
            
        \rotatebox{90}{\hspace{0.35cm} Ours} & 
            \includegraphics[width=0.1\linewidth]{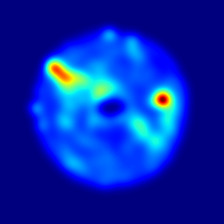} & 
            \includegraphics[width=0.1\linewidth]{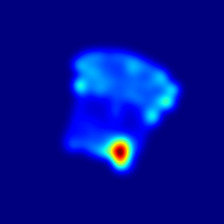} & 
            \includegraphics[width=0.1\linewidth]{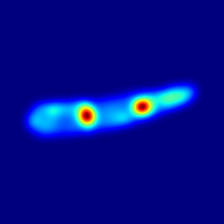} & 
            \includegraphics[width=0.1\linewidth]{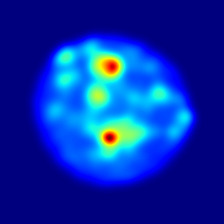} & 
            \includegraphics[width=0.1\linewidth]{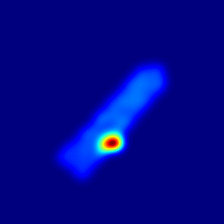} &
            \includegraphics[width=0.1\linewidth]{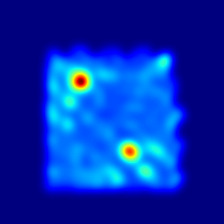} & 
            \includegraphics[width=0.1\linewidth]{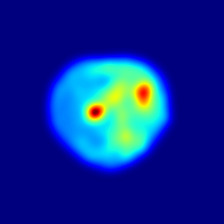} & 
            \includegraphics[width=0.1\linewidth]{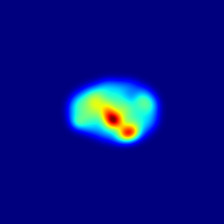} & 
            \includegraphics[width=0.1\linewidth]{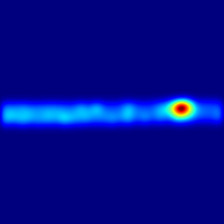} & 
            \includegraphics[width=0.1\linewidth]{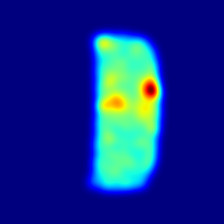} \\
            
      \end{tabular}
  \caption{Qualitative results for each class of the MVTec 3D-AD dataset}
  \label{fig:qualitatives_mvtec_full}
\end{figure*}

\begin{figure*}
  \centering
  \setlength{\tabcolsep}{1pt}
      \begin{tabular}{cccccccccccc}
        &             
        \textbf{Can. C.} &
        \textbf{Cho. C.} &
        \textbf{Cho. P.} &
        \textbf{Conf.} &
        \textbf{Gum. B.} &
        \textbf{Haz. T.} &
        \textbf{Lic. S.} &
        \textbf{Lollip.} &
        \textbf{Marsh.} &
        \textbf{Pep. C.} \\
    
        \rotatebox{90}{\hspace{0.35cm} RGB} & 
            \includegraphics[width=0.1\linewidth]{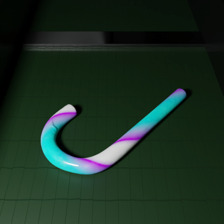} & 
            \includegraphics[width=0.1\linewidth]{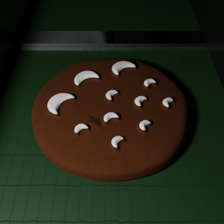} & 
            \includegraphics[width=0.1\linewidth]{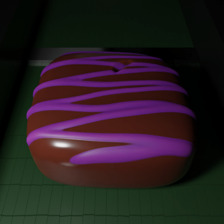} & 
            \includegraphics[width=0.1\linewidth]{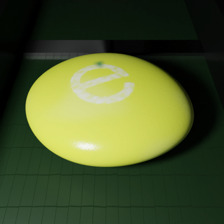} & 
            \includegraphics[width=0.1\linewidth]{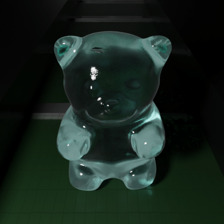} & 
            \includegraphics[width=0.1\linewidth]{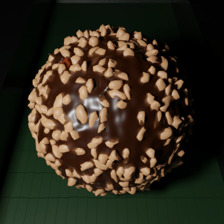} & 
            \includegraphics[width=0.1\linewidth]{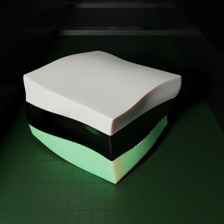} & 
            \includegraphics[width=0.1\linewidth]{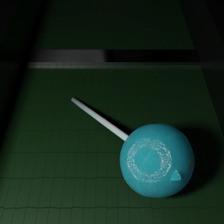} & 
            \includegraphics[width=0.1\linewidth]{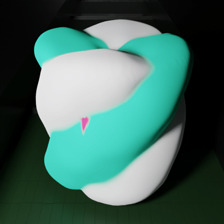} & 
            \includegraphics[width=0.1\linewidth]{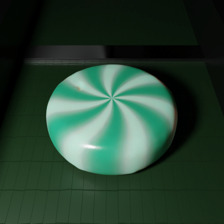} \\
            
        \rotatebox{90}{\hspace{0.45cm} PC} & 
            \includegraphics[width=0.1\linewidth]{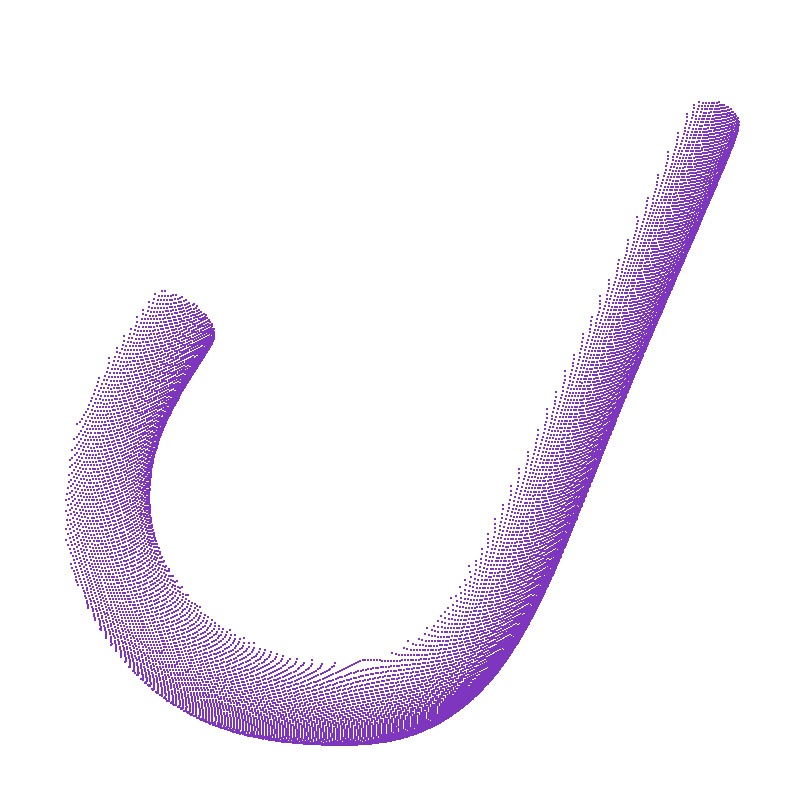} & 
            \includegraphics[width=0.1\linewidth]{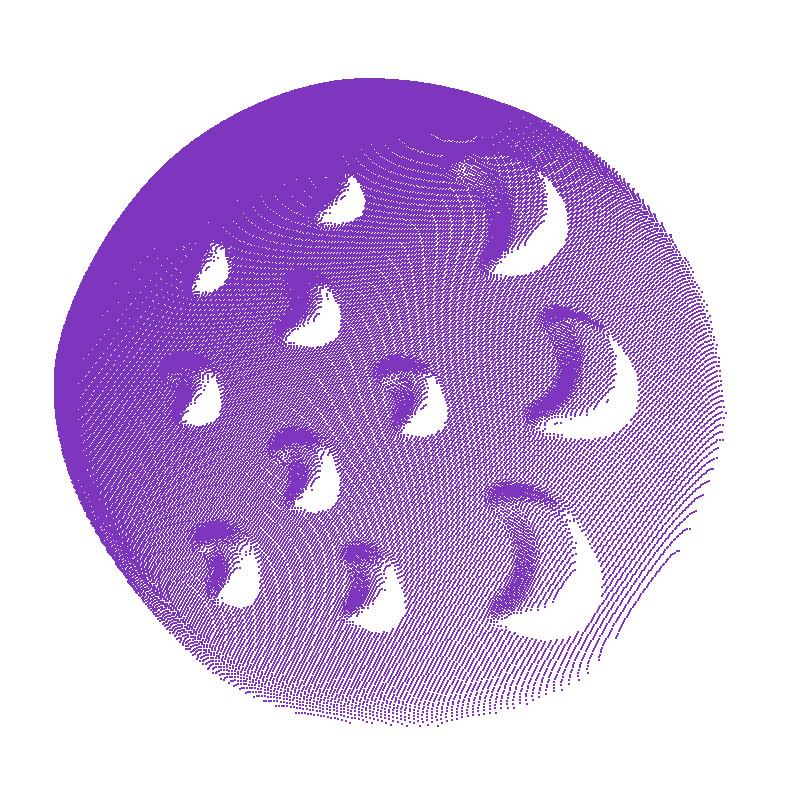} & 
            \includegraphics[width=0.1\linewidth]{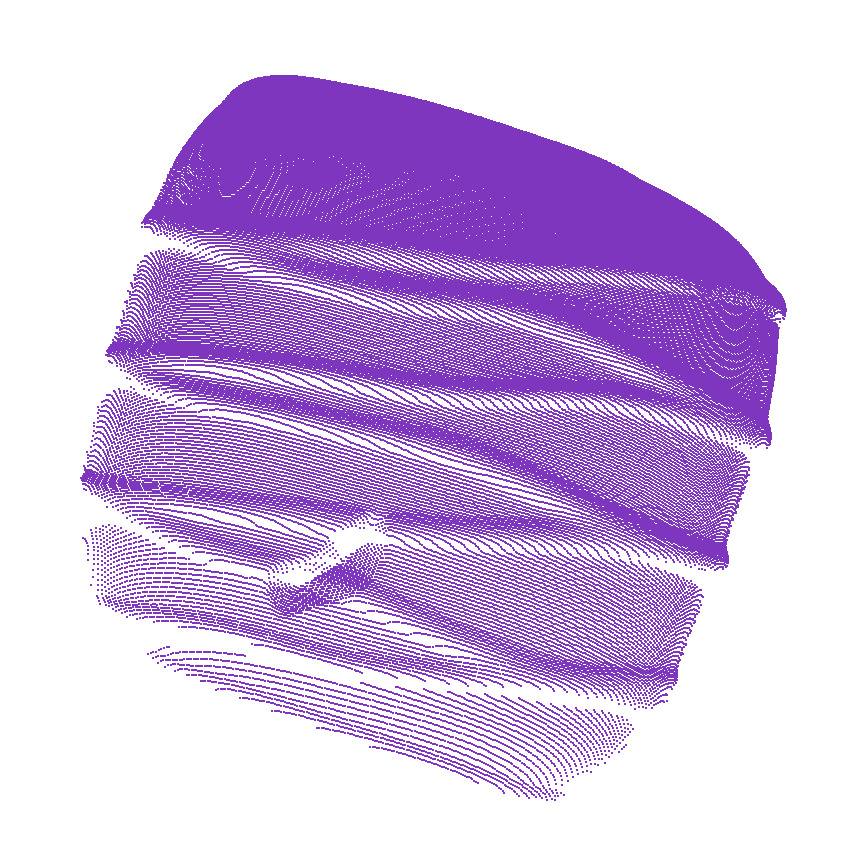} & 
            \includegraphics[width=0.1\linewidth]{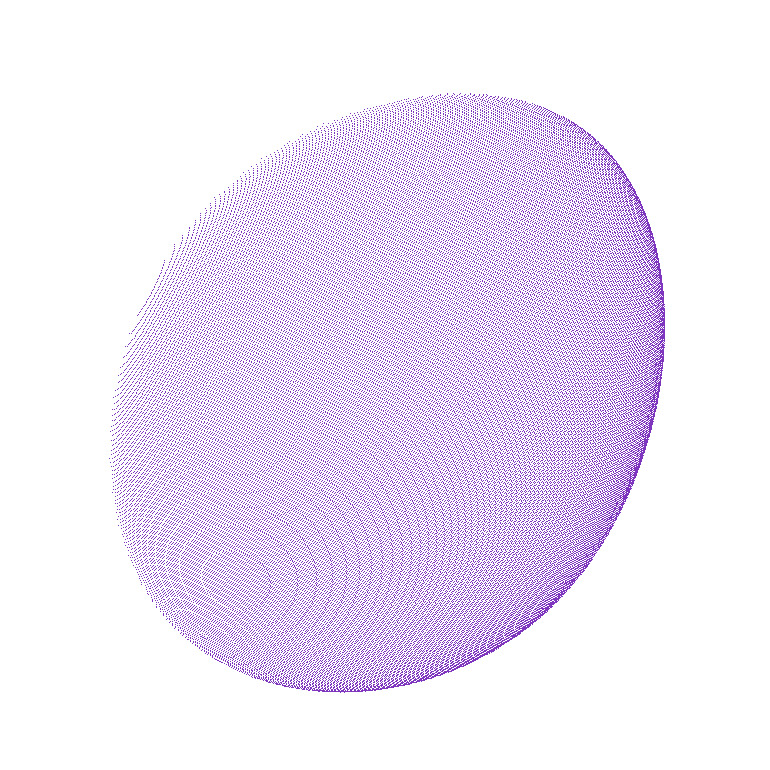} & 
            \includegraphics[width=0.1\linewidth]{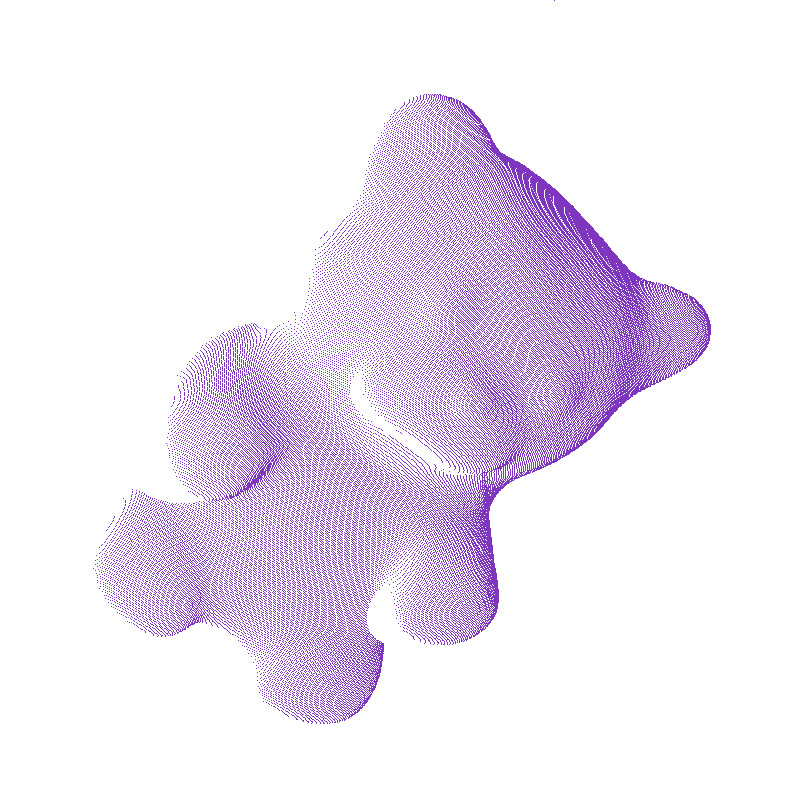} & 
            \includegraphics[width=0.1\linewidth]{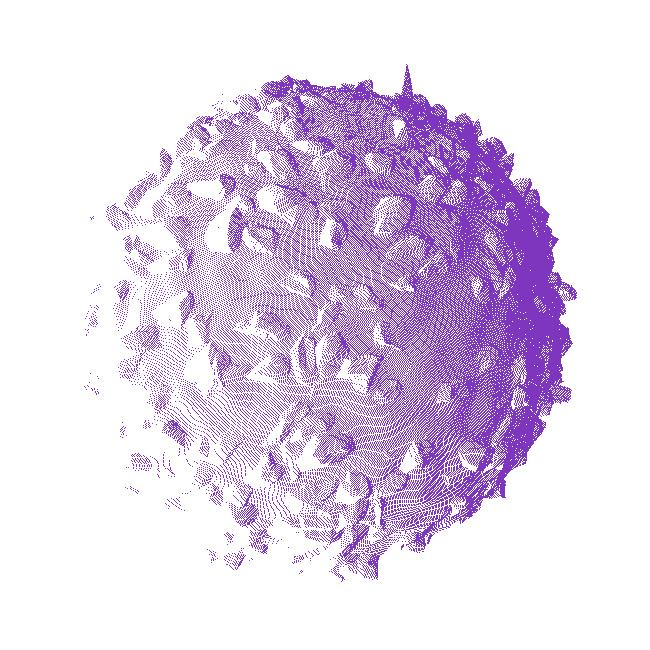} & 
            \includegraphics[width=0.1\linewidth]{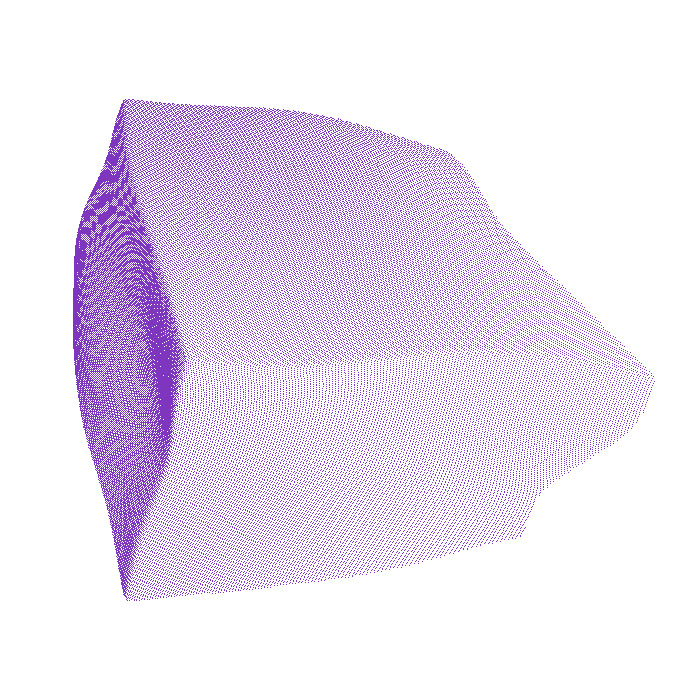} & 
            \includegraphics[width=0.1\linewidth]{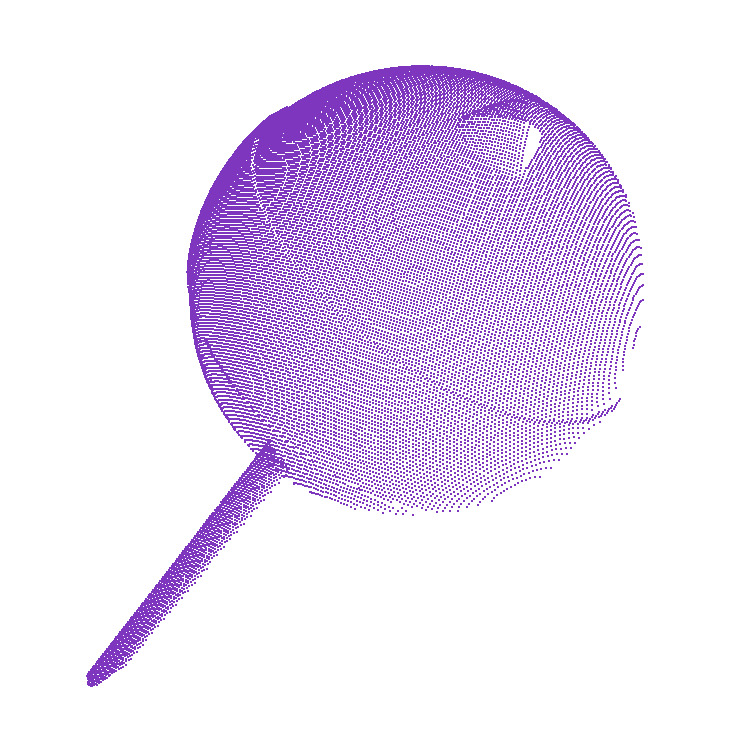} & 
            \includegraphics[width=0.1\linewidth]{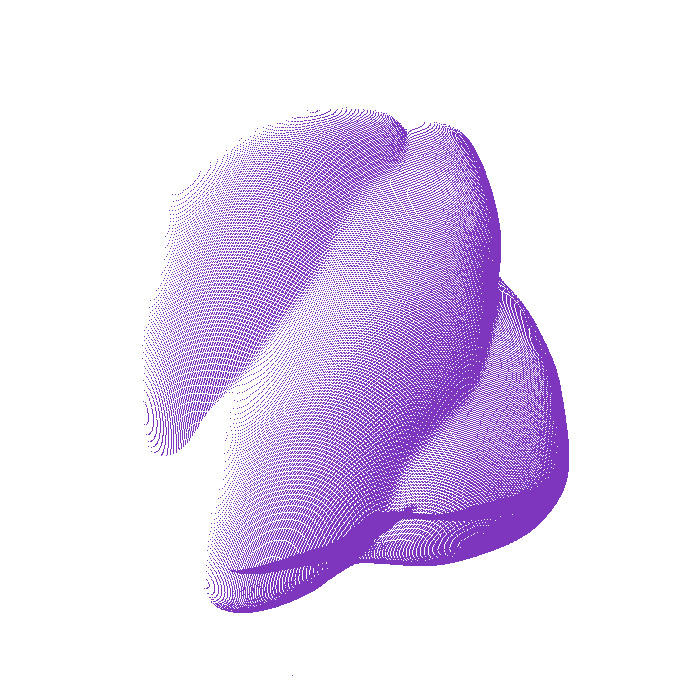} & 
            \includegraphics[width=0.1\linewidth]{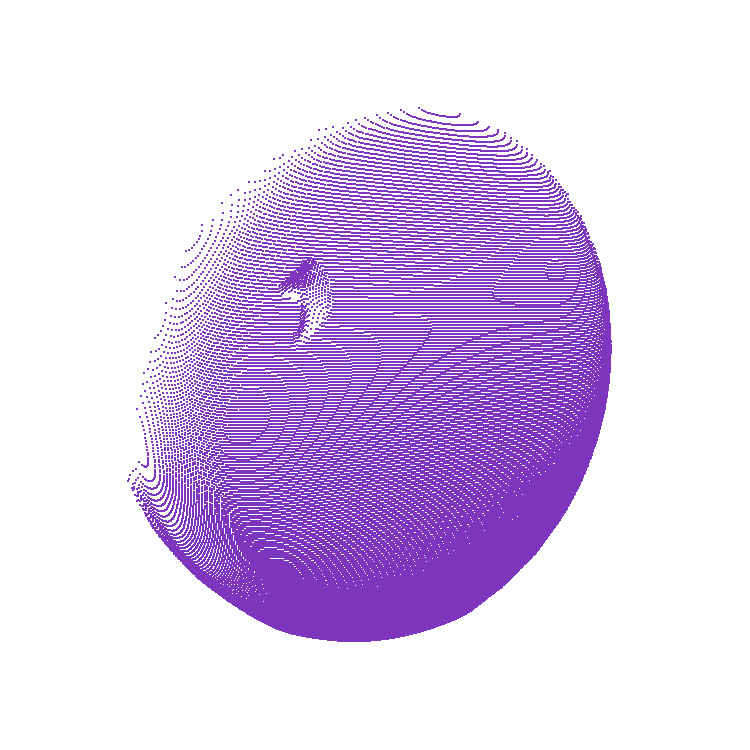} \\
        
        \rotatebox{90}{\hspace{0.45cm} GT} & 
            \includegraphics[width=0.1\linewidth]{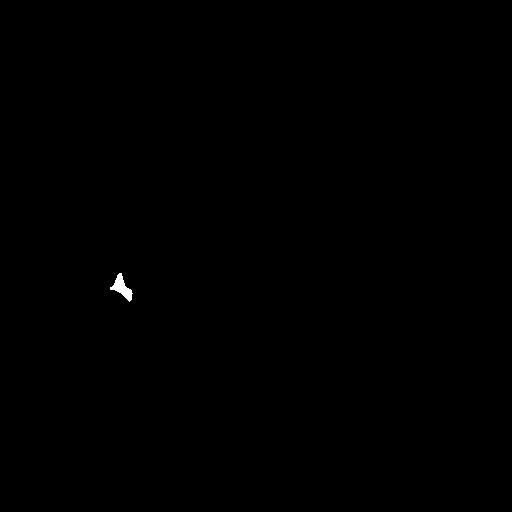} & 
            \includegraphics[width=0.1\linewidth]{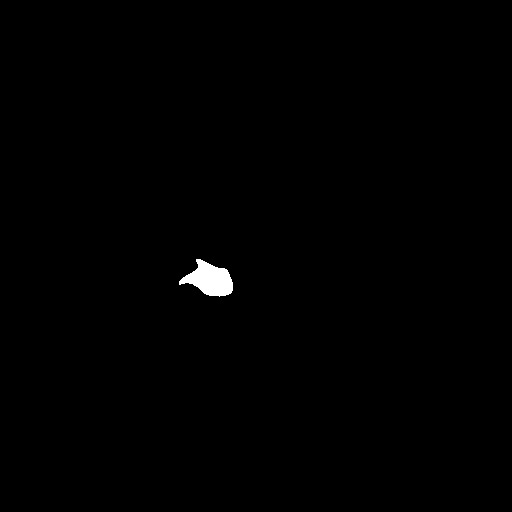} & 
            \includegraphics[width=0.1\linewidth]{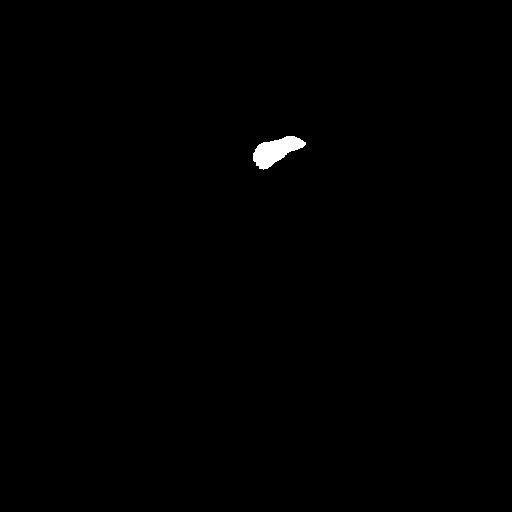} & 
            \includegraphics[width=0.1\linewidth]{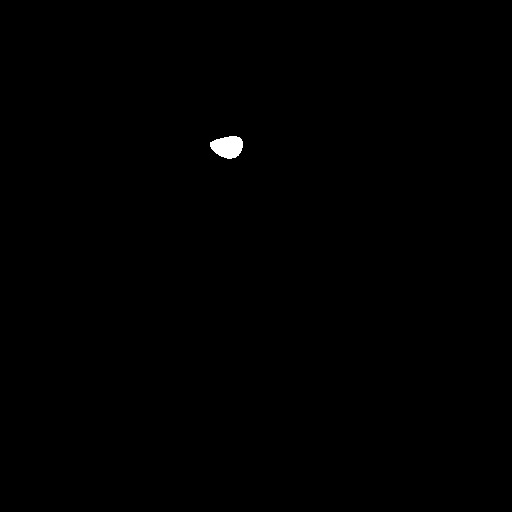} & 
            \includegraphics[width=0.1\linewidth]{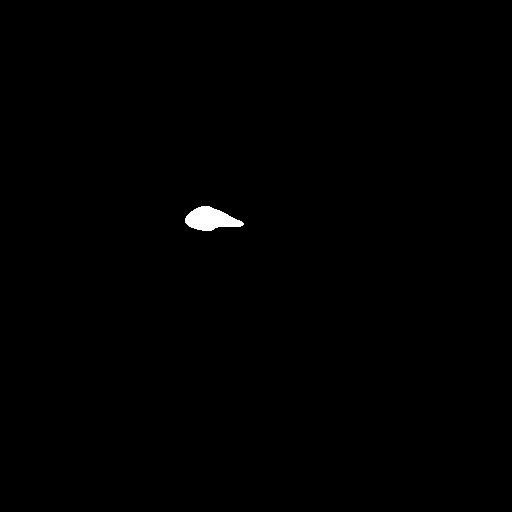} & 
            \includegraphics[width=0.1\linewidth]{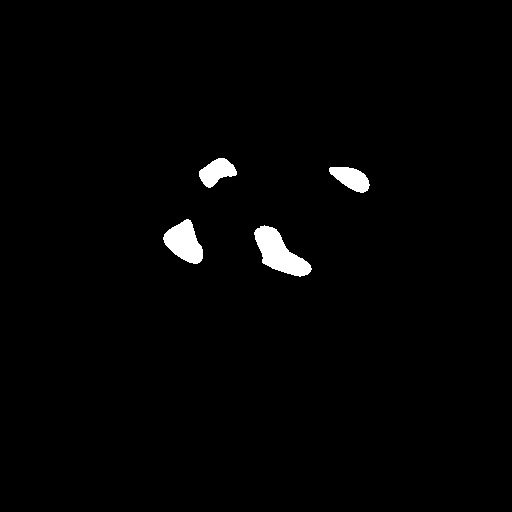} & 
            \includegraphics[width=0.1\linewidth]{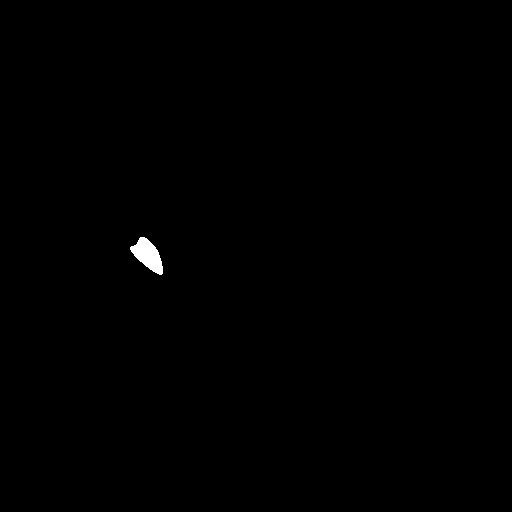} & 
            \includegraphics[width=0.1\linewidth]{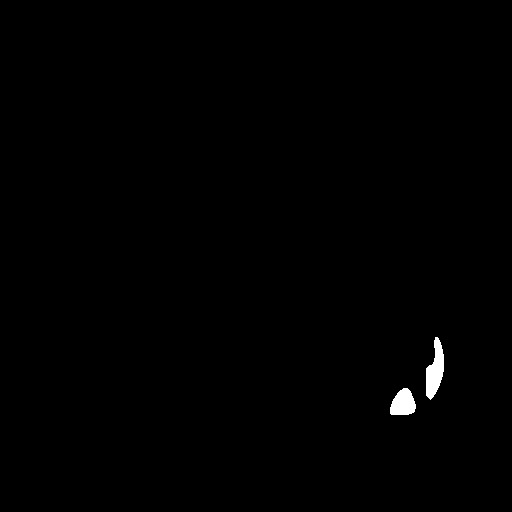} & 
            \includegraphics[width=0.1\linewidth]{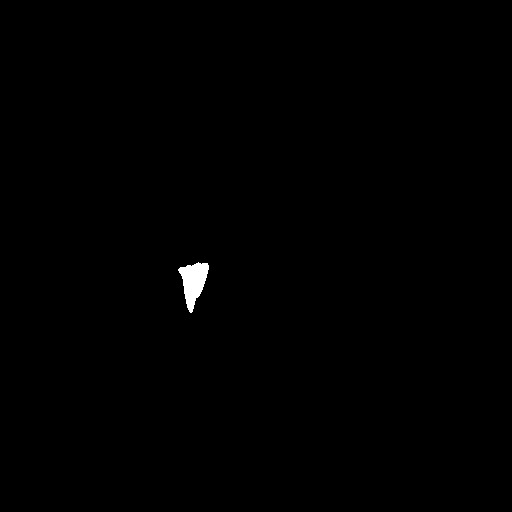} & 
            \includegraphics[width=0.1\linewidth]{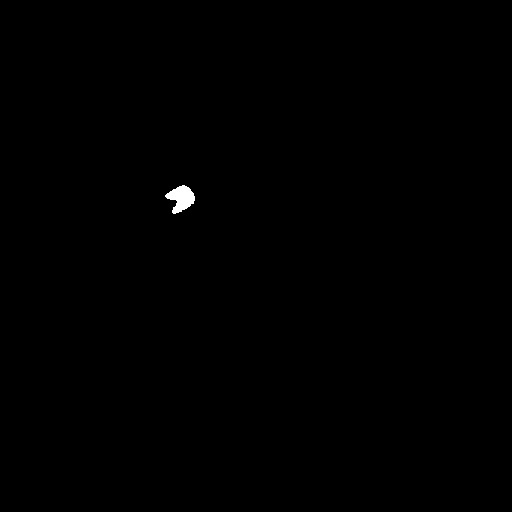} \\
        
        \rotatebox{90}{\hspace{0.30cm} M3DM} & 
            \includegraphics[width=0.1\linewidth]{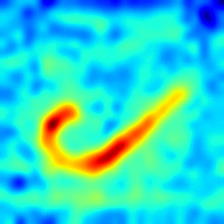} & 
            \includegraphics[width=0.1\linewidth]{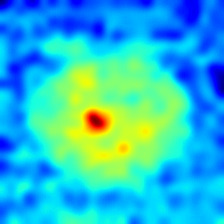} & 
            \includegraphics[width=0.1\linewidth]{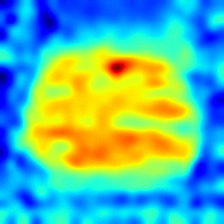} & 
            \includegraphics[width=0.1\linewidth]{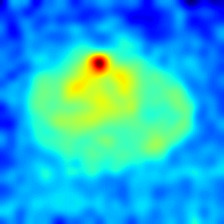} & 
            \includegraphics[width=0.1\linewidth]{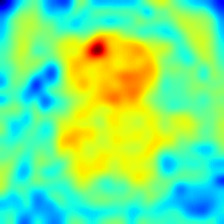} & 
            \includegraphics[width=0.1\linewidth]{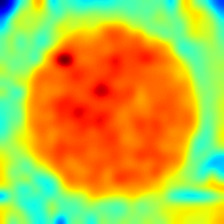} & 
            \includegraphics[width=0.1\linewidth]{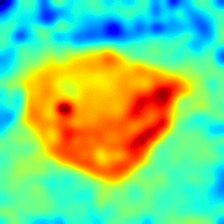} & 
            \includegraphics[width=0.1\linewidth]{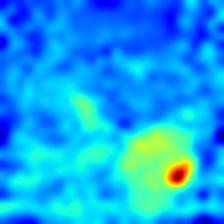} & 
            \includegraphics[width=0.1\linewidth]{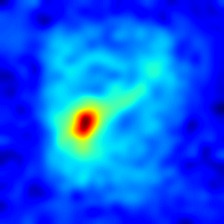} & 
            \includegraphics[width=0.1\linewidth]{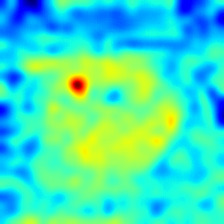} \\
            
        \rotatebox{90}{\hspace{0.35cm} Ours} & 
            \includegraphics[width=0.1\linewidth]{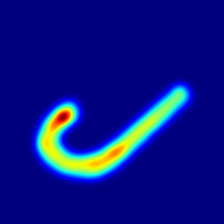} & 
            \includegraphics[width=0.1\linewidth]{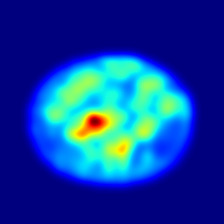} & 
            \includegraphics[width=0.1\linewidth]{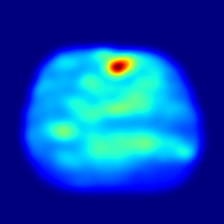} & 
            \includegraphics[width=0.1\linewidth]{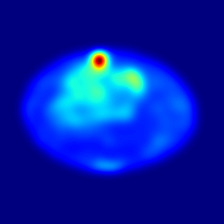} & 
            \includegraphics[width=0.1\linewidth]{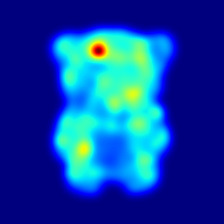} & 
            \includegraphics[width=0.1\linewidth]{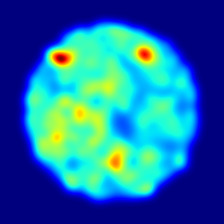} & 
            \includegraphics[width=0.1\linewidth]{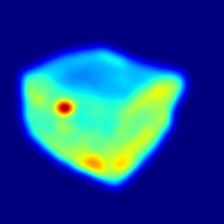} & 
            \includegraphics[width=0.1\linewidth]{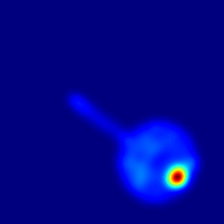} & 
            \includegraphics[width=0.1\linewidth]{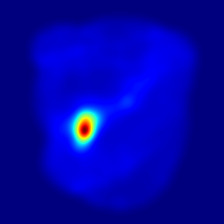} & 
            \includegraphics[width=0.1\linewidth]{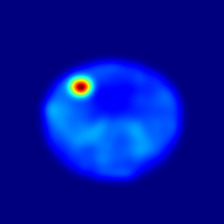} \\

        \\
        \\

        &             
        \textbf{Can. C.} &
        \textbf{Cho. C.} &
        \textbf{Cho. P.} &
        \textbf{Conf.} &
        \textbf{Gum. B.} &
        \textbf{Haz. T.} &
        \textbf{Lic. S.} &
        \textbf{Lollip.} &
        \textbf{Marsh.} &
        \textbf{Pep. C.} \\

        \rotatebox{90}{\hspace{0.35cm} RGB} & 
            \includegraphics[width=0.1\linewidth]{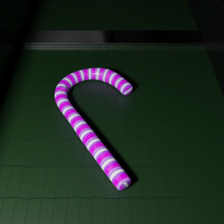} & 
            \includegraphics[width=0.1\linewidth]{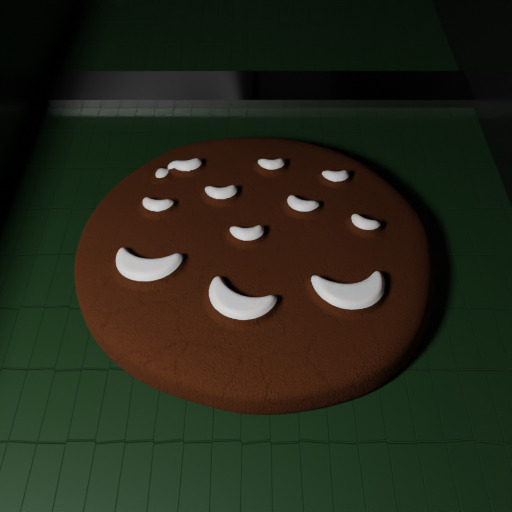} & 
            \includegraphics[width=0.1\linewidth]{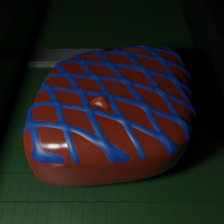} & 
            \includegraphics[width=0.1\linewidth]{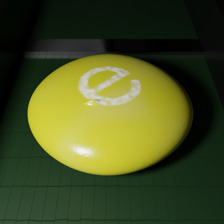} & 
            \includegraphics[width=0.1\linewidth]{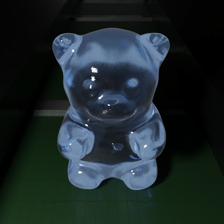} & 
            \includegraphics[width=0.1\linewidth]{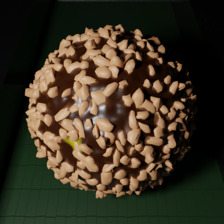} & 
            \includegraphics[width=0.1\linewidth]{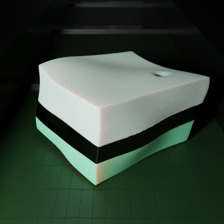} & 
            \includegraphics[width=0.1\linewidth]{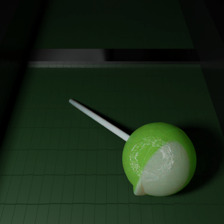} & 
            \includegraphics[width=0.1\linewidth]{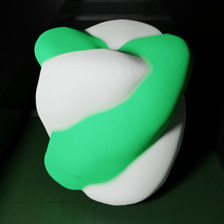} & 
            \includegraphics[width=0.1\linewidth]{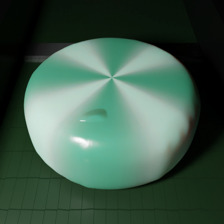} \\
            
        \rotatebox{90}{\hspace{0.45cm} PC} & 
            \includegraphics[width=0.1\linewidth]{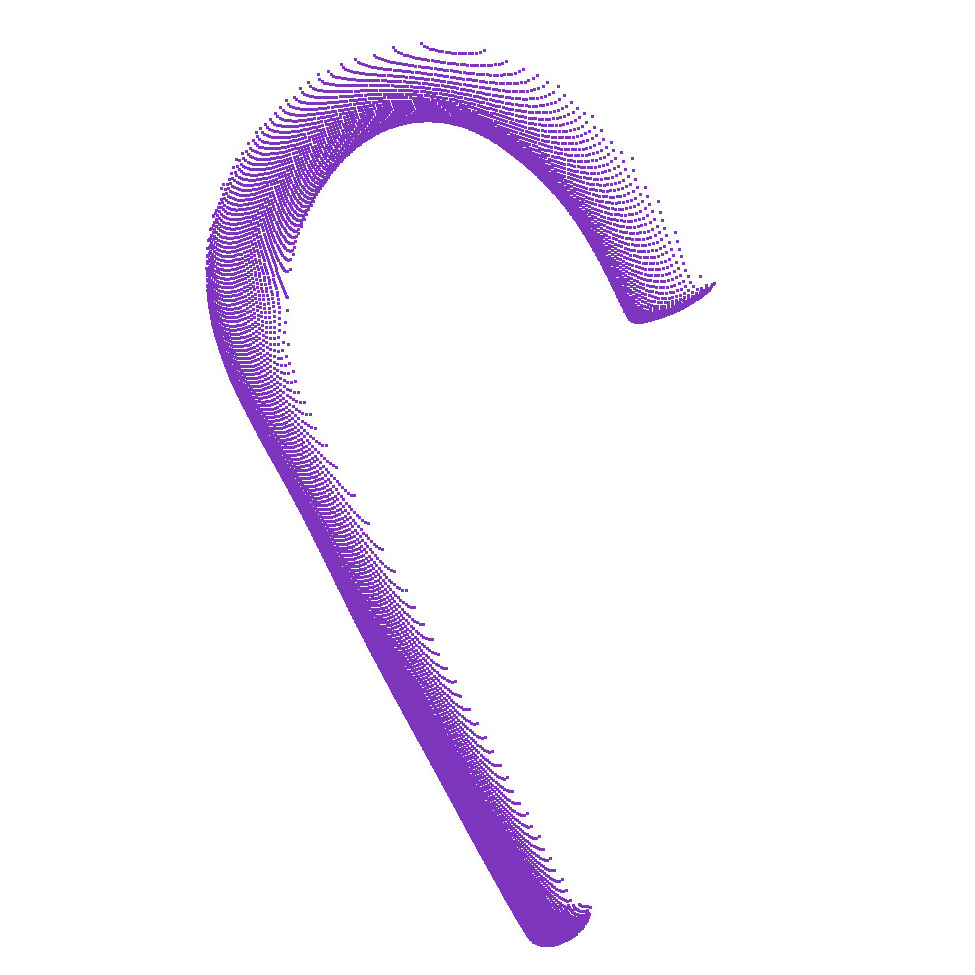} & 
            \includegraphics[width=0.1\linewidth]{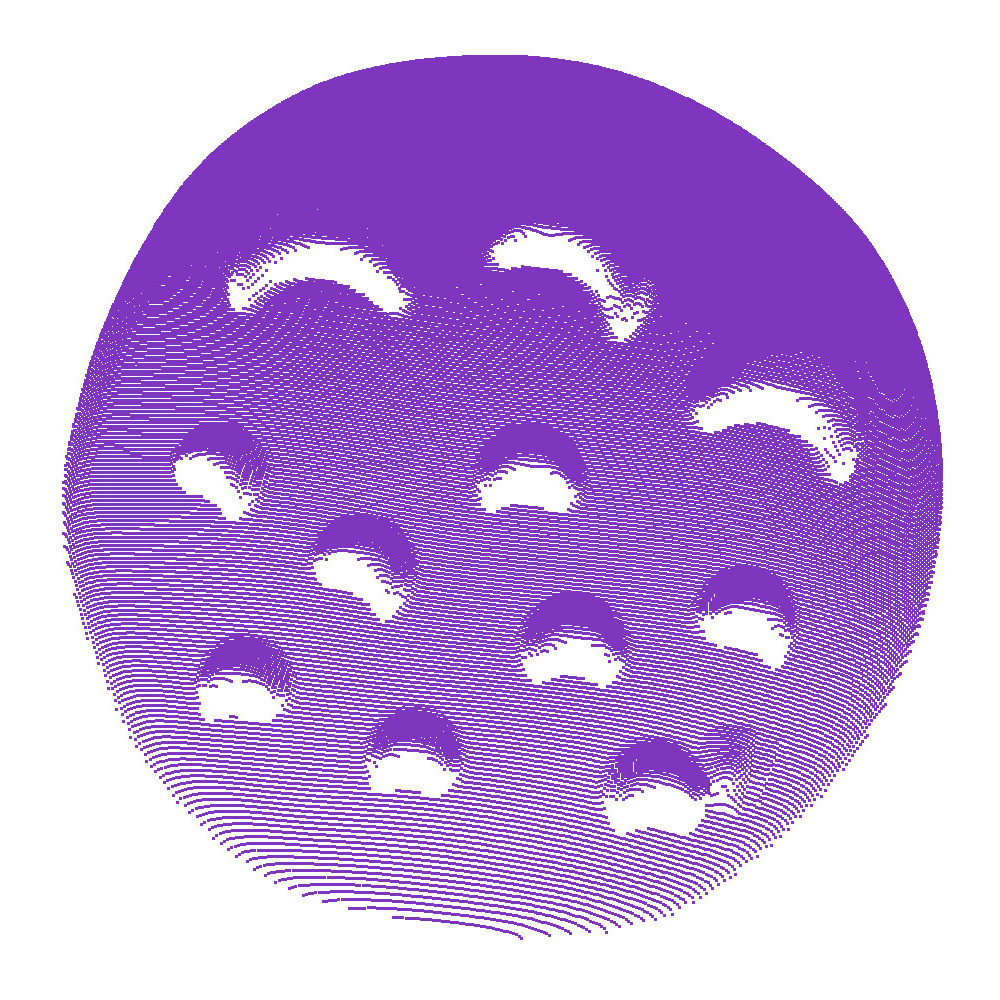} & 
            \includegraphics[width=0.1\linewidth]{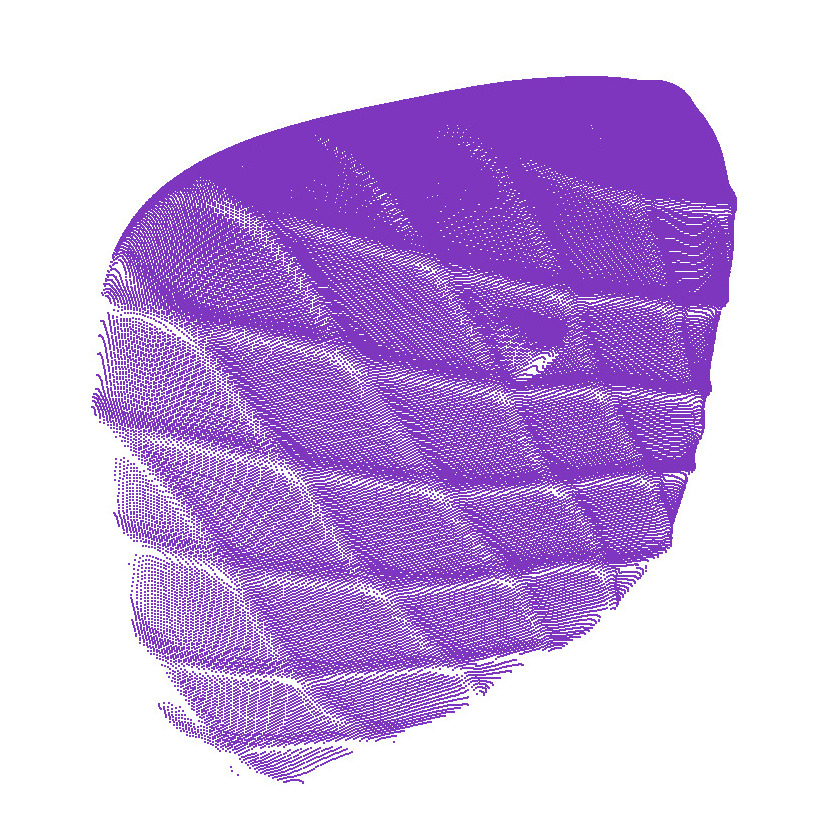} & 
            \includegraphics[width=0.1\linewidth]{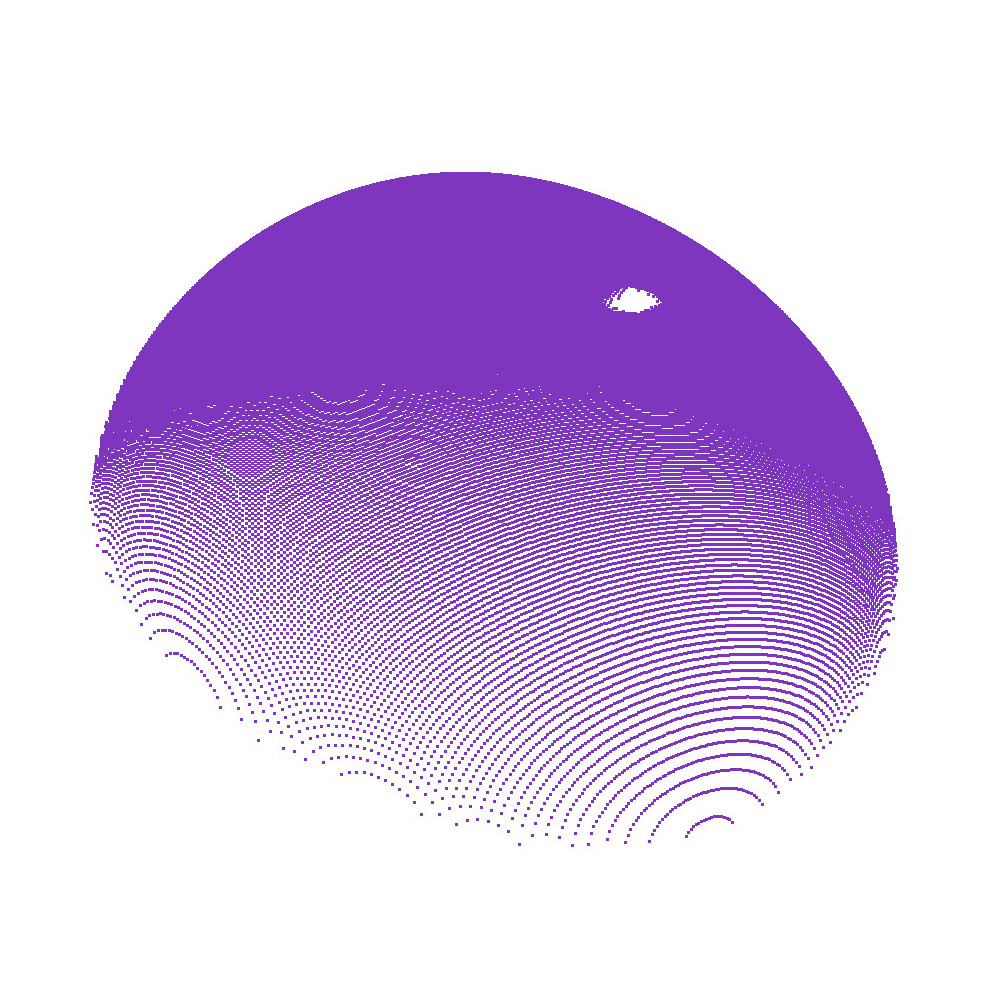} & 
            \includegraphics[width=0.1\linewidth]{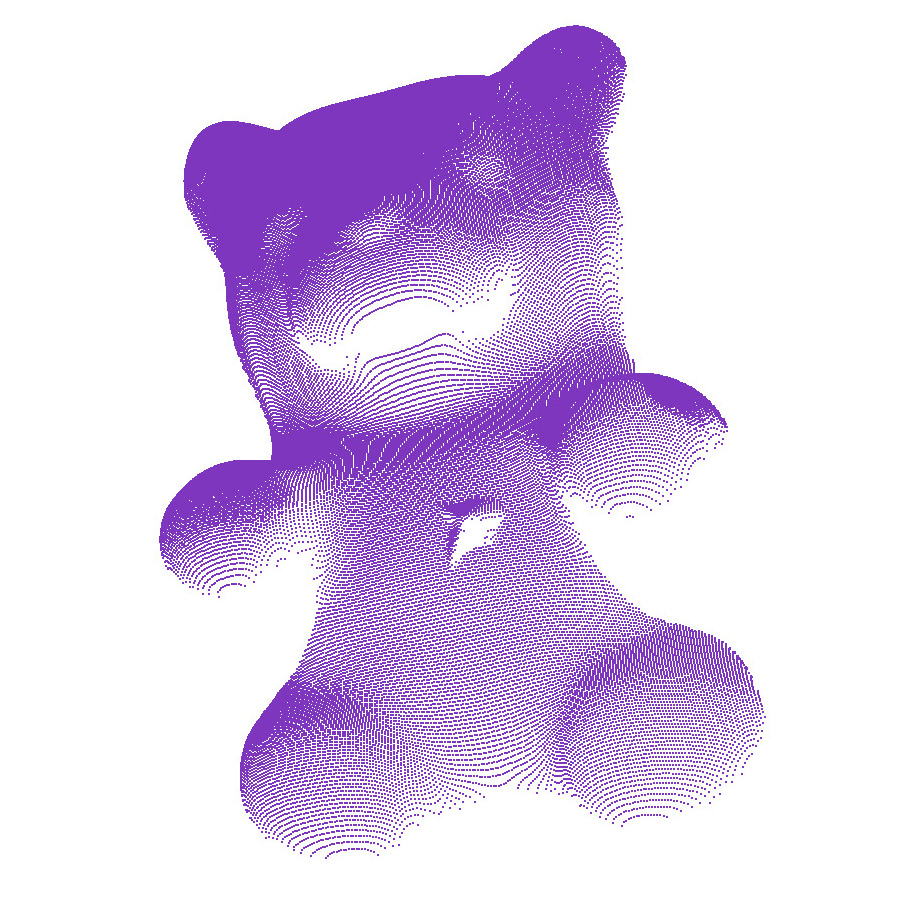} & 
            \includegraphics[width=0.1\linewidth]{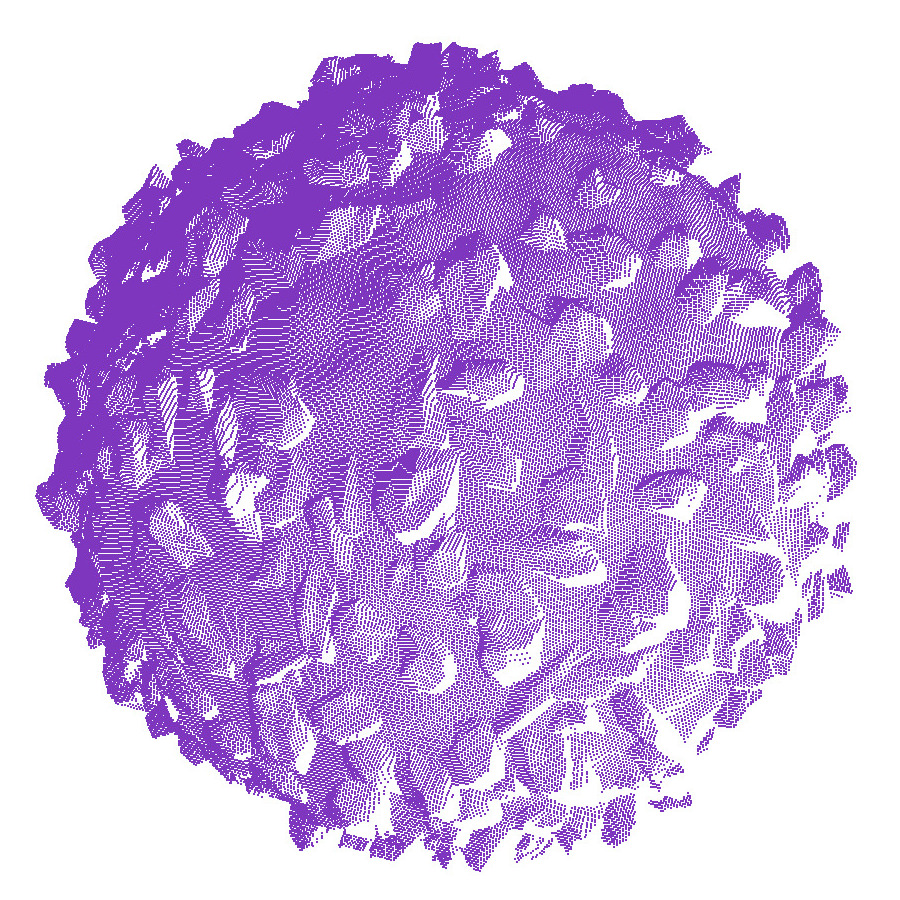} & 
            \includegraphics[width=0.1\linewidth]{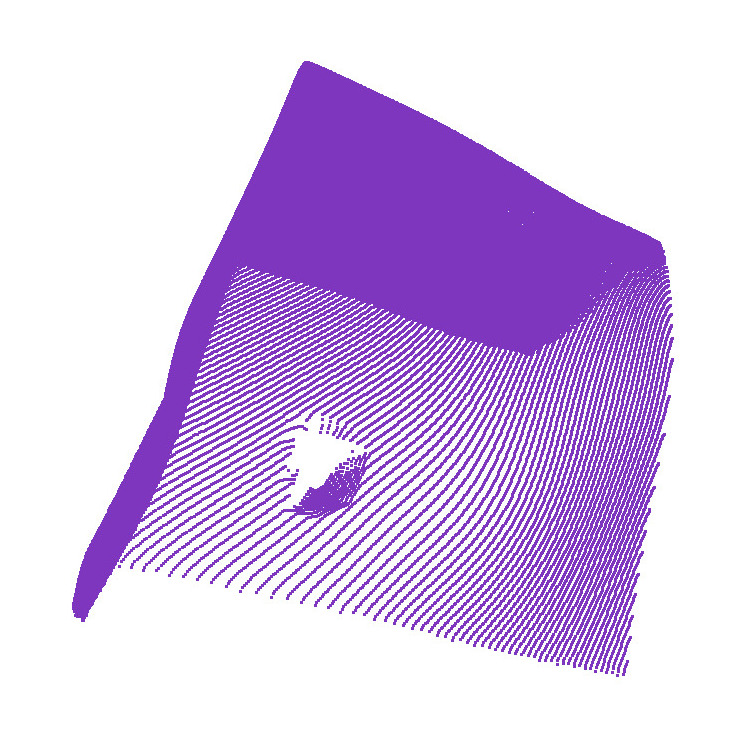} & 
            \includegraphics[width=0.1\linewidth]{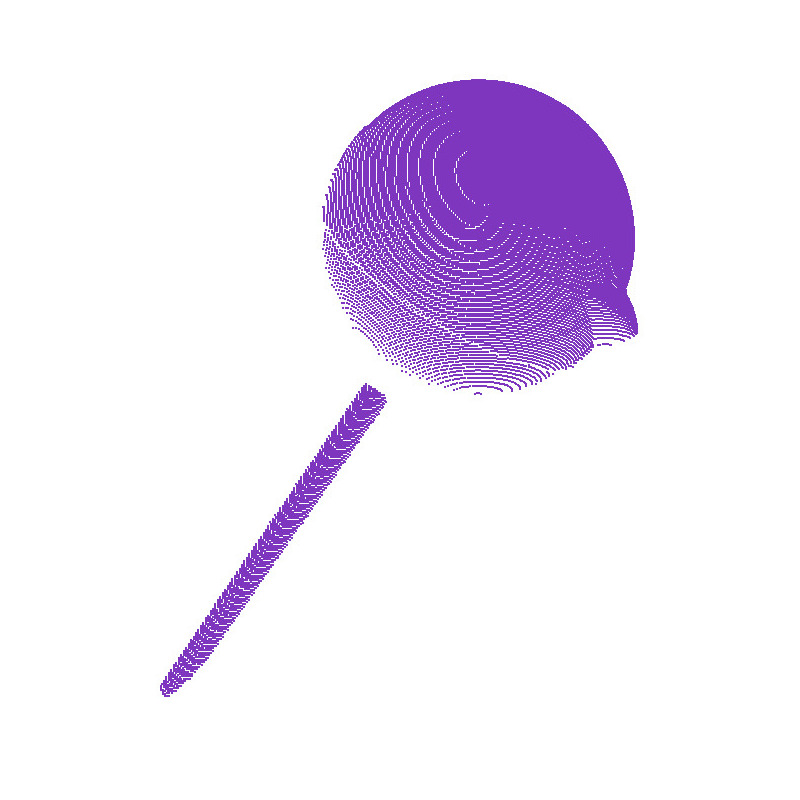} & 
            \includegraphics[width=0.1\linewidth]{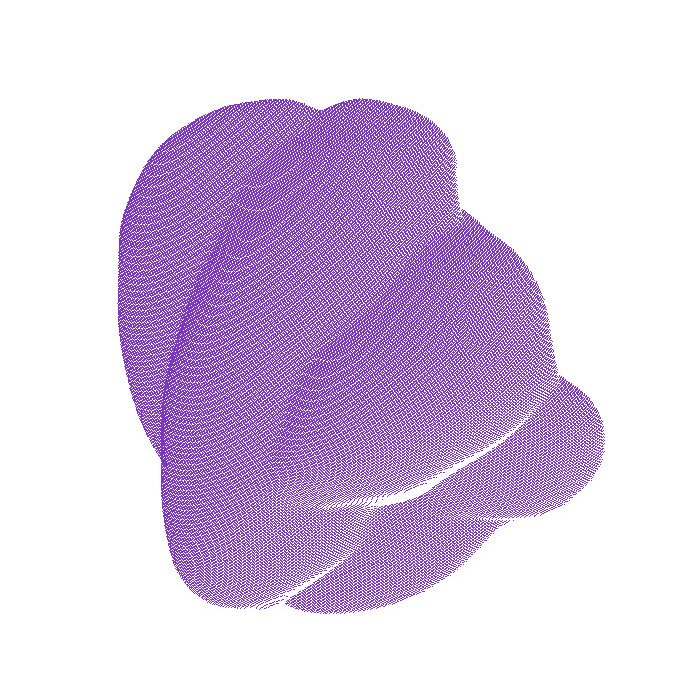} & 
            \includegraphics[width=0.1\linewidth]{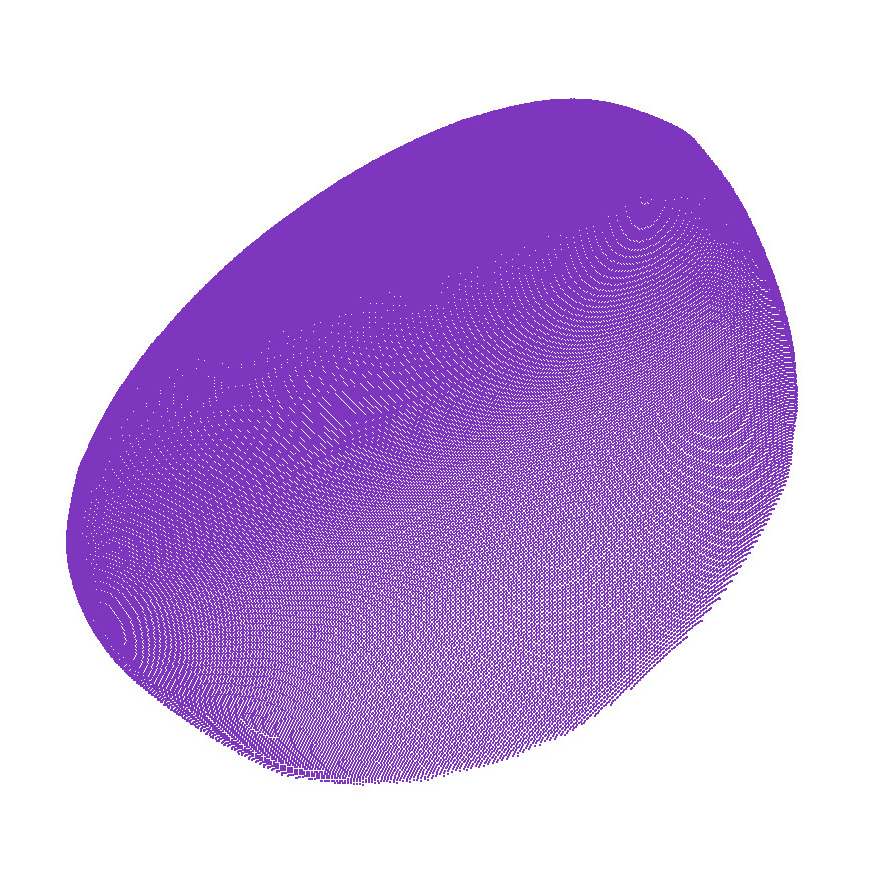} \\
        
        \rotatebox{90}{\hspace{0.45cm} GT} & 
            \includegraphics[width=0.1\linewidth]{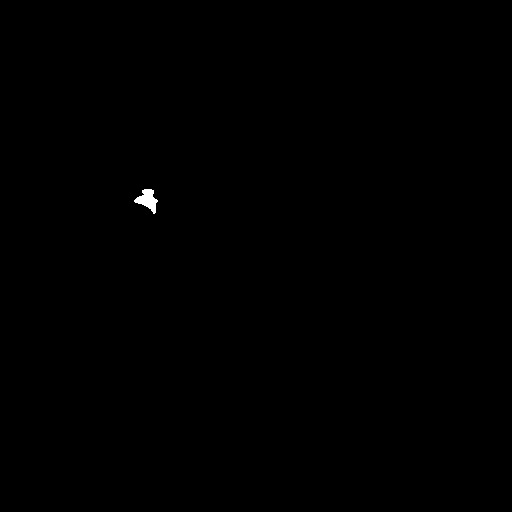} & 
            \includegraphics[width=0.1\linewidth]{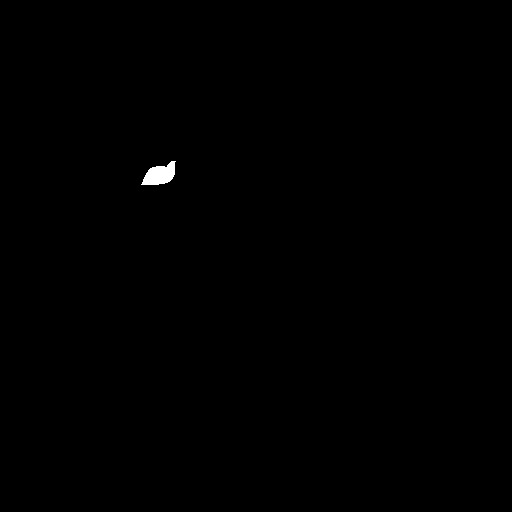} & 
            \includegraphics[width=0.1\linewidth]{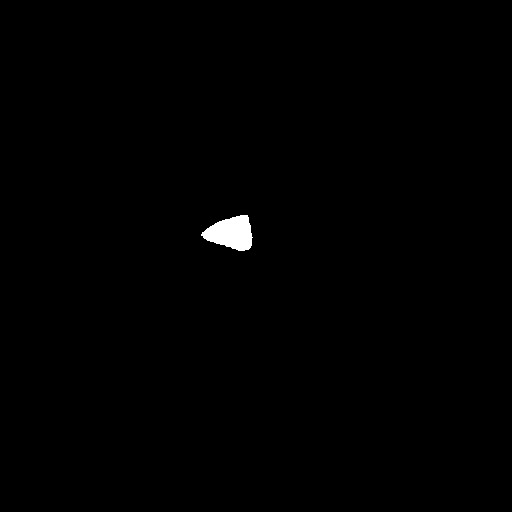} & 
            \includegraphics[width=0.1\linewidth]{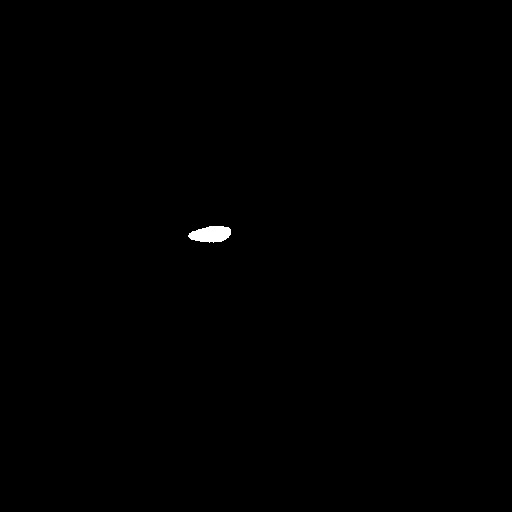} & 
            \includegraphics[width=0.1\linewidth]{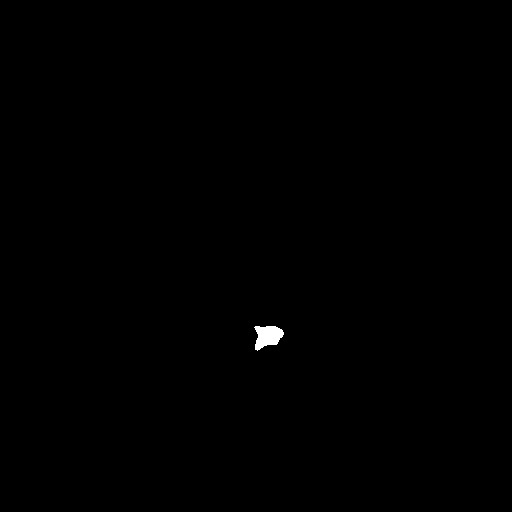} & 
            \includegraphics[width=0.1\linewidth]{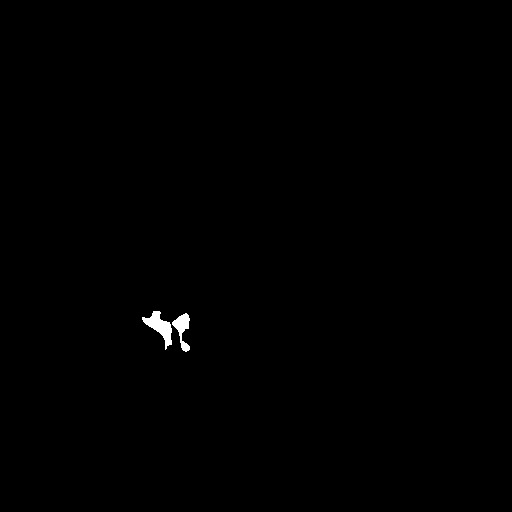} & 
            \includegraphics[width=0.1\linewidth]{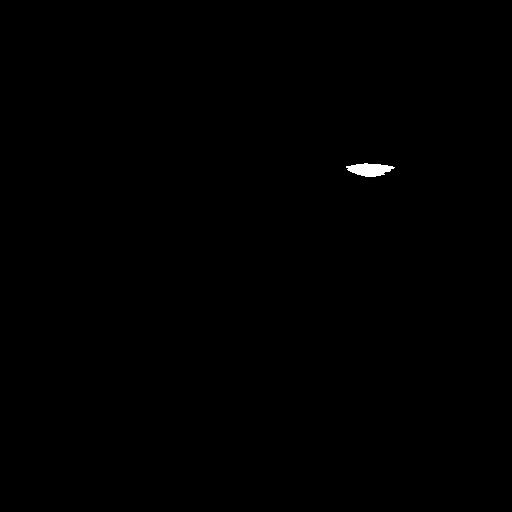} & 
            \includegraphics[width=0.1\linewidth]{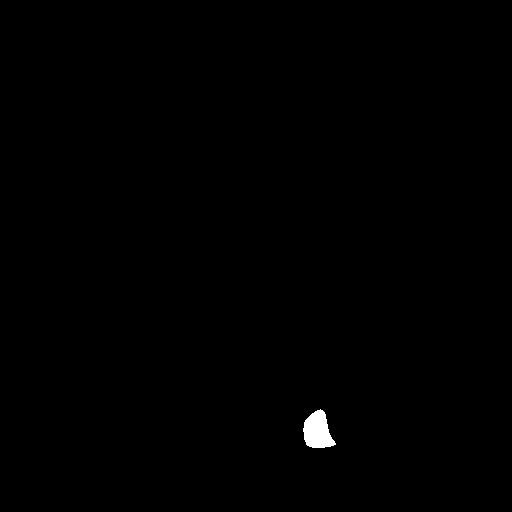} & 
            \includegraphics[width=0.1\linewidth]{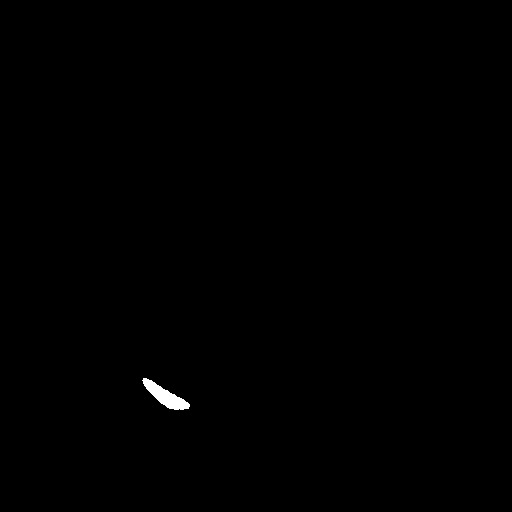} & 
            \includegraphics[width=0.1\linewidth]{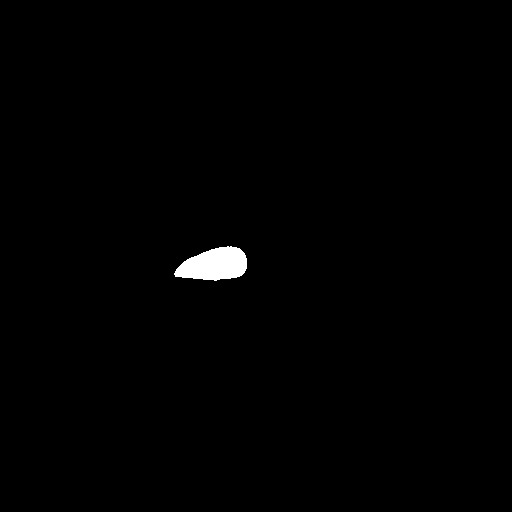} \\

        \rotatebox{90}{\hspace{0.30cm} M3DM} & 
            \includegraphics[width=0.1\linewidth]{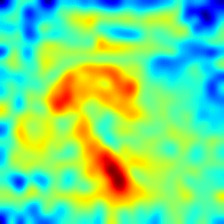} & 
            \includegraphics[width=0.1\linewidth]{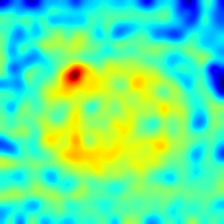} & 
            \includegraphics[width=0.1\linewidth]{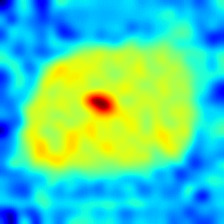} & 
            \includegraphics[width=0.1\linewidth]{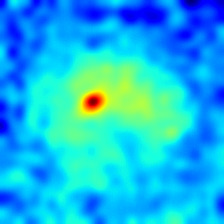} & 
            \includegraphics[width=0.1\linewidth]{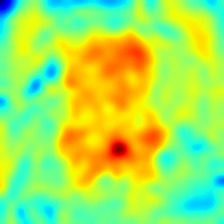} & 
            \includegraphics[width=0.1\linewidth]{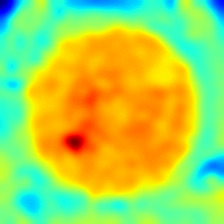} & 
            \includegraphics[width=0.1\linewidth]{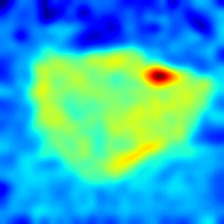} & 
            \includegraphics[width=0.1\linewidth]{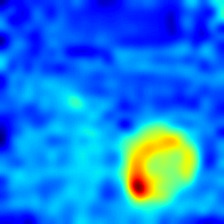} & 
            \includegraphics[width=0.1\linewidth]{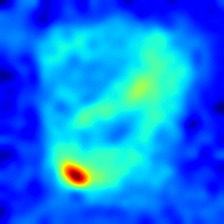} & 
            \includegraphics[width=0.1\linewidth]{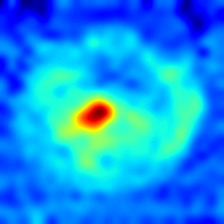} \\
            
        \rotatebox{90}{\hspace{0.35cm} Ours} & 
            \includegraphics[width=0.1\linewidth]{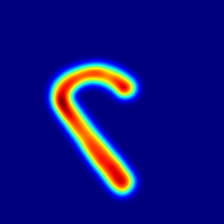} & 
            \includegraphics[width=0.1\linewidth]{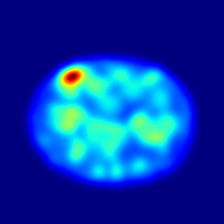} & 
            \includegraphics[width=0.1\linewidth]{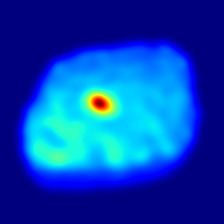} & 
            \includegraphics[width=0.1\linewidth]{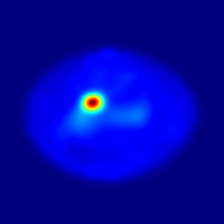} & 
            \includegraphics[width=0.1\linewidth]{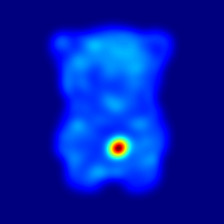} & 
            \includegraphics[width=0.1\linewidth]{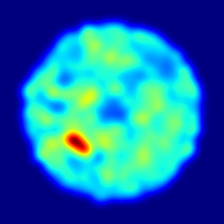} & 
            \includegraphics[width=0.1\linewidth]{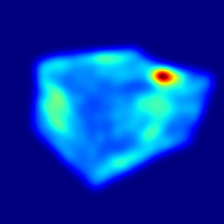} & 
            \includegraphics[width=0.1\linewidth]{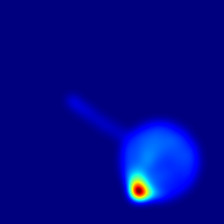} & 
            \includegraphics[width=0.1\linewidth]{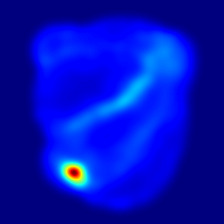} & 
            \includegraphics[width=0.1\linewidth]{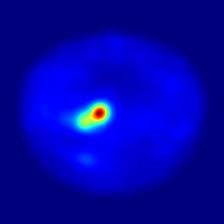} \\
            
      \end{tabular}
  \caption{Qualitative results for each class of the Eyecandies dataset}
  \label{fig:qualitatives_eyecandies_full}
\end{figure*}

        Finally, in \cref{fig:qualitatives_mvtec_full} and \cref{fig:qualitatives_eyecandies_full} we show some additional qualitative results for all the classes of the MVTec 3D-AD and Eyecandies datasets, respectively.
        It is possible to notice how M3DM~\cite{wang2023multimodal} tends to present anomalies on a broader area, highlighting the outline of the underlying object, while our method presents a more localized and less disturbed anomaly map.




\end{document}